%% file: main.tex
\documentclass[runningheads]{llncs}

\usepackage[table,dvipsnames]{xcolor}

\usepackage[mobile]{eccv}

\usepackage{eccvabbrv}

\usepackage{graphicx}
\usepackage{xcolor}
\usepackage{booktabs}
\usepackage{paralist} %
\usepackage{mathtools}
\usepackage{paralist} %
\usepackage{colortbl}
\usepackage{nicefrac}
\usepackage{multirow}
\usepackage{booktabs}
\usepackage{makecell}
\usepackage{gensymb}
\usepackage{tabularx}
\usepackage{comment}

\usepackage[accsupp]{axessibility} 
\usepackage[pagebackref,breaklinks,colorlinks,citecolor=eccvblue]{hyperref}

\usepackage{orcidlink}

\begin{document}

\title{Pixel-GS: Density Control with Pixel-aware Gradient for 3D Gaussian Splatting}

\titlerunning{Pixel-GS}

\author{
  Zheng Zhang\inst{1} \quad
  Wenbo Hu\inst{2}\textsuperscript{\textdagger} \quad
  Yixing Lao\inst{1} \quad \\
  Tong He\inst{3} \quad
  Hengshuang Zhao\inst{1}\textsuperscript{\textdagger}
}

\authorrunning{Z. Zhang et al.}

\institute{
  $^1$The University of Hong Kong \quad
  $^2$Shanghai AI Lab \quad
  $^3$Tencent AI Lab
}

\institute{$^1$The University of Hong Kong \quad $^2$Tencent AI Lab \quad $^3$Shanghai AI Lab}

\maketitle
\let\thefootnote\relax\footnotetext{\textsuperscript{\textdagger}Corresponding author.}

\begin{abstract}
  3D Gaussian Splatting (3DGS) has demonstrated impressive novel view synthesis results while advancing real-time rendering performance.
  However, its efficacy heavily relies on the quality of the initial point cloud, leading to blurring and needle-like artifacts in regions with inadequate initializing points.
  This issue is mainly due to the point cloud growth condition, which only considers the average gradient magnitude of points from observable views, thereby failing to grow for large Gaussians that are observable for many viewpoints while many of them are only covered in the boundaries.
  To address this, we introduce Pixel-GS, a novel approach to take into account the number of pixels covered by the Gaussian in each view during the computation of the growth condition.
  We regard the covered pixel numbers as the weights to dynamically average the gradients from different views, such that the growth of large Gaussians can be prompted.
  As a result, points within the areas with insufficient initializing points can be grown more effectively, leading to a more accurate and detailed reconstruction.
  In addition, we propose a simple yet effective strategy to scale the gradient field according to the distance to the camera, to suppress the growth of floaters near the camera.
  Extensive qualitative and quantitative experiments confirm that our method achieves state-of-the-art rendering quality while maintaining real-time speeds, outperforming on challenging datasets such as Mip-NeRF 360 and Tanks \& Temples.
  Code and demo are available at: \url{https://pixelgs.github.io}
\end{abstract}

\section{Introduction}
\input{sections/introduction}
\section{Related Work}
\input{sections/related_work}
\section{Method}
\input{sections/method}
\section{Experiments}
\input{sections/experiments}
\section{Conclusion}
\input{sections/conclusion}

\appendix 

\section{Additional Results}
Tables~\ref{table:sub1},~\ref{table:sub2_part1},~\ref{table:sub2_part2},~\ref{table:sub3},~\ref{table:sub4_part1} and~\ref{table:sub4_part2} break down the results of Tables~\ref{table:result1},~\ref{table:result2}, and~\ref{table:ablation1} into metrics for each scene.
Our method consistently enhances scene modeling instructions in the vast majority of scenarios, especially in terms of improving the LPIPS index for scene modeling. Notably, LPIPS is more reflective of the human eye's perception of images compared to PSNR and SSIM. Additionally, we strongly suggest that readers watch the videos on the project page for a more direct understanding of how our approach surpasses 3DGS.

In Figure~\ref{subfig}, we showcase the comparative results of Ground Truth, 3DGS with varying point cloud growth thresholds, and Pixel-GS (Ours).
The top and bottom rows correspond respectively to the rendering results and the point clouds that produced these results, along with the SFM point clouds obtained through COLMAP. The rendering quality (in terms of LPIPS $\downarrow$) and memory consumption are displayed in the top right corner of the image.
Through the comparison of the metrics in the top-right corners of (c) and (d), Pixel-GS achieves better reconstruction results with fewer points and significantly enhances the reconstruction capability in areas with insufficient initial SFM points.
It is noteworthy that the points grown by Pixel-GS are more uniformly distributed, while most of the points grown by 3DGS with a reduced threshold are still located in denser areas of the point cloud, which do not significantly enhance the rendering quality. The improvement in rendering performance for 3DGS with a reduced threshold, compared to 3DGS with the original threshold, often comes from additional points grown at the edges of dense point cloud areas. These points can enhance the modeling capability in the surrounding areas where the point cloud is sparser.

\input{sections/table_and_figure/360scene_compare}
\input{sections/table_and_figure/tankscene_compare}

\input{sections/table_and_figure/ablation_mip}
\input{sections/table_and_figure/ablation_tank}

\input{sections/table_and_figure/sub_qua}

\clearpage
\bibliographystyle{splncs04}
\bibliography{main}

\end{document}

%% file: sections/introduction.tex
Novel View Synthesis (NVS) is a fundamental problem in computer vision and graphics.
Recently, 3D Gaussian Splatting (3DGS)~\cite{kerbl20233d} has drawn increasing attention for its explicit point-based representation of 3D scenes and real-time rendering performance.

\input{sections/table_and_figure/qualitative1}

3DGS represents the scene as a set of points associated with geometry (Gaussian scales) and appearance (opacities and colors) attributes.
These attributes can be effectively learned by the differentiable rendering, while the optimization of the point cloud's density is challenging.
3DGS carefully initializes the point cloud using the sparse points produced by the Structure from Motion (SfM) process and presents an adaptive density control mechanism to split or clone the points during the optimization process.
However, this mechanism relies heavily on the initial point cloud's quality and cannot effectively grow points in areas where the initial point cloud is sparse, resulting in blurry or needle-like artifacts in the synthesized images.
In practice, the initial point cloud from SfM unavoidably suffers from insufficient points in areas with repetitive textures and few observations.
As shown in the first and second columns of Figure~\ref{fig:results1}, the blurry regions in the RGB images are well aligned with the areas where few points are initialized, and 3DGS fails to generate enough points in these areas.

In essence, this issue is mainly attributed to the condition of when to split or clone a point.
3DGS decides it by checking whether the average gradient magnitude of the points in the Normalized Device Coordinates (NDC) is larger than a threshold.
The magnitude of the gradient is equally averaged across different viewpoints, and the threshold is fixed.
Large Gaussians are usually visible in many viewpoints, and the size of their projection area varies significantly across views, leading to the number of pixels involved in the gradient calculation varies significantly. 
According to the mathematical form of the Gaussian distribution, a few pixels near the center of the projected Gaussian contribute much more to the gradient than the pixels far away from the center. 
Larger Gaussians often have many viewpoints where the area near the projected center point is not within the screen space, thereby lowering the average gradient, making them difficult to split or clone. 
This issue cannot be solved by merely lowering the threshold, as it would more likely encourage growing points in areas with sufficient points, as shown in the third column of Figure~\ref{fig:results1}, still leaving blurry artifacts in the areas with insufficient points.

In this paper, we propose to consider the calculation of the mean gradient magnitude of points from the perspective of pixels.
During the computation of the average gradient magnitude for a Gaussian, we take into account the number of pixels covered by the Gaussian in each view by replacing the averaging across views with the weighted average across views by the number of covered pixels.
The motivation behind this is to amplify the gradient contribution of large Gaussians while leaving the conditions for splitting or cloning small Gaussians unchanged, such that we can effectively grow points in the areas with large Gaussians.
In the meanwhile, for small Gaussians, the weighted average only slightly impacts the final gradient since the variation of covered pixel numbers across different viewpoints is minimal.
Therefore, the final number of points in areas with sufficient initial points would not change significantly to avoid unnecessary memory consumption and processing time, but importantly, points in areas with insufficient initial points can be effectively grown to reconstruct fine-grained details.
As shown in the last column of Figure~\ref{fig:results1}, our method effectively grows points in areas with insufficient initial points and renders high-fidelity images, while directly lowering the threshold in 3DGS to maintain a similar number of final points fails to render blurring-free results.
Besides, we observe that ``floaters'' tend to appear near the camera, which are points that are not well aligned with the scene geometry and are not contributing to the final rendering.
To this end, we propose to scale the gradient field in NDC space according to the depth value of the points,
thereby suppressing the growth of ``floaters'' near the camera.

To evaluate the effectiveness of our method, we conducted extensive experiments on the challenging Mip-NeRF 360~\cite{barron2022mip} and Tanks~\&~Temples~\cite{knapitsch2017tanks} datasets.
Experimental results validate that our method consistently outperforms the original 3DGS, both quantitatively (17.8\% improvement in terms of LPIPS) and qualitatively.
We also demonstrate that our method is more robust to the sparsity of the initial point cloud by manually discarding a certain proportion (up to 99\%) of the initial SfM point clouds.
In summary, we make the following contributions:
\begin{compactenum}[--]
    \item We analyzed the reason for the blurry artifacts in 3DGS and propose to optimize the number of points from the perspective of pixels, thereby enabling effectively growing points in areas with insufficient initial points.
    \item We present a simple yet effective gradient scaling strategy to suppress the ``floater'' artifacts near the camera.
    \item Our method achieves state-of-the-art performance on the challenging Mip-NeRF 360 and Tanks~\&~Temples datasets and is more robust to the quality of initial points.
\end{compactenum}

%% file: sections/table_and_figure/qualitative1.tex
\begin{figure}[t]
    \centering
    \tiny
    \resizebox{1.0\linewidth}{!}{
        \begin{tabular}{*{4}{>{\centering\arraybackslash}m{0.25\linewidth}}}
            \includegraphics[width=\linewidth]{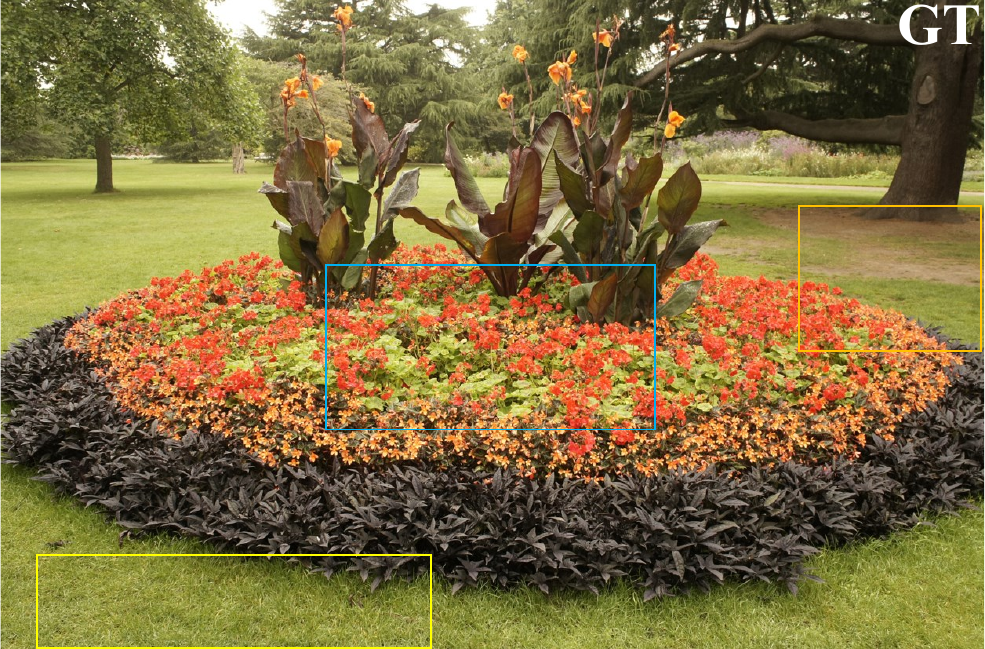} &
            \includegraphics[width=\linewidth]{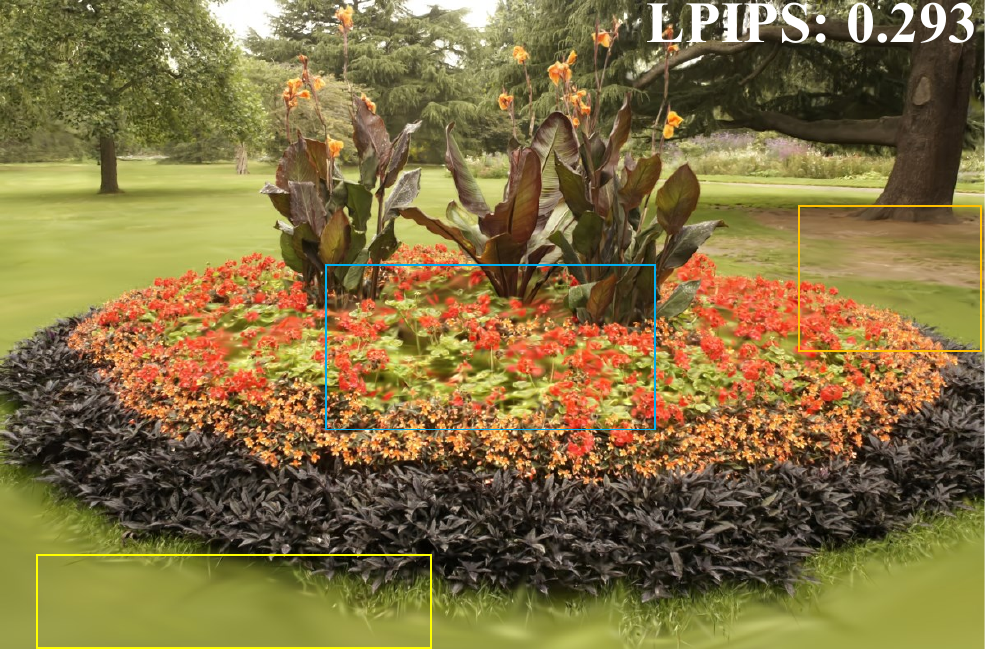} &
            \includegraphics[width=\linewidth]{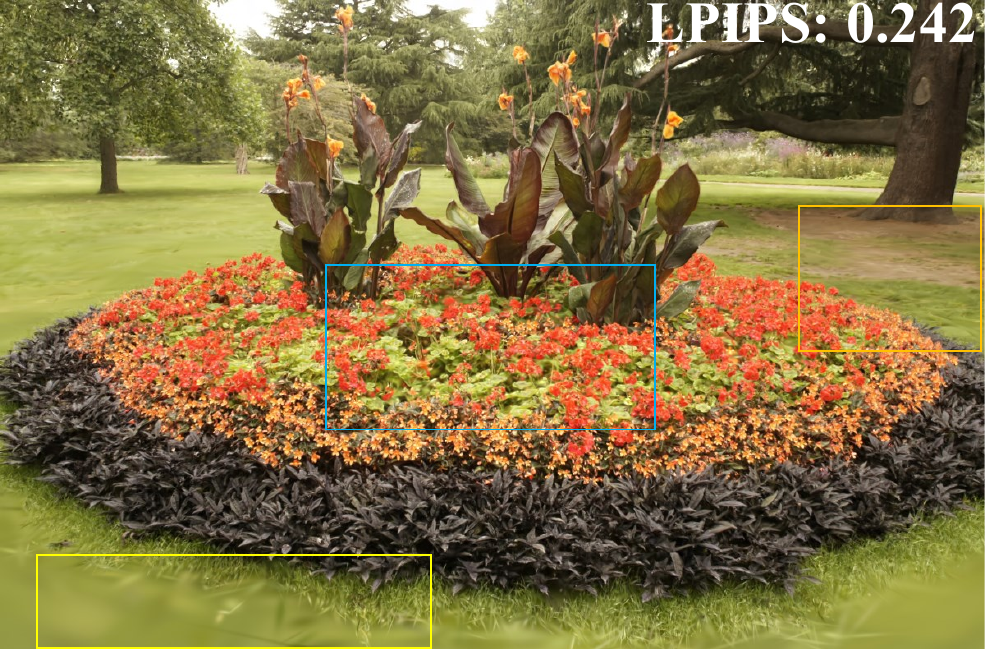} &
            \includegraphics[width=\linewidth]{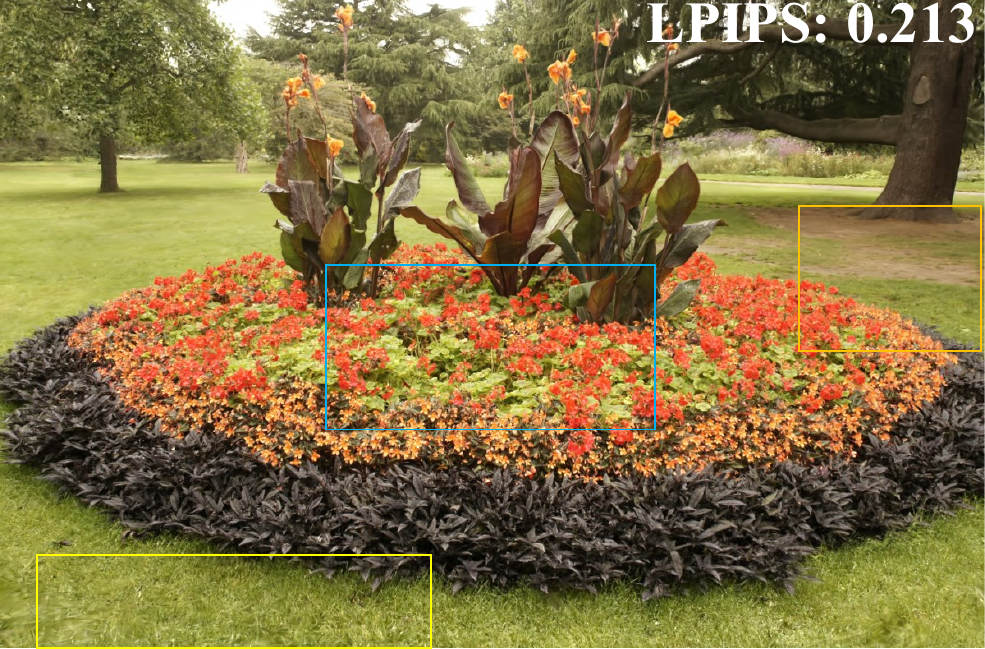} \\
            \includegraphics[width=\linewidth]{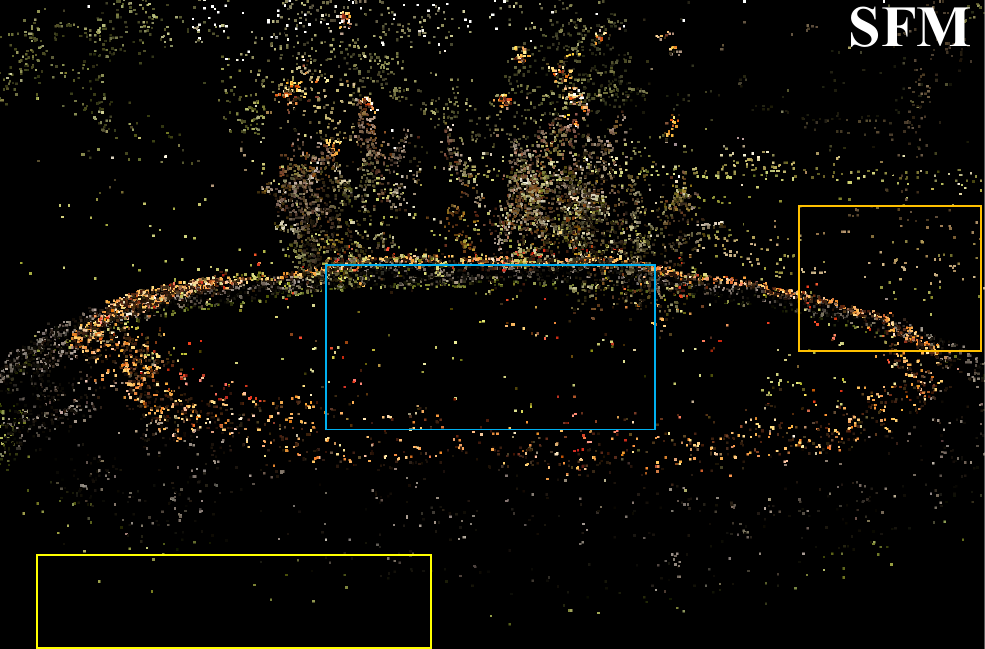} &
            \includegraphics[width=\linewidth]{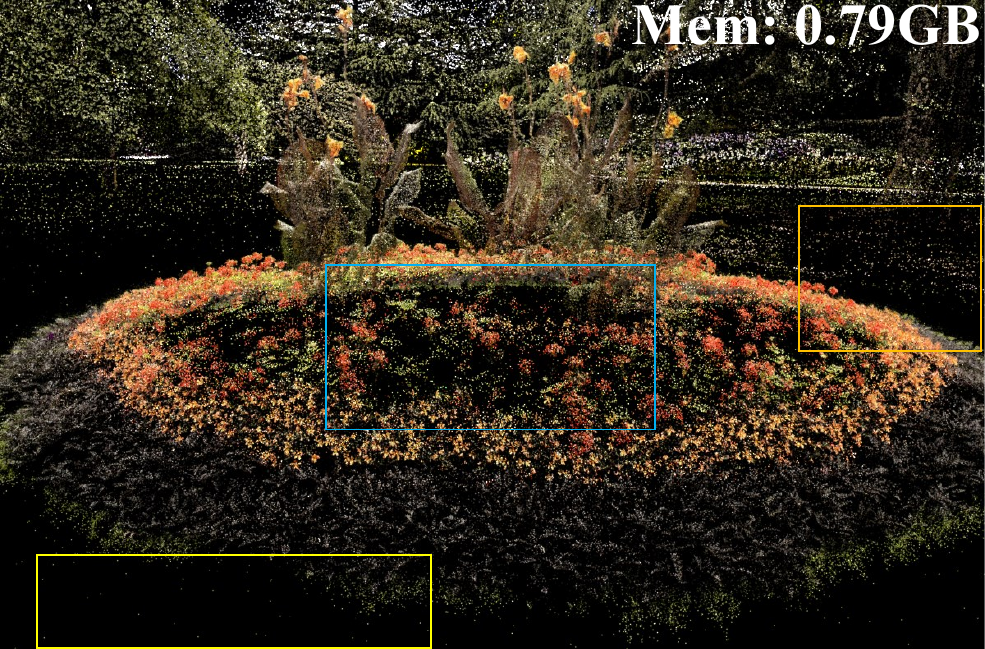} &
            \includegraphics[width=\linewidth]{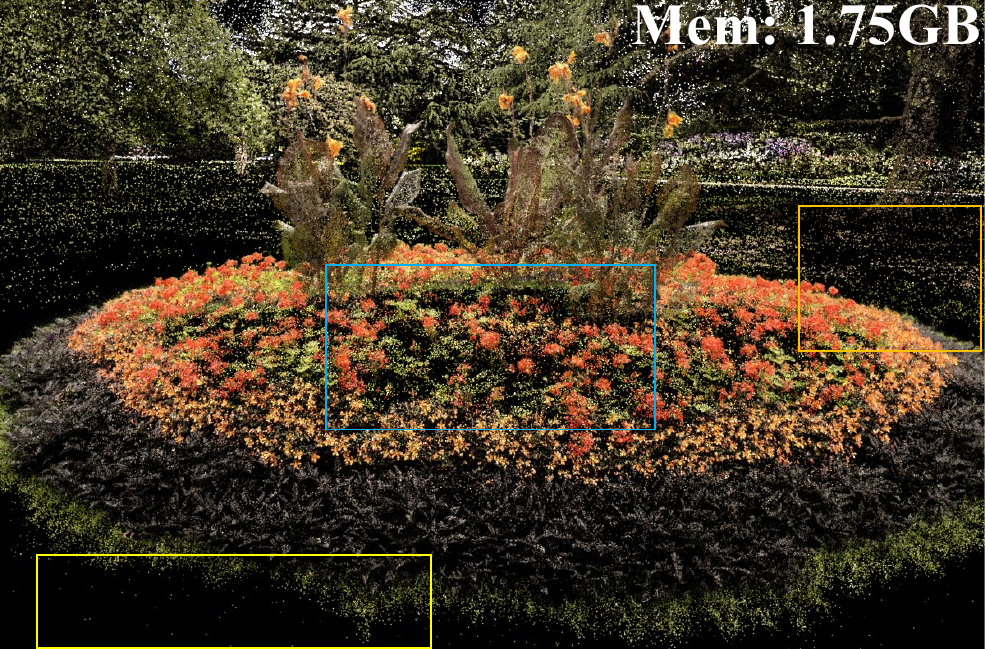} &
            \includegraphics[width=\linewidth]{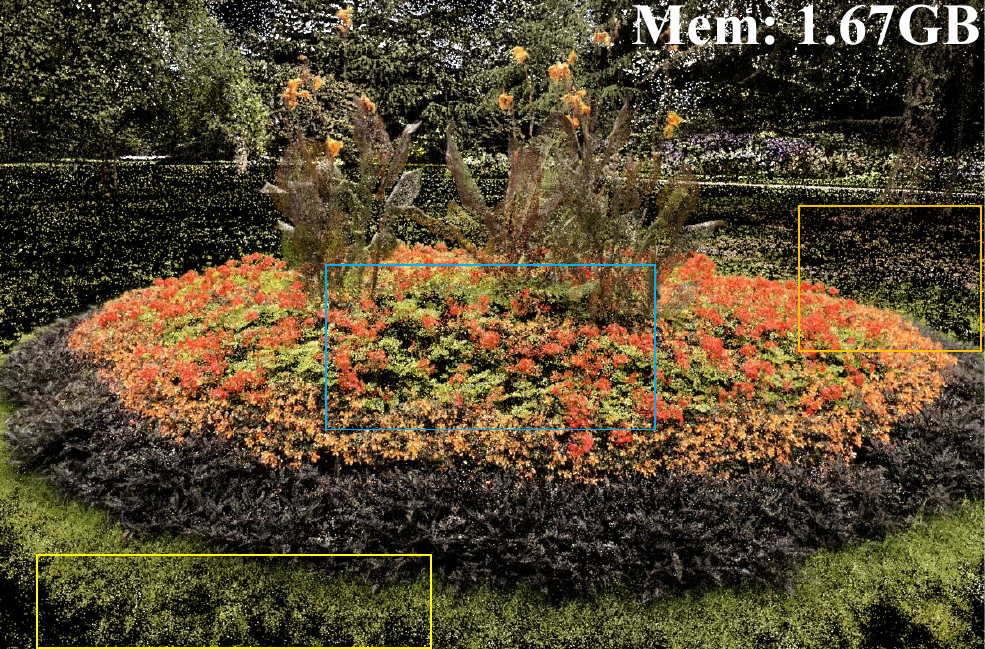} \\
            (a) Ground Truth & (b) 3DGS$^*$ (original threshold) & (c) 3DGS$^*$ (lower threshold) & (d) Pixel-GS (Ours) \\
        \end{tabular}}
    \vspace{1mm} %
    \noindent %
    \parbox{\linewidth}{\small\centering
    To convert b to d, adjust densification from $\frac{\sum{\|\mathbf{g}\|}}{\sum{1}}>\tau_{\text{pos}}$ to $\frac{\sum{\text{pixel}\cdot\|\mathbf{g}\|}}{\sum{\text{pixel}}}>\tau_{\text{pos}}$.
   }
    \vspace{-3mm} %
    \caption{
        Our Pixel-GS effectively grows points in areas with insufficient initializing points (a), leading to a more accurate and detailed reconstruction (d). In contrast, 3D Gaussian Splatting (3DGS) suffers from blurring and needle-like artifacts in these areas, even with a lower threshold of splitting and cloning to encourage more grown points (c). The rendering quality (in LPIPS $\downarrow$) and memory consumption are shown in the results. 3DGS{$^*$} is our retrained 3DGS model with better performance.
    }

    \vspace{-3mm}
    \label{fig:results1}
\end{figure}

%% file: sections/related_work.tex
\noindent\textbf{Novel view synthesis.}
The task of novel view synthesis refers to the process of generating images from perspectives different from the original input viewpoints. Recently, NeRF~\cite{mildenhall2021NeRF} has achieved impressive results in novel view synthesis by using neural networks to approximate the radiance field and employing volumetric rendering~\cite{drebin1988volume, levoy1990efficient, max1995optical, max2005local} techniques for rendering. These approaches use implicit functions (such as MLPs~\cite{mildenhall2021NeRF, barron2021mip, barron2022mip}, feature grid-based representations~\cite{chen2022tensorf, fridovich2022plenoxels, liu2020neural, muller2022instant, sun2022direct}, or feature point-based representations~\cite{kerbl20233d, xu2022point}) to fit the scene's radiance field and utilize a rendering formula for rendering.
Due to the requirement to process each sampled point along a ray through an MLP to obtain its density and color, during the volume rendering, these works significantly suffer from low rendering speed.
Subsequent methods~\cite{hedman2021baking, reiser2021kiloNeRF, reiser2023merf, yariv2023bakedsdf, yu2021plenoctrees} have refined a pre-trained NeRF into a sparse representation, thus achieving real-time rendering of NeRF.
Although some advanced scene representations~\cite{chen2022tensorf, chen2023neurbf, fridovich2022plenoxels, liu2020neural, kulhanek2023tetra, muller2022instant, sun2022direct, barron2021mip, barron2022mip, barron2023zip,hu2023tri} have been proposed to improve one or more aspects of NeRF, such as training cost, rendering results, and rendering speed, 3D Gaussian Splatting (3DGS)~\cite{kerbl20233d} still draws increasing attention due to its explicit representation, high-fidelity results, and real-time rendering speed.
Some subsequent works on 3DGS have further improved it from perspectives such as anti-aliasing~\cite{yu2023mip, yan2023multi}, reducing memory usage~\cite{fan2023lightgaussian, niedermayr2023compressed, navaneet2023compact3d, lee2023compact, morgenstern2023compact, lu2023scaffold}, replacing spherical harmonics functions to enhance the modeling capability of high-frequency signals based on reflective surfaces~\cite{yang2024spec}, and modeling dynamic scenes~\cite{luiten2023dynamic, yang2023deformable, duan20244d, wu20234d, yang2023real, katsumata2023efficient, kratimenos2023dynmf, huang2023sc}.
However, 3DGS still tends to exhibit blurring and needle-like artifacts in areas where the initial points are sparse. This is because 3DGS initializes the scale of each Gaussian based on the distance to neighboring Gaussians, making it challenging for the point cloud growth mechanism of 3DGS to generate sufficient points to accurately model these areas.

\noindent\textbf{Point-based radiance field.}
Point-based representations (such as point clouds) commonly represent scenes using fixed-size, unstructured points, and are rendered by rasterization using GPUs~\cite{botsch2005high, ren2002object, sainz2004point}. Although this is a simple and convenient solution to address topological changes, it often results in holes or outliers, leading to artifacts during rendering. To mitigate issues of discontinuity, researchers have proposed differentiable rendering based on points, utilizing points to model local domains~\cite{gross2011point, insafutdinov2018unsupervised, lin2018learning, yifan2019differentiable, xu2022point, kerbl20233d, wiles2020synsin}. Among these approaches, ~\cite{aliev2020neural, kopanas2021point} employs neural networks to represent point features and utilizes 2D CNNs for rendering. Point-NeRF~\cite{xu2022point} models 3D scenes using neural 3D points and presents strategies for pruning and growing points to repair common holes and outliers in point-based radiance fields. 3DGS~\cite{kerbl20233d} renders using a rasterization approach, which significantly speeds up the rendering process. It starts with a sparse point cloud initialization from SfM and fits each point's influence area and color features using three-dimensional Gaussian distributions and spherical harmonics functions, respectively. To enhance the representational capability of this point-based spatial function, 3DGS introduces a density control mechanism based on the gradient of each point's NDC (Normalized Device Coordinates) coordinates and opacity, managing the growth and elimination of the point cloud. Recent work~\cite{cheng2024gaussianpro} on 3DGS has improved the point cloud growth process by incorporating depth and normals to enhance the fitting ability in low-texture areas.
In contrast, our Pixel-GS does not require any additional priors or information resources, \eg depths and normals, and can directly grow points in areas with insufficient initializing points, reducing blurring and needle-like artifacts.

\noindent\textbf{Floater artifacts.}
Most radiance field scene representation methods encounter floater artifacts, which predominantly appear near the camera and are more severe with sparse input views.
Some papers~\cite{roessle2022dense, chung2023depth} address floaters by introducing depth priors. NeRFshop~\cite{jambon2023NeRFshop} proposes an editing method to remove floaters. Mip-NeRF 360~\cite{barron2022mip} introduces a distortion loss by adding a prior that the density distribution along each ray is unimodal, effectively reducing floaters near the camera. NeRF in the Dark~\cite{mildenhall2022NeRF} suggests a variance loss of weights to decrease floaters. FreeNeRF~\cite{yang2023freeNeRF} introduces a penalty term for the density of points close to the camera as a loss to reduce floaters near the camera. Most of these methods suppress floaters by incorporating priors through loss or editing methods, while ``Floaters No More''~\cite{philip2023floaters} attempts to explore the fundamental reason for the occurrence of floaters and points out that floaters primarily arise because, for two regions of the same volume and shape, the number of pixels involved in the computation is proportional to the inverse square of each region's distance from the camera. Under the same learning rate, areas close to the camera rapidly complete optimization and, after optimization, block the optimization of areas behind them, leading to an increased likelihood of floaters near the camera.
Our method is inspired by this analysis and deals with floaters by a simple yet effective strategy, \ie, scaling the gradient field by the distance to the camera.

%% file: sections/method.tex
We first review the point cloud growth condition of ``Adaptive density control'' in 3DGS. Then, we propose a method for calculating the average gradient magnitude in the point cloud growth condition from a pixel perspective, significantly enhancing the reconstruction capability in areas with insufficient initial points. Finally, we show that by scaling the spatial gradient field that controls point growth, floaters near the input cameras can be significantly suppressed.

\subsection{Preliminaries}

In 3D Gaussian Splatting, Gaussian $i$ under viewpoint $k$ generates a 2D covariance matrix $\varSigma _{2D}^{i,k}=\left( \begin{matrix}
            a^{i,k} & b^{i,k} \\
            b^{i,k} & c^{i,k} \\
        \end{matrix} \right) $, and the corresponding influence range radius $R_{k}^{i}$ can be determined by:
\begin{equation}\label{radii}
    R_{k}^{i}=3\times \left( \frac{a^{i,k}+c^{i,k}}{2}+\sqrt{\left( \frac{a^{i,k}+c^{i,k}}{2} \right) ^2-\left( a^{i,k}c^{i,k}-\left( b^{i,k} \right) ^2 \right)} \right),
\end{equation}
which covers 99$\%$ of the probability in the Gaussian distribution.
For Gaussian $i$, under viewpoint $k$, the coordinates in the camera coordinate system are $\left( \mu _{c,x}^{i,k},\mu _{c,y}^{i,k},\mu _{c,z}^{i,k} \right) $, and in the pixel coordinate system, they are $\left( \mu _{p,x}^{i,k},\mu _{p,y}^{i,k},\mu _{p,z}^{i,k} \right)$. With the image width being $W$ pixels and the height $H$ pixels, Gaussian $i$ participates in the calculation for viewpoint $k$ when it simultaneously satisfies the following six conditions:
\begin{equation}\label{part_cal}
    \begin{cases}
        R_{k}^{i}>0, \mu _{c,z}^{i,k}>0.2,               \\
        -R_{k}^{i}-0.5<\mu _{p,x}^{i,k}<R_{k}^{i}+W-0.5, \\
        -R_{k}^{i}-0.5<\mu _{p,y}^{i,k}<R_{k}^{i}+H-0.5.
    \end{cases}
\end{equation}

In 3D Gaussian Splatting, whether a point is split or cloned is determined by the average magnitude of the gradient of the NDC coordinates for the viewpoints in which the Gaussian participates in the calculation.
Specifically, for Gaussian $i$ under viewpoint $k$, the NDC coordinate is $( \mu _\mathrm{ndc,x}^{i,k},\mu _\mathrm{ndc,y}^{i,k},\mu _\mathrm{ndc,z}^{i,k} )$, and the loss under viewpoint $k$ is $L_k$. During ``Adaptive Density Control'' every 100 iterations, Gaussian $i$ participates in the calculation for $M^i$ viewpoints. The threshold $\tau _\mathrm{pos}$ is set to 0.0002 in 3D Gaussian Splatting. When Gaussian satisfies
\begin{equation}\label{split}
    \frac{\sum_{k=1}^{M^i}{\sqrt{\left( \frac{\partial L_k}{\partial \mu _\mathrm{ndc,x}^{i,k}} \right) ^2+\left( \frac{\partial L_k}{\partial \mu _\mathrm{ndc,y}^{i,k}} \right) ^2}}}{M^i}>\tau _\mathrm{pos},
\end{equation}
it is transformed into two Gaussians.

\input{sections/table_and_figure/method_pic}

\subsection{Pixel-aware Gradient}

Although the current criteria used to decide whether a point should split or clone are sufficient for appropriately distributing Gaussians in most areas, artifacts tend to occur in regions where initial points are sparse. In 3DGS, the lengths of the three axes of the ellipsoid corresponding to Gaussian $i$ are initialized using the values calculated by:
\begin{equation}\label{r_init}
    r^i=\sqrt{\frac{\left( d_{1}^{i} \right) ^2+\left( d_{2}^{i} \right) ^2+\left( d_{3}^{i} \right) ^2}{3}},
\end{equation}
where $d_{1}^{i}$, $d_{2}^{i}$, and $d_{3}^{i}$ are the distances to the three nearest points to Gaussian $i$, respectively. We observed that areas inadequately modeled often have very sparse initial SfM point clouds, leading to the initialization of Gaussians in these areas with ellipsoids having larger axis lengths. This results in their involvement in the computation from too many viewpoints. These Gaussians exhibit larger gradients only in viewpoints where the center point, after projection, is within or near the pixel space. This implies that, from these viewpoints, the large Gaussians cover a larger area in the pixel space after projection. This results in these points having a smaller average gradient size of their NDC coordinates during the ``Adaptive Density Control'' process every 100 iterations (Eq.~\ref{split}), because they participate in the computation from too many viewpoints and only have significant gradient sizes in individual viewpoints. Consequently, it is difficult for these points to split or clone, leading to poor modeling in these areas.

Below, we analyze through equations why the Gaussians in the previously mentioned sparser areas can only obtain larger NDC coordinate gradients from viewpoints with sufficient coverage, whereas for viewpoints that only affect the edge areas, the NDC coordinate gradients are smaller.
The contribution of a pixel under viewpoint $k$ to the NDC coordinate gradient of Gaussian $i$ can be computed as:
\begin{equation}\label{dL_dndc_i_k}
    \left( \begin{array}{c}
            \frac{\partial L_k}{\partial \mu _\mathrm{ndc,x}^{i,k}} \\
            \frac{\partial L_k}{\partial \mu _\mathrm{ndc,y}^{i,k}} \\
        \end{array} \right) =\sum_{pix=1}^{m_{k}^{i}}{\sum_{j=1}^3{\left( \frac{\partial L_k}{\partial c_{j}^{pix}}\times \frac{\partial c_{j}^{pix}}{\partial \alpha _{k,pix}^{i}}\times \left( \begin{array}{c}
            \frac{\partial \alpha _{k,pix}^{i}}{\partial \mu _\mathrm{ndc,x}^{i,k}} \\
            \frac{\partial \alpha _{k,pix}^{i}}{\partial \mu _\mathrm{ndc,y}^{i,k}} \\
        \end{array} \right) \right)}},
\end{equation}
where both $\frac{\partial \alpha _{k,pix}^{i}}{\partial \mu _\mathrm{ndc,x}^{i,k}}$ and $\frac{\partial \alpha _{k,pix}^{i}}{\partial \mu _\mathrm{ndc,y}^{i,k}}$ contain factor $\alpha _{k}^{i}$, which can be calculated as:
\begin{equation}\label{alpha_i}
    \alpha _{k,pix}^{i}=\sigma ^i\times \exp \left( -\frac{1}{2}\left( \begin{array}{c}
                pix_x-\mu _{p,x}^{i,k} \\
                pix_y-\mu _{p,y}^{i,k} \\
            \end{array} \right) ^T\left( \varSigma _{2D}^{i,k} \right) ^{-1}\left( \begin{array}{c}
                pix_x-\mu _{p,x}^{i,k} \\
                pix_y-\mu _{p,y}^{i,k} \\
            \end{array} \right) \right),
\end{equation}
where $c_{j}^{pix}$ represents the color of the $j$th channel of the current pixel, and $m_{k}^{i}$ represents the number of pixels involved in the calculation for Gaussian $i$ under viewpoint $k$. $\alpha _{k,pix}^{i}$ as a function of the distance between the center of the projected Gaussian and the pixel center, exhibits exponential decay as the distance increases.

This results in a few pixels close to the center position of the projected Gaussian making a primary contribution to the NDC coordinate gradient of this Gaussian. For large Gaussians, many viewpoints will only affect the edge areas, projecting onto pixels in these viewpoints, leading to the involvement of these viewpoints in the calculation but with very small NDC coordinate gradients. On the other hand, we observe that for these points, for a given viewpoint, when a large number of pixels are involved in the calculation after projection, these points often exhibit larger gradients of NDC coordinates in this viewpoint. This is easy to understand because, when a large number of pixels are involved in the calculation after projection, the projected center point tends to be within the pixel plane, and according to previous calculations, a few pixels near the center point are the main contributors to the gradient of the NDC coordinates.

To solve this problem, we assign a weight to the gradient size of the NDC coordinates for each Gaussian at every viewpoint, where the weight is the number of pixels involved in the computation for that Gaussian from the corresponding viewpoint. The advantage of this computational approach is that, for large Gaussians, the number of pixels involved in the calculations varies significantly across different viewpoints. According to previous derivations, these large Gaussians only receive larger gradients in viewpoints where a higher number of pixels are involved in the calculations. Weighting the magnitude of gradients by the number of participating pixels in an average manner can more rationally promote the splitting or cloning of these Gaussians. Additionally, for smaller Gaussians, the variation in the number of pixels involved across different viewpoints is minimal. The current averaging method does not produce a significant change compared to the original conditions and does not result in excessive additional memory consumption. The modified equation to decide whether a Gaussian undergoes split or clone is given by:
\begin{equation}\label{split2}
    \frac{\sum_{k=1}^{M^i}{m_{k}^{i}\times \sqrt{\left( \frac{\partial L_k}{\partial \mu _\mathrm{ndc,x}^{i,k}} \right) ^2+\left( \frac{\partial L_k}{\partial \mu _\mathrm{ndc,y}^{i,k}} \right) ^2}}}{\sum_{k=1}^{M^i}{m_{k}^{i}}}>\tau _\mathrm{pos},
\end{equation}
where $M^i$ is the number of viewpoints in which Gaussian $i$ participates in the computation during the corresponding 100 iterations of ``Adaptive Density Control'', $m_{k}^{i}$ is the number of pixels Gaussian $i$ participates in at viewpoint $k$, and $\frac{\partial L_k}{\partial \mu _\mathrm{ndc,x}^{i,k}}$ and $\frac{\partial L_k}{\partial \mu _\mathrm{ndc,y}^{i,k}}$ respectively represent the gradients of Gaussian $i$ in the $x$ and $y$ directions of NDC space at viewpoint $k$. The conditions under which a Gaussian participates in the computation for a pixel is given by:
\begin{equation}\label{pixel}
    \begin{cases}
        \sqrt{\left( pix_x-\mu _{p,x}^{i,k} \right) ^2+\left( pix_y-\mu _{p,y}^{i,k} \right) ^2}<R_{k}^{i}, \\
        \prod_{j=1}^i{\left( 1-\alpha _{k,pix}^{j} \right)}\geqslant 10^{-4},                               \\
        \alpha _{k,pix}^{i}\geqslant \frac{1}{255},
    \end{cases}
\end{equation}
while the conditions under which a Gaussian participates in the computation from a viewpoint is given by
Eq.~\ref{part_cal}.

\subsection{Scaled Gradient Field}

While using ``Pixel-aware Gradient'' to decide whether a point should split or clone (Eq.~\ref{split2}) can address artifacts in modeling areas with insufficient viewpoints and repetitive texture, we found that this condition for point cloud growth also exacerbates the presence of floaters near the camera. This is mainly because floaters near the camera occupy a large screen space and have significant gradients in their NDC coordinates, leading to an increasing number of floaters during the point cloud growth process. To address this issue, we scale the gradient field of the NDC coordinates.

Specifically, we use the radius to determine the scale of the scene, where the radius is calculated by:
\begin{equation}\label{radius}
    \mathrm{radius}=1.1\cdot \max_i \left\{ \left\| \mathbf{C}_i-\frac{1}{N}\sum_{j=1}^N{\mathbf{C}_j} \right\| _2 \right\}.
\end{equation}
In the training set, there are $N$ viewpoints, with $\mathbf{C}_j$  representing the coordinates of the $j$th viewpoint's camera in the world coordinate system. We scale the gradient of the NDC coordinates for each Gaussian $i$ under the $k$th viewpoint, with the scaling factor $f\left( i,k \right)$ being calculated by:
\begin{equation}\label{scale}
    f\left( i,k \right) =\mathrm{clip}\left( \left( \frac{\mu _{c,z}^{i,k}}{\gamma _\mathrm{depth}\times \mathrm{radius}} \right) ^2,0,1 \right),
\end{equation}
where $\mu _{c,z}^{i,k}$ is the z-coordinate of Gaussian $i$ in the camera coordinate system under the $k$th viewpoint, indicating the depth of this Gaussian from the viewpoint, and $\gamma _{\mathrm{depth}}$ is a hyperparameter set manually.

The primary inspiration for using squared terms as scaling coefficients in Eq.~\ref{scale} comes from ``Floaters No More''~\cite{philip2023floaters}. This paper notes that floaters in NeRF~\cite{mildenhall2021NeRF} are mainly due to regions close to the camera occupying more pixels after projection, which leads to receiving more gradients during optimization. This results in these areas being optimized first, consequently obscuring the originally correct spatial positions from being optimized. The number of pixels occupied is inversely proportional to the square of the distance to the camera, hence the scaling of gradients by the squared distance.

In summary, a major issue with pixel-based optimization is the imbalance in the spatial gradient field, leading to inconsistent optimization speeds across different areas. Adaptive scaling of the gradient field in different spatial regions can effectively address this problem.
Therefore, the final calculation equation that determines whether a Gaussian undergoes a ``split'' or ``clone'' is given by:
\begin{equation}\label{split3}
    \frac{\sum_{k=1}^{M^i}{m_{k}^{i}\times f\left( i,k \right) \times \sqrt{\left( \frac{\partial L_k}{\partial \mu _\mathrm{ndc,x}^{i,k}} \right) ^2+\left( \frac{\partial L_k}{\partial \mu _\mathrm{ndc,y}^{i,k}} \right) ^2}}}{\sum_{k=1}^{M^i}{m_{k}^{i}}}>\tau _\mathrm{pos}.
\end{equation}

%% file: sections/table_and_figure/method_pic.tex
\definecolor{mycolor}{RGB}{0, 0, 0} %

\begin{figure}[t]
\centering
\includegraphics[width=0.9\textwidth]{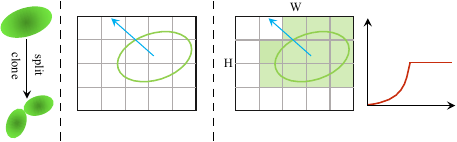}
\put(-243,7){\color{mycolor}$\frac{\sum{\parallel \mathbf{g}_i\parallel}}{\sum{1}} > \textcolor{mycolor}{\tau}_{\mathrm{pos}}$} 
\put(-120,7){\color{mycolor}$\frac{\sum{\mathrm{p}_i}\cdot f\left( \parallel \mathbf{g}_i\parallel \right)}{\sum{\mathrm{p}_i}}> \textcolor{mycolor}{\tau}_{\mathrm{pos}}$} 
\put(-64.5,87){\color{mycolor}$f$} 
\put(-13.5,12){\color{mycolor}$\mathrm{depth}$} 
\put(-237,90){\color{mycolor}$\mathbf{g}_i$} 
\put(-130,90){\color{mycolor}$\mathbf{g}_i$} 

\caption{\textbf{Pipeline of Pixel-GS.} $\mathrm{p}_i$ represents the number of pixels participating in the calculation for the Gaussian from this viewpoint, and $\mathbf{g}_i$ represents the gradient of the Gaussian's NDC coordinates. We changed the condition for deciding whether a Gaussian should split or clone from the left to the right side.}
\label{fig:method}
\vspace{-15pt}
\end{figure}

%% file: sections/experiments.tex
\subsection{Experimental Setup}

\noindent\textbf{Datasets and benchmarks.}
We evaluated our method across a total of 30 real-world scenes, including all scenes from Mip-NeRF 360 (9 scenes)~\cite{barron2022mip} and Tanks~\&~Temples (21 scenes)~\cite{knapitsch2017tanks}, which are two most widely used datasets in the field of 3D reconstruction.
They contain both bounded indoor scenes and unbounded outdoor scenes, allowing for a comprehensive evaluation of our method's performance.

\noindent\textbf{Evaluation metrics.}
We assess the quality of reconstruction through PSNR$\uparrow$, SSIM$\uparrow$~\cite{wang2004image}, and LPIPS$\downarrow$~\cite{zhang2018unreasonable}.
Among them, PSNR reflects pixel-aware errors but does not quite correspond to human visual perception as it treats all errors as noise without distinguishing between structural and non-structural distortions.
SSIM accounts for structural transformations in luminance, contrast, and structure, thus more closely mirroring human perception of image quality.
LPIPS uses a pre-trained deep neural network to extract features and measures the high-level semantic differences between images, offering a similarity that is closer to human perceptual assessment compared to PSNR and SSIM.

\noindent\textbf{Implementation details.}
Our method only requires minor modifications to the original code of 3DGS, so it is compatible with almost all subsequent works on 3DGS.
We use the default parameters of 3DGS to ensure consistency with the original implementation, including maintaining the same threshold $\tau_{pos}$ for splitting and cloning points as in the original 3DGS.
For all scenes, we set a constant $\gamma _\mathrm{depth}$ value in Eq.~\ref{scale} as 0.37 which is obtained through experimentations. All experiments were conducted on one RTX 3090 GPU with 24GB memory.

\subsection{Main Results}
We select several representative methods for comparison, including the NeRF methods, \eg, Plenoxels~\cite{fridovich2022plenoxels}, INGP~\cite{muller2022instant}, and Mip-NeRF 360~\cite{barron2022mip}, and the 3DGS method~\cite{kerbl20233d}.
We used the official implementation for all of the compared methods, and the same training/testing split as Mip-NeRF 360, selecting one out of every eight photos for testing.

\noindent\textbf{Quantitative results.}
The quantitative results (PSNR, SSIM, and LPIPS) on the Mip-NeRF 360 and Tanks~\&~Temples datasets are presented in Tables~\ref{table:result1} and~\ref{table:result2}, respectively.
We also provide the results of three challenging scenes for each dataset for more detailed information.
Here, we retrained the 3DGS (noted as 3DGS$^*$) as doing so yields a better performance than the original 3DGS (noted as 3DGS).
We can see that our method consistently outperforms all the other methods, especially in terms of the LPIPS metric, while maintaining real-time rendering speed (to be discussed later).
Besides, compared to 3DGS, our method shows significant improvements in the three challenging scenes in both datasets and achieves better performance over the entire dataset.
It quantitatively validates the effectiveness of our method in improving the quality of reconstruction.

\input{sections/table_and_figure/comparison_table_360.tex}
\input{sections/table_and_figure/comparison_table_tank}

\noindent\textbf{Qualitative results.}
In Figures~\ref{fig:results1} and~\ref{fig:results2}, we showcase the comparisons between our method and 3DGS$^*$.
We can see our approach significantly reduces the blurring and needle-like artifacts, \eg the region of the flowers in the second row and the blow-up region in the last row, compared against the 3DGS$^*$.
These regions are initialized with insufficient points from SfM, and our method effectively grows points in these areas, leading to a more accurate and detailed reconstruction.
Please refer to the supplemental materials for the point cloud comparison.
These examples clearly validate that our method is more robust to the quality of initialization point clouds and can reconstruct high-fidelity details.
\subsection{Ablation Studies}
To evaluate the effectiveness of individual components of our method, \ie the pixel-aware gradient and the scaled gradient field, we conducted ablation studies on the Mip-NeRF 360 and Tanks~\&~Temples datasets.
The quantitative and qualitative results are presented in Table~\ref{table:ablation1} and Figure~\ref{fig:abltion}, respectively.
We can see that both the pixel-aware gradient and the scaled gradient field contribute to the improvement of the reconstruction quality in the Mip-NeRF 360 dataset.
However, the pixel-aware gradient strategy reduces the reconstruction quality in the Tanks~\&~Temples dataset.
This is mainly due to floaters that tend to appear near the camera in some large scenes in Tanks~\&~Temples and the pixel-aware gradient encourages more Gaussians, as shown in column (b) of Figure~\ref{fig:abltion}.
Notably, this phenomenon also exists for the 3DGS when the threshold $\tau _\mathrm{pos}$ is lowered, which also promots more Gaussians, as shown in Table~\ref{tab:efficiency}.
But importantly, the combination of both proposed strategies achieves the best performance in the Tanks \& Temples dataset, as shown in Table~\ref{table:ablation1}, since the scaled gradient field can suppress the growth of floaters near the camera.
In summary, the ablation studies demonstrate the effectiveness of our proposed individual components and the necessity of combining them to achieve the best performance.

\input{sections/table_and_figure/qualitative2}

\input{sections/table_and_figure/ablation_qua}
\input{sections/table_and_figure/ablation_table}
\input{sections/table_and_figure/tau_table_all}
\input{sections/table_and_figure/drop_figure}

\subsection{Analysis}
\noindent\textbf{The impact of lowering the threshold $\tau _\mathrm{pos}$.}
As the blurring and needle-like artifacts in 3DGS mainly occur in areas with insufficient initializing points, one straightforward solution would be to lower the threshold $\tau _\mathrm{pos}$ to encourage the growth of more points.
To verify this, we experimented on the Mip-NeRF 360 and Tanks~\&~Temples datasets by lowering the threshold $\tau _\mathrm{pos}$ from $2e-4$ to $1.28e-4$ for 3DGS to make the final optimized number of points comparable to ours.
From Table~\ref{tab:efficiency}, we can see that lowering the threshold $\tau _\mathrm{pos}$ for 3DGS significantly increases the memory consumption and decreases the rendering speed, while still falling behind ours in terms of reconstruction quality.
As can be seen from the qualitative comparison in Figure~\ref{fig:results1}, this is because the point cloud growth mechanism of 3DGS struggles to generate points in areas with insufficient initializing points and only yields unnecessary points in areas where the initial SfM point cloud is already dense.
In contrast, although our method also results in additional memory consumption, our method's point cloud distribution is more uniform, enabling effectively growing points in areas with insufficient initializing points, thereby leading to a more accurate and detailed reconstruction while still maintaining real-time rendering speed.

\noindent\textbf{Robustness to the quality of initialization point clouds.}
Finally, SfM algorithms often fail to produce high-quality point clouds in some areas, \eg, too few observations, repetitive textures, or low textures.
The point cloud produced by SfM is usually the necessary input for 3DGS and our method.
Therefore, we explored the robustness of our method to the quality of initialization point clouds by randomly dropping points from the SfM point clouds used for initialization and compared the results with that of 3DGS.
Figure~\ref{fig:drop} shows how the reconstruction quality varies with the proportion of dropped points.
We can see that our method consistently outperforms 3DGS in terms of all the metrics (PSNR, SSIM, and LPIPS).
And more importantly, our method is less affected by the dropping rate than 3DGS.
Notably, even though the $99\%$ initializing points have been dropped, the reconstruction quality of our method still surpasses that of 3DGS initialized with complete SfM point clouds, in terms of LPIPS.
These results demonstrate the robustness of our method to the quality of initialization point clouds, which is crucial for real-world applications.

%% file: sections/table_and_figure/comparison_table_360.tex
\begin{table}[t]
	\centering
	\small
	\caption{\textbf{Quantitative results on the Mip-NeRF 360 dataset.} Cells are highlighted as follows: \colorbox{red!40}{best}, \colorbox{orange!40}{second best}, and \colorbox{yellow!40}{third best}. We also show the results of three challenging scenes. 3DGS{$^*$} is our retrained 3DGS model with better performance.}
	\label{table:result1}
	\scalebox{0.71}{
		\begin{tabular}{l|>{\centering\arraybackslash}p{1.2cm}>{\centering\arraybackslash}p{1.2cm}>{\centering\arraybackslash}p{1.2cm}|ccc|ccc|ccc}

			                                        & \multicolumn{3}{c|}{Mip-NeRF 360 (all scenes)} & \multicolumn{3}{c|}{Flowers} & \multicolumn{3}{c|}{Bicycle} & \multicolumn{3}{c}{Stump}                                                                                                                                                                                                                                          \\
			Method
			                                        & PSNR$\uparrow$                                 & SSIM$\uparrow$               & LPIPS$\downarrow$
			                                        & PSNR$\uparrow$                                 & SSIM$\uparrow$               & LPIPS$\downarrow$
			                                        & PSNR$\uparrow$                                 & SSIM$\uparrow$               & LPIPS$\downarrow$
			                                        & PSNR$\uparrow$                                 & SSIM$\uparrow$               & LPIPS$\downarrow$                                                                                                                                                                                                                                                                                 \\
			\hline
			Plenoxels~\cite{fridovich2022plenoxels} & 23.08                                          & 0.625                        & 0.463                        & 20.10                      & 0.431                      & 0.521                      & 21.91                      & 0.496                      & 0.506                      & 20.66                      & 0.523                      & 0.503                      \\
			INGP-Base~\cite{muller2022instant}      & 25.30                                          & 0.671                        & 0.371                        & 20.35                      & 0.450                      & 0.481                      & 22.19                      & 0.491                      & 0.487                      & 23.63                      & 0.574                      & 0.450                      \\
			INGP-Big~\cite{muller2022instant}       & 25.59                                          & 0.699                        & 0.331                        & 20.65                      & 0.486                      & 0.441                      & 22.17                      & 0.512                      & 0.446                      & 23.47                      & 0.594                      & 0.421                      \\
			Mip-NeRF 360~\cite{barron2022mip}       & \cellcolor{yellow!40}27.69                     & 0.792                        & 0.237                        & \cellcolor{yellow!40}21.73 & 0.583                      & 0.344                      & 24.37                      & 0.685                      & 0.301                      & 26.40                      & 0.744                      & 0.261                      \\
			\hline
			3DGS~\cite{kerbl20233d}                 & 27.21                                          & \cellcolor{yellow!40}0.815   & \cellcolor{yellow!40}0.214   & 21.52                      & \cellcolor{yellow!40}0.605 & \cellcolor{yellow!40}0.336 & \cellcolor{yellow!40}25.25 & \cellcolor{yellow!40}0.771 & \cellcolor{yellow!40}0.205 & \cellcolor{yellow!40}26.55 & \cellcolor{yellow!40}0.775 & \cellcolor{yellow!40}0.210 \\
			3DGS$^*$~\cite{kerbl20233d}             & \cellcolor{orange!40}27.71                     & \cellcolor{orange!40}0.826   & \cellcolor{orange!40}0.202   & \cellcolor{orange!40}21.89 & \cellcolor{orange!40}0.622 & \cellcolor{orange!40}0.328 & \cellcolor{orange!40}25.63 & \cellcolor{orange!40}0.778 & \cellcolor{orange!40}0.204 & \cellcolor{orange!40}26.90 & \cellcolor{orange!40}0.785 & \cellcolor{orange!40}0.207 \\
			Pixel-GS (Ours)                       & \cellcolor{red!40}27.88                        & \cellcolor{red!40}0.834      & \cellcolor{red!40}0.176      & \cellcolor{red!40}21.94    & \cellcolor{red!40}0.652    & \cellcolor{red!40}0.251    & \cellcolor{red!40}25.74    & \cellcolor{red!40}0.793    & \cellcolor{red!40}0.173    & \cellcolor{red!40}27.11    & \cellcolor{red!40}0.796    & \cellcolor{red!40}0.181
		\end{tabular}
	}
\end{table}

%% file: sections/table_and_figure/comparison_table_tank.tex
\begin{table}[t]
	\centering
	\small
	\caption{\textbf{Quantitative results on the Tanks~\&~Temples dataset.} We also show the results of three challenging scenes. $^*$ indicates retraining for better performance.}
	\label{table:result2}
	\scalebox{0.71}{
		\begin{tabular}{l|>{\centering\arraybackslash}p{1.32cm}>{\centering\arraybackslash}p{1.32cm}>{\centering\arraybackslash}p{1.32cm}|ccc|ccc|ccc}

			                            & \multicolumn{3}{c|}{Tanks~\&~Temples (all scenes)} & \multicolumn{3}{c|}{Train} & \multicolumn{3}{c|}{Barn}  & \multicolumn{3}{c}{Caterpillar}                                                                                                                                                                                                                                         \\
			Method
			                            & PSNR$\uparrow$                                     & SSIM$\uparrow$             & LPIPS$\downarrow$
			                            & PSNR$\uparrow$                                     & SSIM$\uparrow$             & LPIPS$\downarrow$
			                            & PSNR$\uparrow$                                     & SSIM$\uparrow$             & LPIPS$\downarrow$
			                            & PSNR$\uparrow$                                     & SSIM$\uparrow$             & LPIPS$\downarrow$                                                                                                                                                                                                                                                                                    \\
			\hline

			3DGS$^*$~\cite{kerbl20233d} & \cellcolor{orange!40}24.19                         & \cellcolor{orange!40}0.844 & \cellcolor{orange!40}0.194 & \cellcolor{orange!40}22.02      & \cellcolor{orange!40}0.812 & \cellcolor{orange!40}0.209 & \cellcolor{orange!40}28.46 & \cellcolor{orange!40}0.869 & \cellcolor{orange!40}0.182 & \cellcolor{orange!40}23.79 & \cellcolor{orange!40}0.809 & \cellcolor{orange!40}0.211 \\
			Pixel-GS (Ours)          & \cellcolor{red!40}24.38                            & \cellcolor{red!40}0.850    & \cellcolor{red!40}0.178    & \cellcolor{red!40}22.13         & \cellcolor{red!40}0.823    & \cellcolor{red!40}0.180    & \cellcolor{red!40}29.00    & \cellcolor{red!40}0.888    & \cellcolor{red!40}0.144    & \cellcolor{red!40}24.08    & \cellcolor{red!40}0.832    & \cellcolor{red!40}0.173
		\end{tabular}
	}
\end{table}

%% file: sections/table_and_figure/qualitative2.tex
\begin{figure}[h!]
    \centering
    \scriptsize
    \resizebox{0.915\linewidth}{!}{
        \begin{tabular}{@{\hspace{0.0mm}}c@{\hspace{0.5mm}}c@{\hspace{0.5mm}}c@{\hspace{0.0mm}}}
            \includegraphics[width=0.33\linewidth]{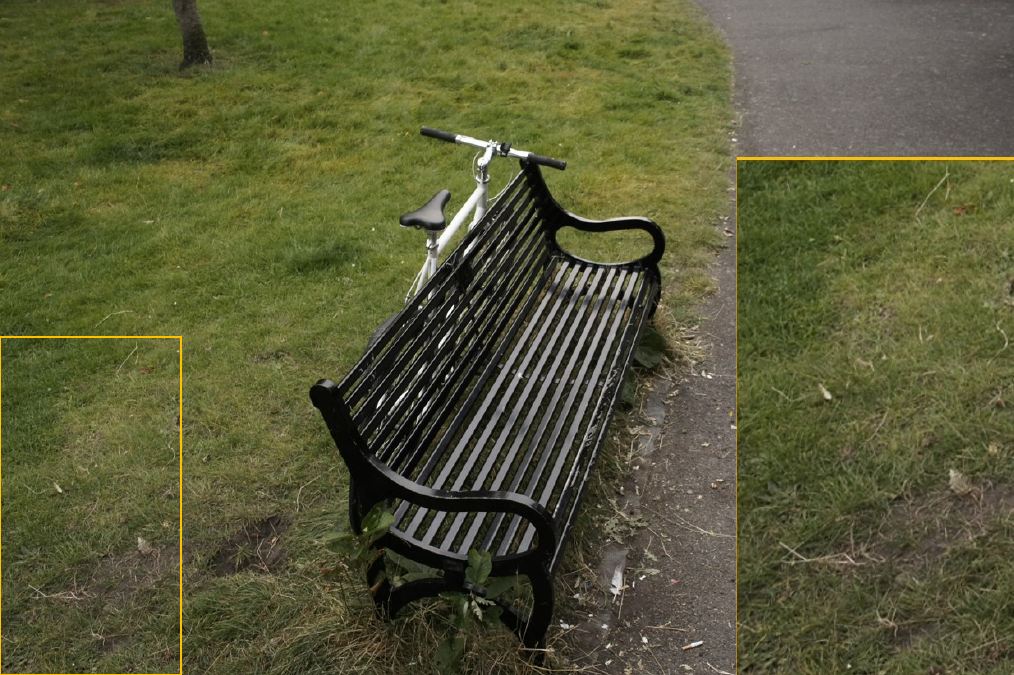}       &
            \includegraphics[width=0.33\linewidth]{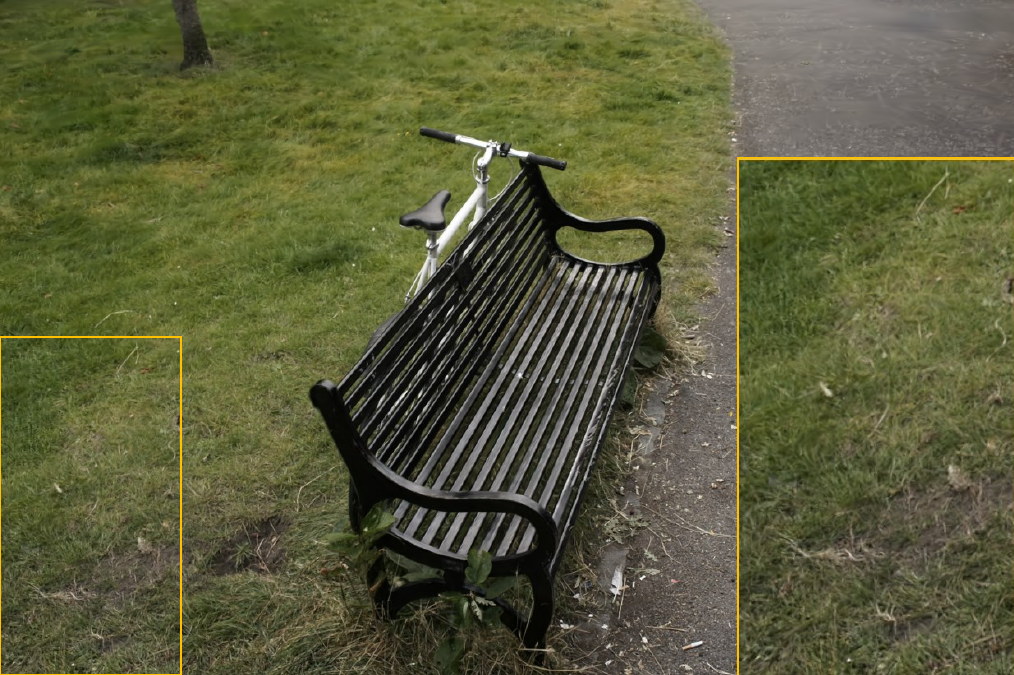}     &
            \includegraphics[width=0.33\linewidth]{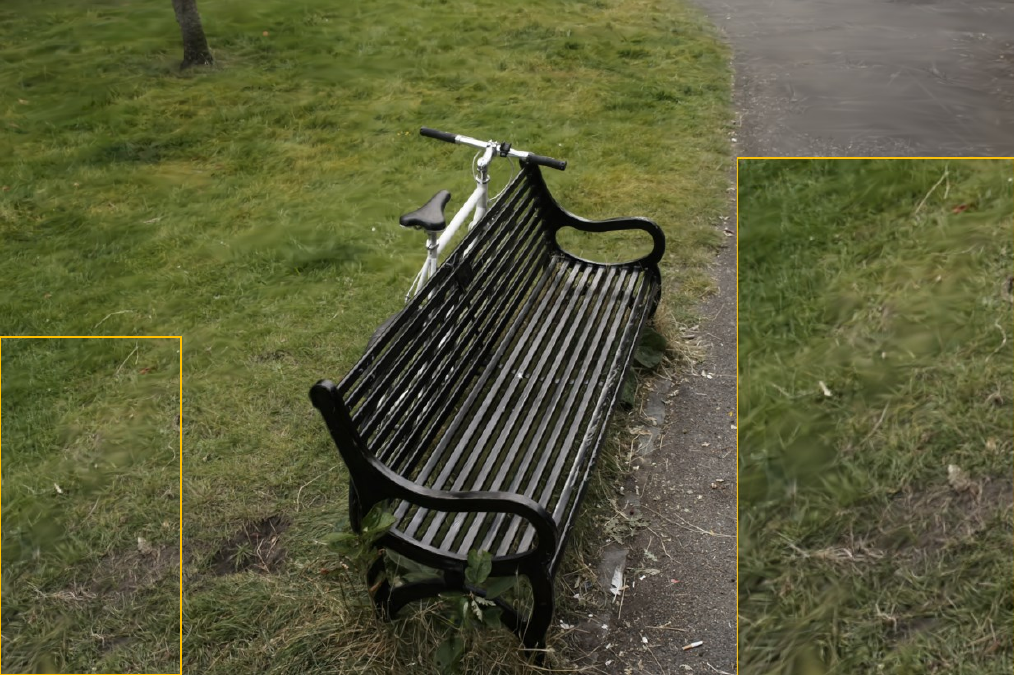}                                                    \\
            \includegraphics[width=0.33\linewidth]{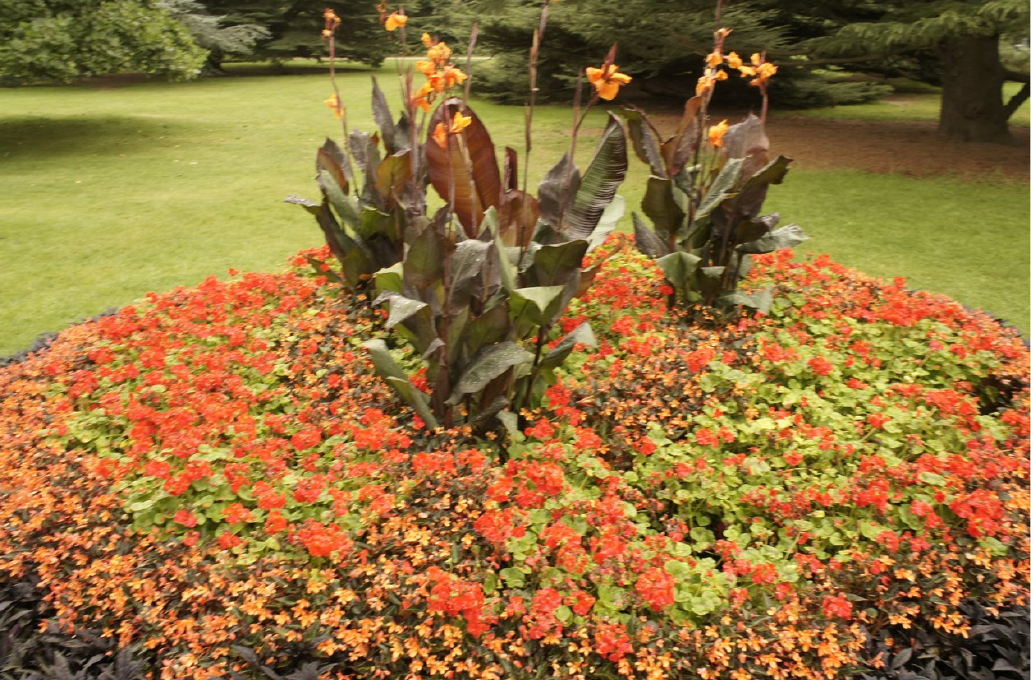}       &
            \includegraphics[width=0.33\linewidth]{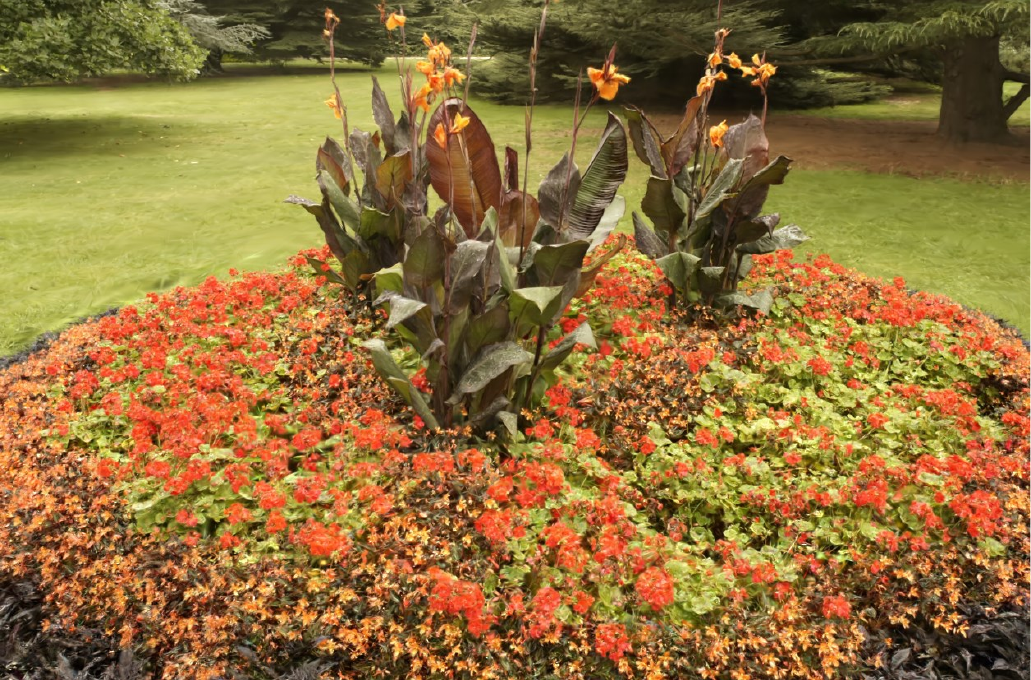}     &
            \includegraphics[width=0.33\linewidth]{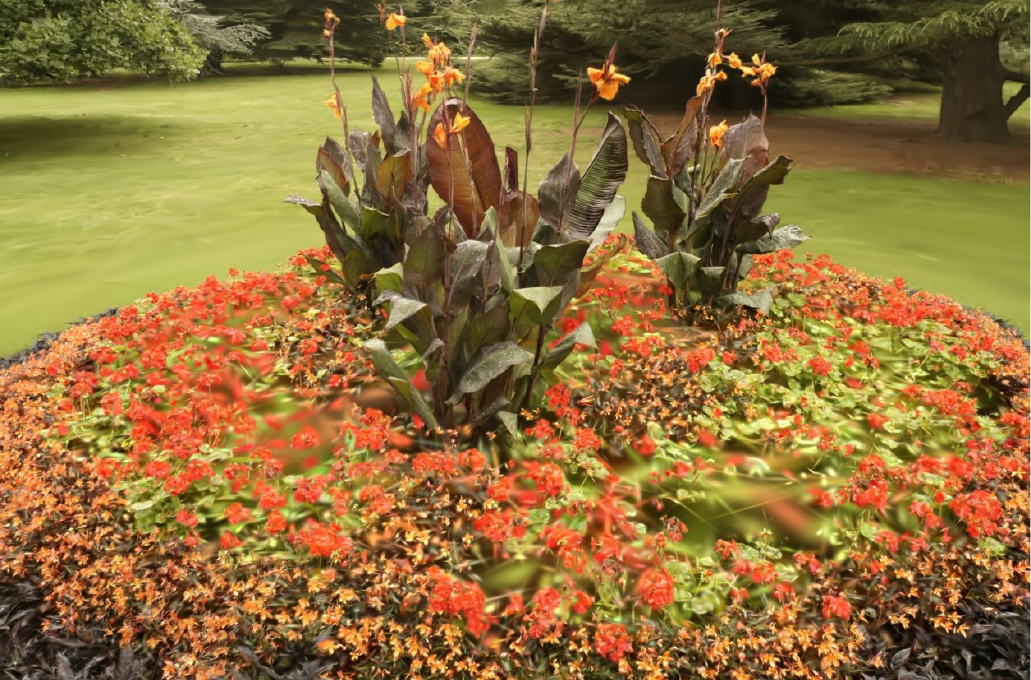}                                                    \\
            \includegraphics[width=0.33\linewidth]{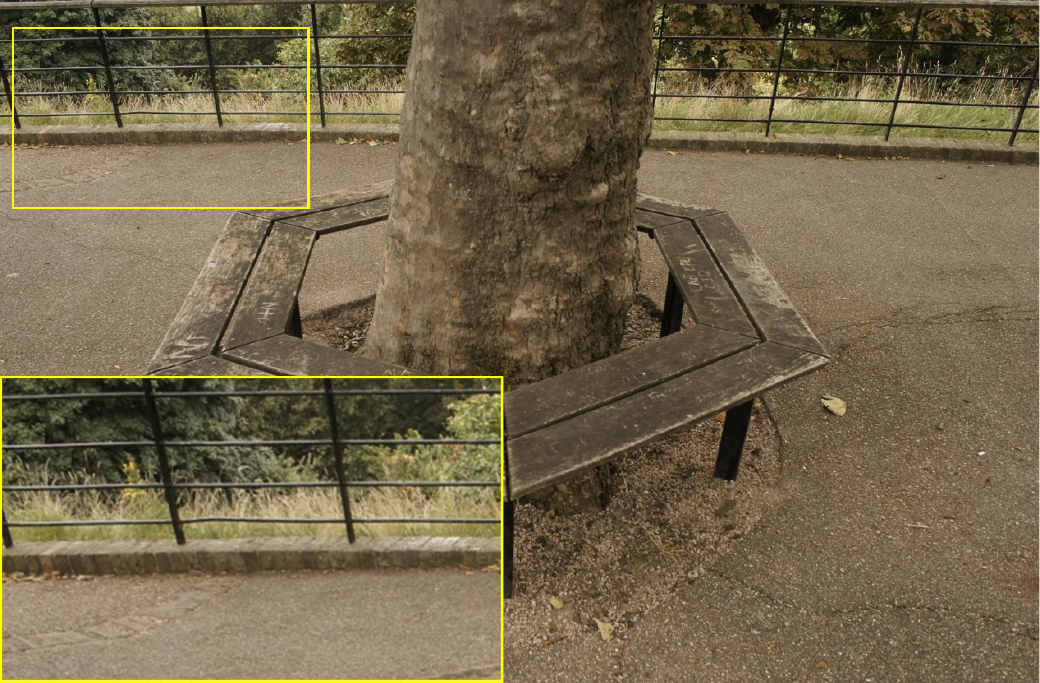}      &
            \includegraphics[width=0.33\linewidth]{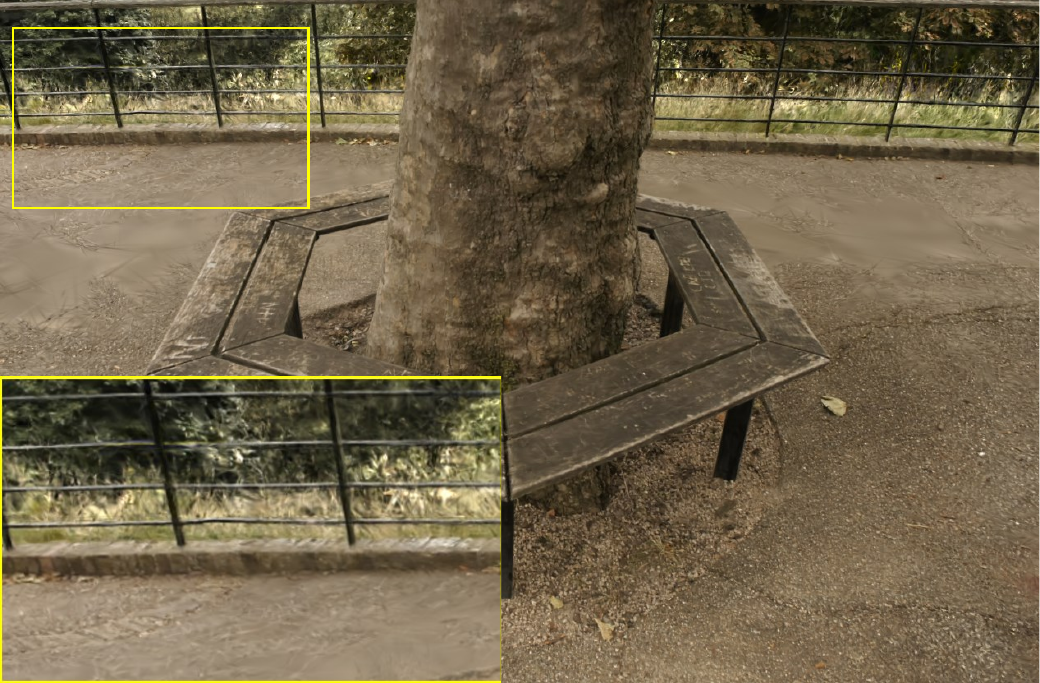}    &
            \includegraphics[width=0.33\linewidth]{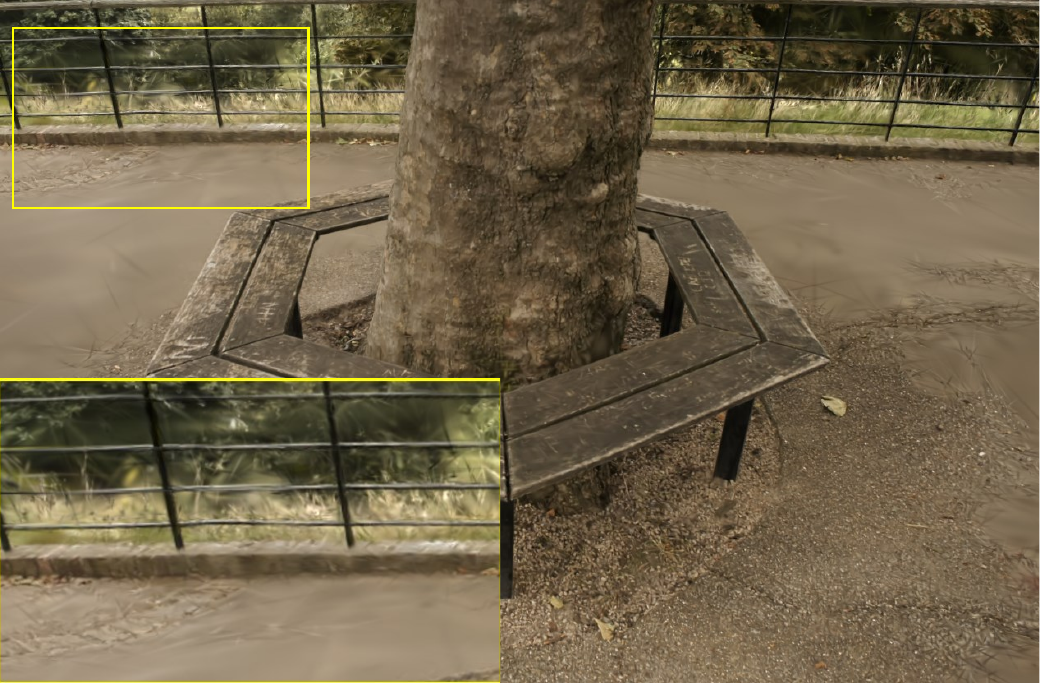}                                                   \\
            \includegraphics[width=0.33\linewidth]{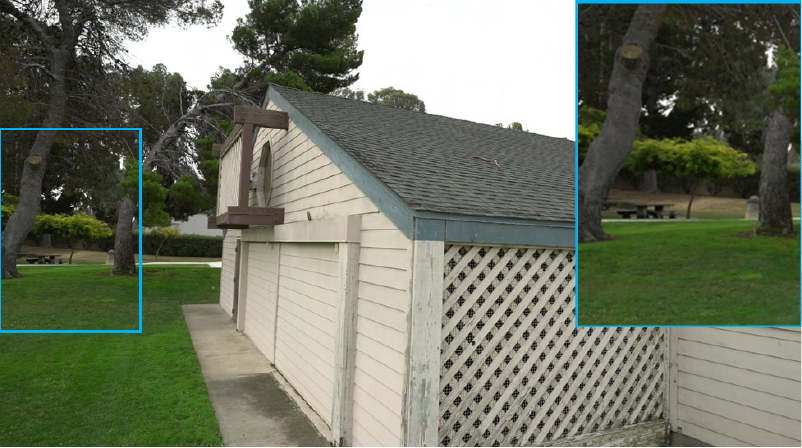}          &
            \includegraphics[width=0.33\linewidth]{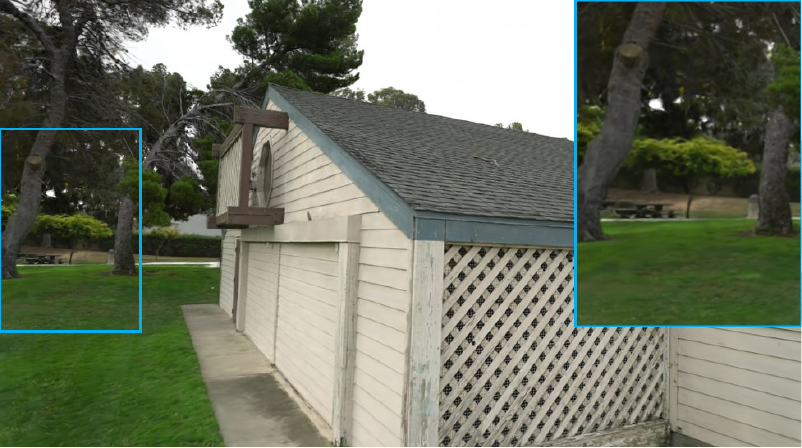}        &
            \includegraphics[width=0.33\linewidth]{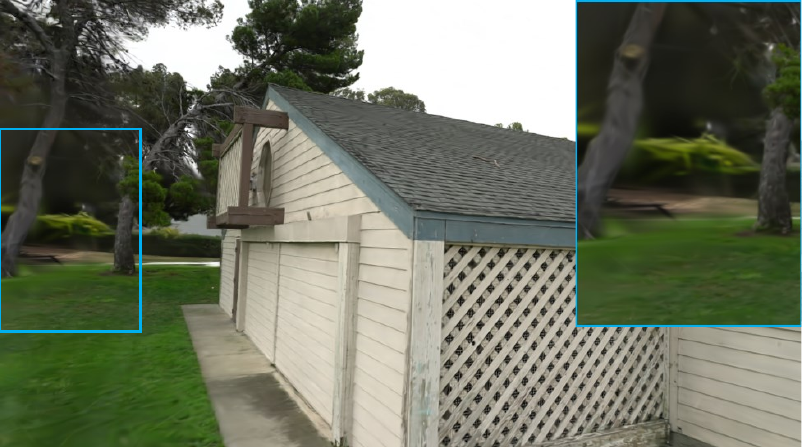}                                                       \\
            \includegraphics[width=0.33\linewidth]{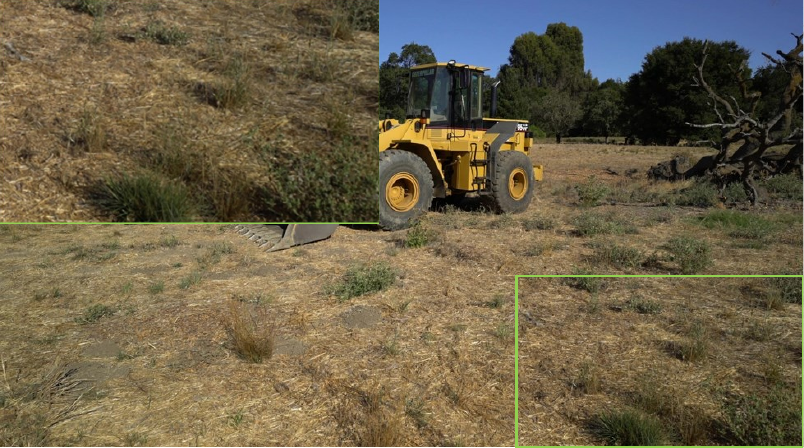}   &
            \includegraphics[width=0.33\linewidth]{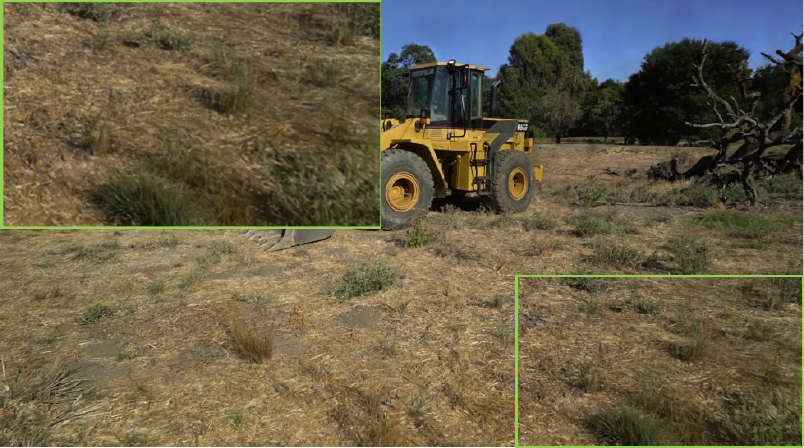} &
            \includegraphics[width=0.33\linewidth]{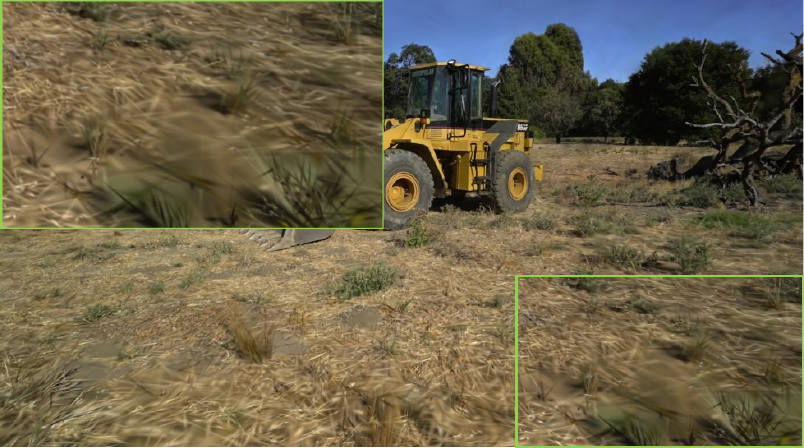}                                                \\
            \includegraphics[width=0.33\linewidth]{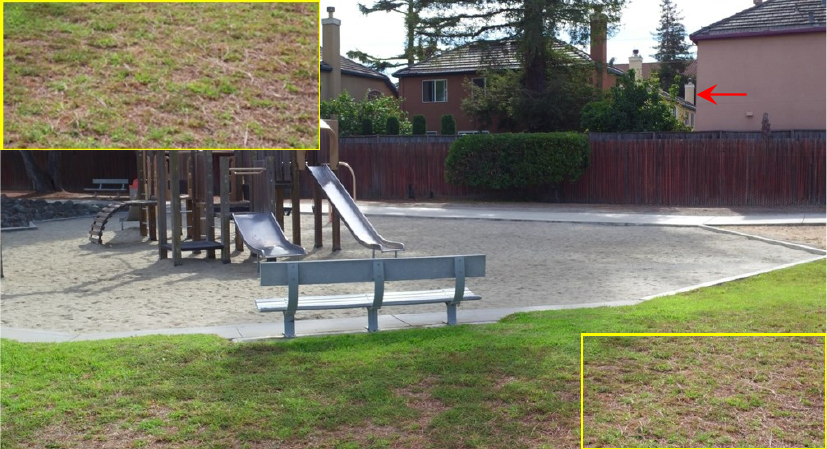}    &
            \includegraphics[width=0.33\linewidth]{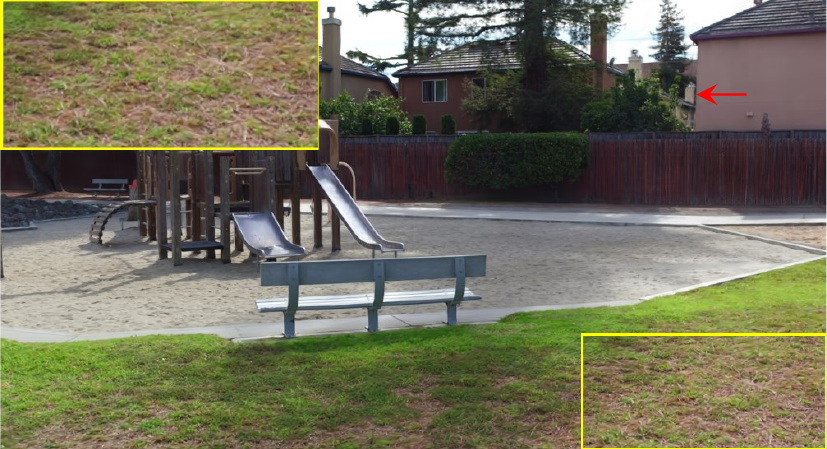}  &
            \includegraphics[width=0.33\linewidth]{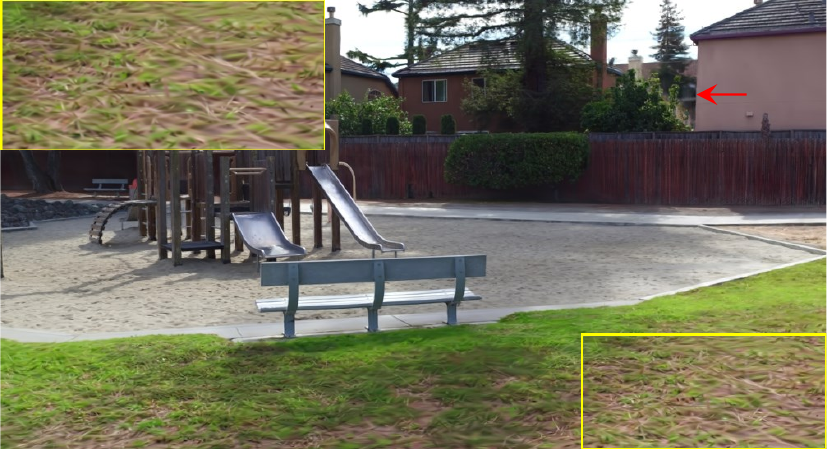}                                                 \\
            \includegraphics[width=0.33\linewidth]{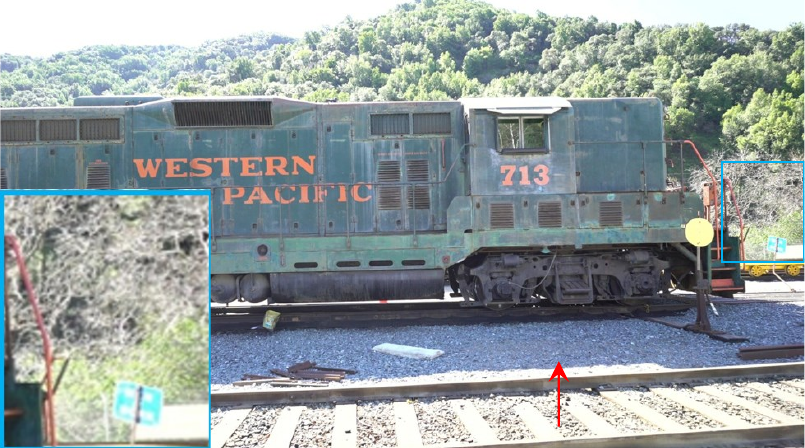}         &
            \includegraphics[width=0.33\linewidth]{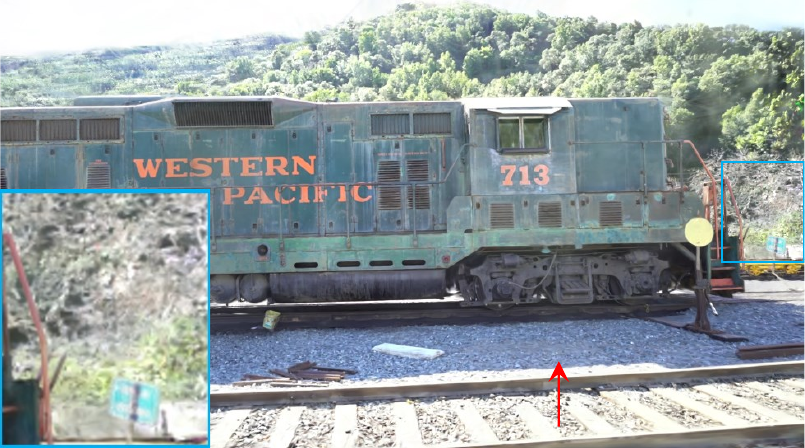}       &
            \includegraphics[width=0.33\linewidth]{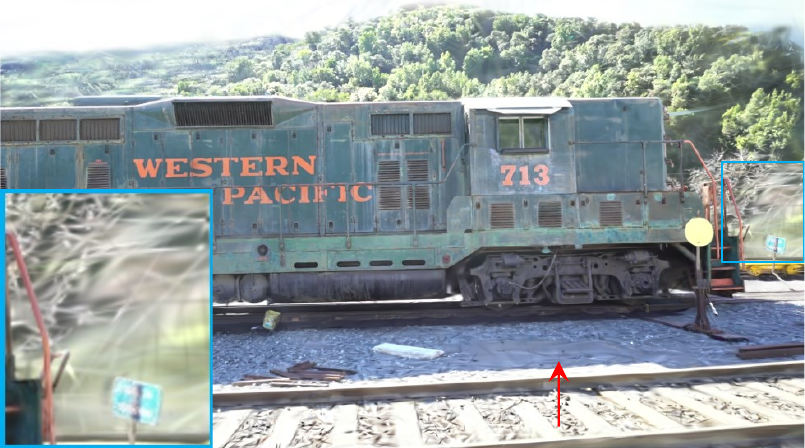}                                                      \\
            (a) Ground Truth                                                     & (b) Pixel-GS (Ours) & (c) 3DGS{$^*$}~\cite{kerbl20233d} \\
        \end{tabular}}
    \captionof{figure}{
        \textbf{Qualitative comparison between Pixel-GS (Ours) and 3DGS{$^*$}.} The first three scenes are from the Mip-NeRF 360 dataset (\textit{Bicycle, Flowers, and Treehill}), while the last four scenes are from the Tanks~\&~Temples dataset (\textit{Barn, Caterpillar, Playground, and Train}). The blow-up regions or arrows highlight the parts with distinct differences in quality. 3DGS{$^*$} is our retrained 3DGS model with better performance.
    }
    \label{fig:results2}
\end{figure}

%% file: sections/table_and_figure/ablation_qua.tex
\begin{figure}[!h]
    \centering
    \tiny
    \resizebox{1.0\linewidth}{!}{
        \begin{tabular}{@{\hspace{0.0mm}}c@{\hspace{0.5mm}}c@{\hspace{0.5mm}}c@{\hspace{0.5mm}}c@{\hspace{0.0mm}}}
            \includegraphics[width=0.25\linewidth]{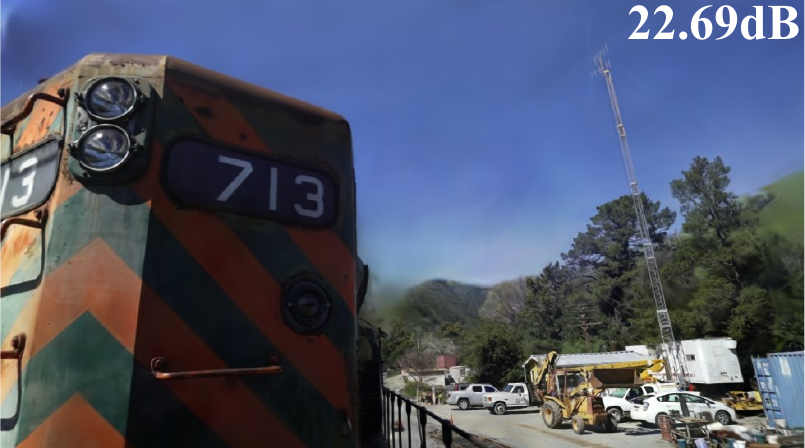}  &
            \includegraphics[width=0.25\linewidth]{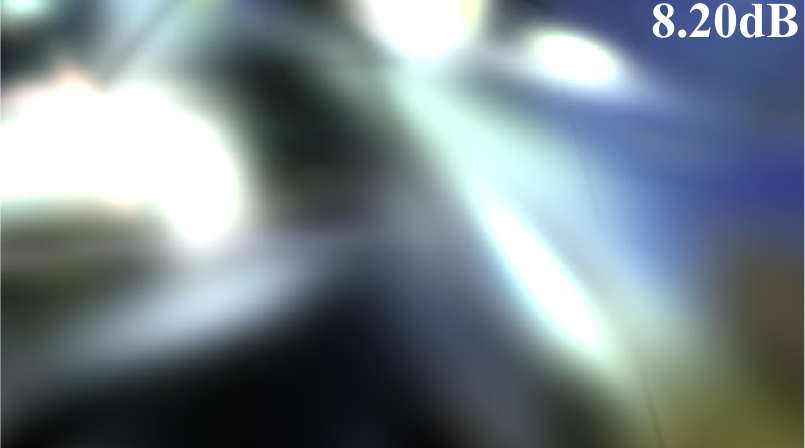}     &
            \includegraphics[width=0.25\linewidth]{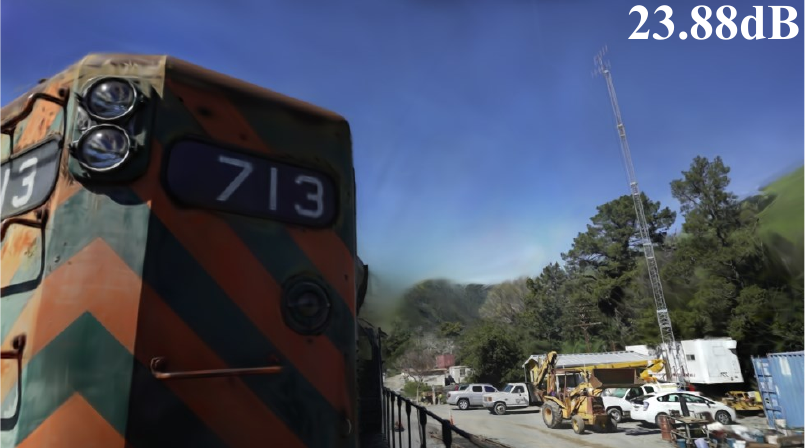}    &
            \includegraphics[width=0.25\linewidth]{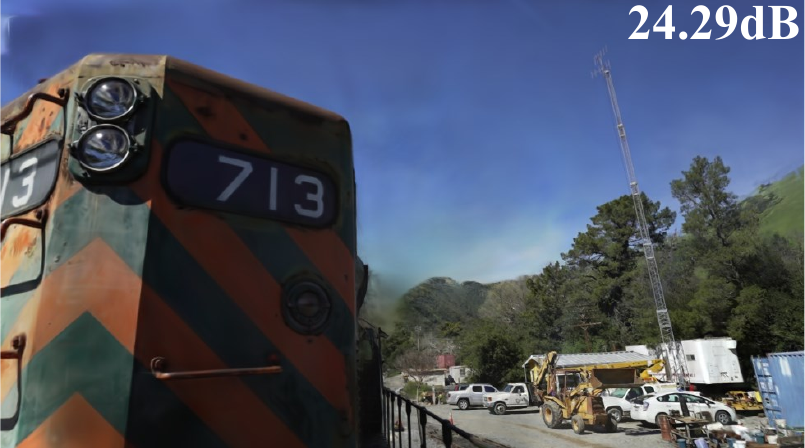}                                                                          \\
            \includegraphics[width=0.25\linewidth]{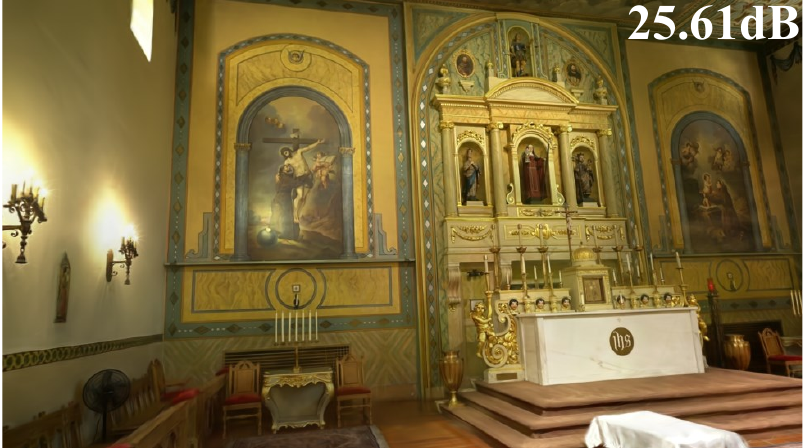} &
            \includegraphics[width=0.25\linewidth]{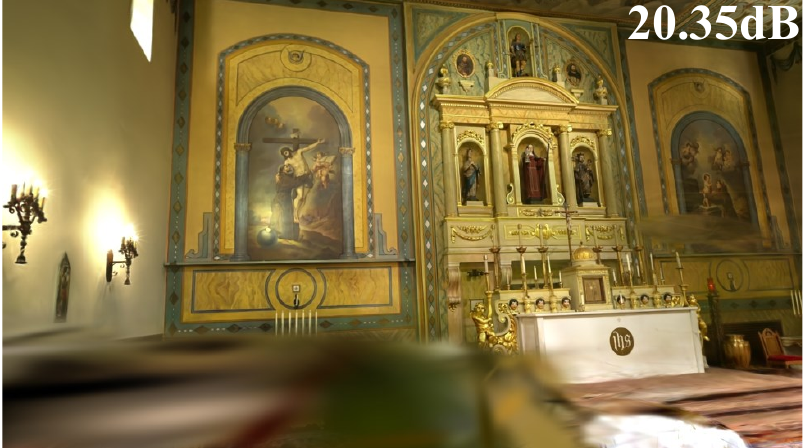}    &
            \includegraphics[width=0.25\linewidth]{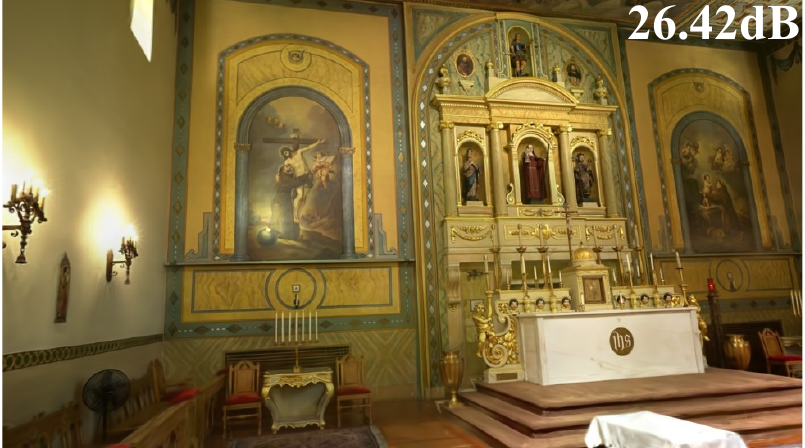}   &
            \includegraphics[width=0.25\linewidth]{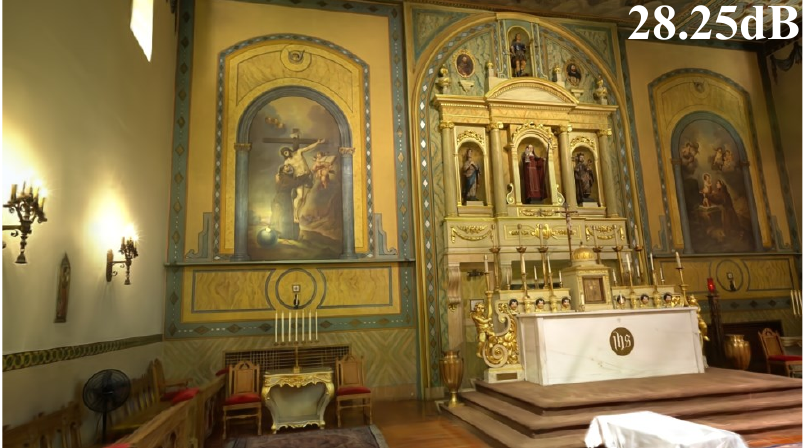}                                                                         \\
            \includegraphics[width=0.25\linewidth]{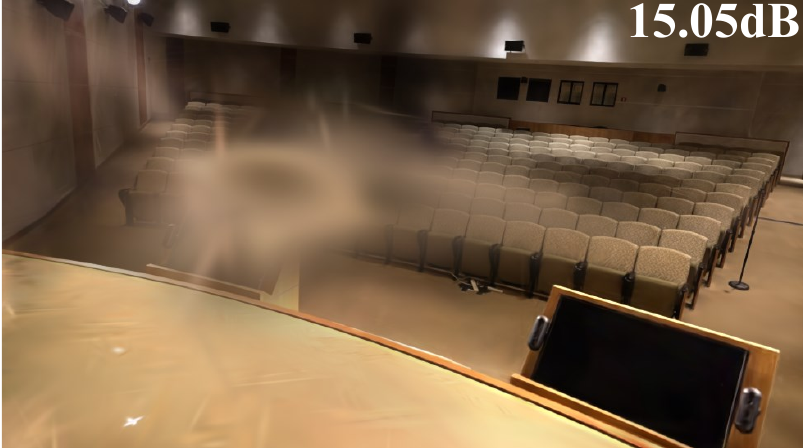} &
            \includegraphics[width=0.25\linewidth]{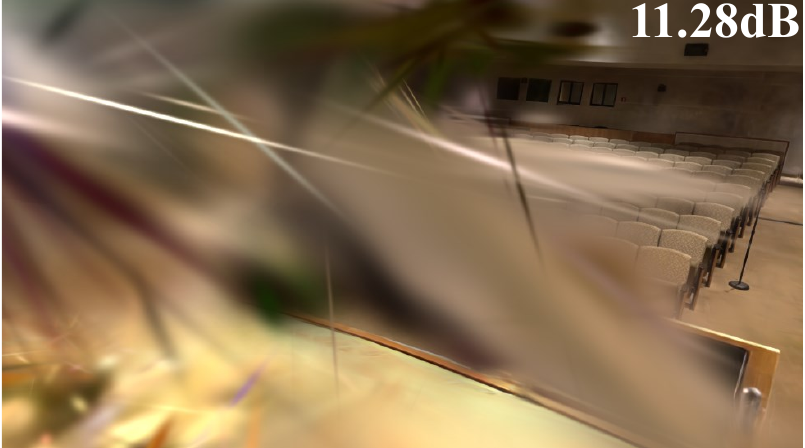}    &
            \includegraphics[width=0.25\linewidth]{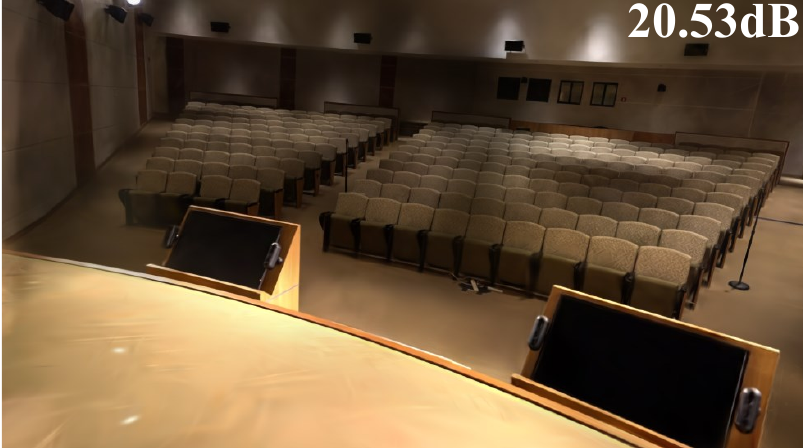}   &
            \includegraphics[width=0.25\linewidth]{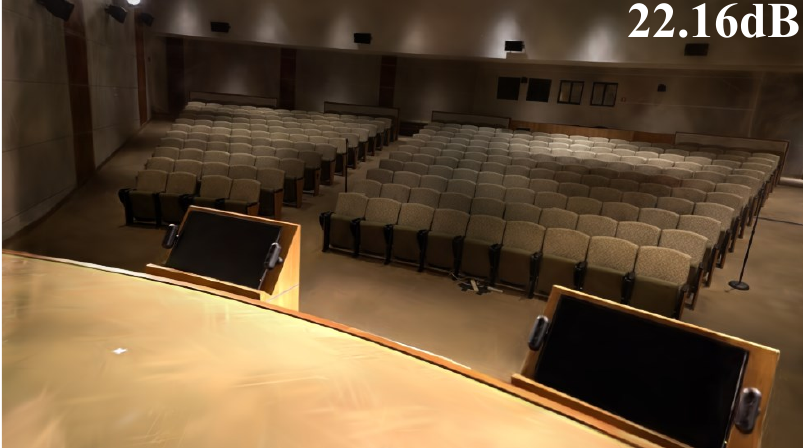}                                                                         \\
            \includegraphics[width=0.25\linewidth]{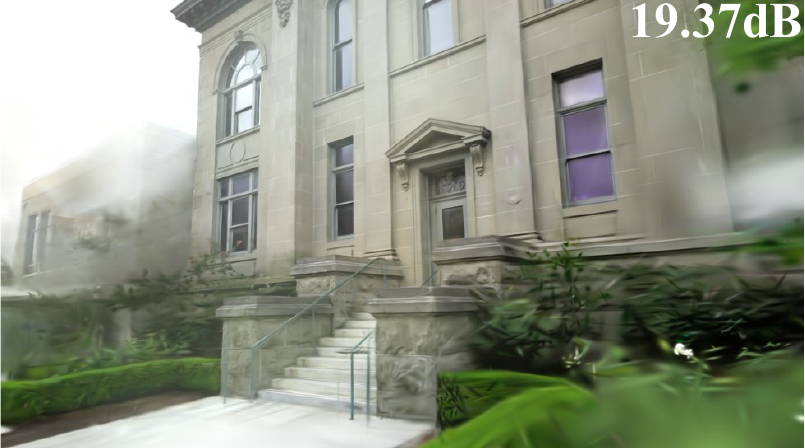} &
            \includegraphics[width=0.25\linewidth]{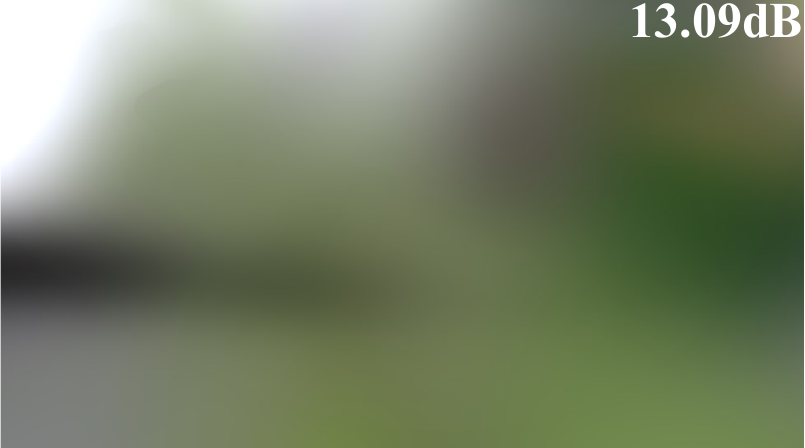}    &
            \includegraphics[width=0.25\linewidth]{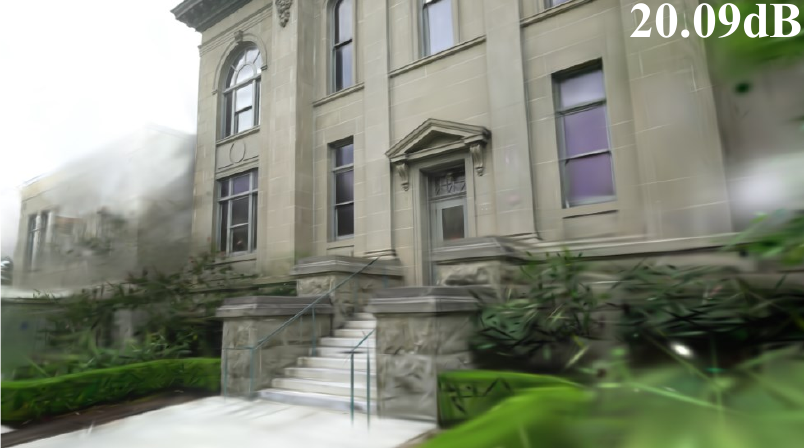}   &
            \includegraphics[width=0.25\linewidth]{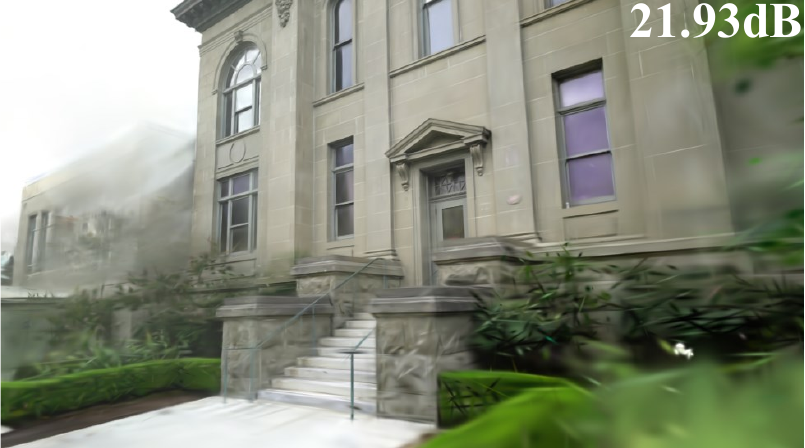}                                                                         \\
            (a) 3DGS$^*$                                                                     & (b) Pixel-aware Gradient & (c) Scaled Gradient Field & (d) Complete Model \\
        \end{tabular}}
    \captionof{figure}{\textbf{Qualitative results of the ablation study.} The PSNR$\uparrow$ results are shown on the corresponding images.}
    \label{fig:abltion}
\end{figure}

%% file: sections/table_and_figure/ablation_table.tex
\begin{table}[t]
	\centering
	\small
	\caption{\textbf{Ablation study.} The metrics are derived from the average values across all scenes of the Mip-NeRF 360 and Tanks~\&~Temples datasets, respectively.}
	\label{table:ablation1}
	\scalebox{0.9}{
		\begin{tabular}{l|ccc|ccc}

			~                     & \multicolumn{3}{c|}{Mip-NeRF 360} & \multicolumn{3}{c}{Tanks~\&~Temples}                                                                                                                     \\
			Method
			                            & PSNR$\uparrow$                    & SSIM$\uparrow$                       & LPIPS$\downarrow$
			                            & PSNR$\uparrow$                    & SSIM$\uparrow$                       & LPIPS$\downarrow$                                                                                                 \\
			\hline

			3DGS$^*$~\cite{kerbl20233d} & 27.71                             & \cellcolor{yellow!40}0.826           & \cellcolor{yellow!40}0.202 & \cellcolor{yellow!40}24.23 & \cellcolor{orange!40}0.844 & \cellcolor{orange!40}0.194 \\

			Pixel-aware Gradient        & \cellcolor{orange!40}27.74        & \cellcolor{orange!40}0.833           & \cellcolor{red!40}0.176    & 21.80                      & 0.791                      & 0.239                      \\

			Scaled Gradient Field       & \cellcolor{yellow!40}27.72        & 0.825                                & \cellcolor{yellow!40}0.202 & \cellcolor{orange!40}24.34 & \cellcolor{yellow!40}0.843 & \cellcolor{yellow!40}0.198 \\

			Complete Model              & \cellcolor{red!40}27.88           & \cellcolor{red!40}0.834              & \cellcolor{red!40}0.176    & \cellcolor{red!40}24.38    & \cellcolor{red!40}0.850    & \cellcolor{red!40}0.178
		\end{tabular}
	}
\end{table}

%% file: sections/table_and_figure/tau_table_all.tex
\begin{table}[!t]
    \centering
    \small
    \caption{\textbf{Impact of lowering $\tau _\mathrm{pos}$.} We show the corresponding quality and efficiency metrics when lowering the threshold $\tau _\mathrm{pos}$ of point growth for 3DGS$^*$ and our method.}
    \label{tab:efficiency}
    \scalebox{0.80}{
        \begin{tabular}{l|c|cccccc}
            Dataset & Strategy                                & PSNR$\uparrow$             & SSIM$\uparrow$             & LPIPS$\downarrow$          & Train                       & FPS                     & Memory                      \\ \hline
            \multirow{3}{*}{Mip-NeRF 360}
                    & 3DGS$^*$ $(\tau _\mathrm{pos}=2e-4)$    & \cellcolor{yellow!40}27.71 & \cellcolor{yellow!40}0.826 & \cellcolor{yellow!40}0.202 & \cellcolor{red!40}25m40s    & \cellcolor{red!40}126   & \cellcolor{red!40}0.72GB    \\
                    & 3DGS$^*$ $(\tau _\mathrm{pos}=1.28e-4)$ & \cellcolor{orange!40}27.83 & \cellcolor{orange!40}0.833 & \cellcolor{orange!40}0.181 & \cellcolor{yellow!40}43m23s & \cellcolor{orange!40}90 & \cellcolor{yellow!40}1.4GB  \\
                    & Ours $(\tau _\mathrm{pos}=2e-4)$        & \cellcolor{red!40}27.88    & \cellcolor{red!40}0.834    & \cellcolor{red!40}0.176    & \cellcolor{orange!40}41m25s & \cellcolor{yellow!40}89 & \cellcolor{orange!40}1.2GB  \\ \hline
            \multirow{3}{*}{Tanks~\&~Temples}
                    & 3DGS$^*$ $(\tau _\mathrm{pos}=2e-4)$    & \cellcolor{orange!40}24.19 & \cellcolor{orange!40}0.844 & \cellcolor{yellow!40}0.194 & \cellcolor{red!40}16m3s     & \cellcolor{red!40}135   & \cellcolor{red!40}0.41GB    \\
                    & 3DGS$^*$ $(\tau _\mathrm{pos}=1e-4)$    & \cellcolor{yellow!40}23.86 & \cellcolor{yellow!40}0.842 & \cellcolor{orange!40}0.187 & \cellcolor{yellow!40}27m59s & \cellcolor{yellow!40}87 & \cellcolor{yellow!40}0.94GB \\
                    & Ours $(\tau _\mathrm{pos}=2e-4)$        & \cellcolor{red!40}24.38    & \cellcolor{red!40}0.850    & \cellcolor{red!40}0.178    & \cellcolor{orange!40}26m36s & \cellcolor{orange!40}92 & \cellcolor{orange!40}0.84GB \\
        \end{tabular}
    }
    \vspace{-10pt}
\end{table}

%% file: sections/table_and_figure/drop_figure.tex
\begin{figure}[t]
  \centering
  \scriptsize
  \resizebox{1.0\linewidth}{!}{
    \begin{tabular}{@{\hspace{0.0mm}}c@{\hspace{0.0mm}}c@{\hspace{0.0mm}}c@{\hspace{0.0mm}}}
      \includegraphics[width=0.33\linewidth]{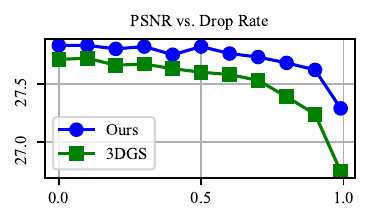} &
      \includegraphics[width=0.33\linewidth]{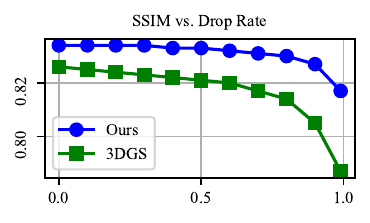} &
      \includegraphics[width=0.33\linewidth]{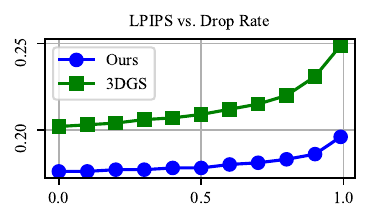}  \\
    \end{tabular}}
  \caption{
    \textbf{Reconstruction quality (PSNR$\uparrow$, SSIM$\uparrow$, and LPIPS$\downarrow$) \vs. Dropping rate of initializing points.} Here, the dropping rate refers to the percentage of points dropped from the original SfM point clouds for initializing Gaussians. The results are obtained on the Mip-NeRF 360 dataset.
  }
  \label{fig:drop}
\end{figure}

%% file: sections/conclusion.tex
The blurring and needle-like artifacts in 3DGS are mainly attributed to its inability to grow points in areas with insufficient initializing points.
To address this issue, we propose Pixel-GS, which considers the number of pixels covered by a Gaussian in each view to dynamically weigh the gradient of each view during the computation of the growth condition.
This strategy effectively grows Gaussians with large scales, which are more likely to exist in areas with insufficient initializing points, such that our method can adaptively grow points in these areas while avoiding unnecessary growth in areas with enough points.
We also introduce a simple yet effective strategy to deal with floaters, \ie, scaling the gradient field by the distance to the camera.
Extensive experiments demonstrate that our method significantly reduces blurring and needle-like artifacts and effectively suppresses floaters, achieving state-of-the-art performance in terms of rendering quality.
Meanwhile, although our method consumes slightly more memory consumption, the increased points are mainly distributed in areas with insufficient initializing points, which are necessary for high-quality reconstruction, and our method still maintains real-time rendering speed.
Finally, our method is more robust to the number of initialization points, thanks to our effective pixel-aware gradient and scaled gradient field.

%% file: sections/table_and_figure/360scene_compare.tex
\begin{table}[h]
	\centering
	\small
	\caption{\textbf{Per-scene quantitative results from the Mip-NeRF 360.}}
	\label{table:sub1}
	\scalebox{0.73}{
		\begin{tabular}{l|ccccccccc}
			\multicolumn{10}{c}{PSNR$\uparrow$}                                                                                                                                                                        \\
			\hline
			         & Bicycle                               & Flowers                              & Garden                               & Stump                              & Treehill                               & Room                          & Counter                                & Kitchen &Bonsai                               \\
			\hline
			Plenoxels  &21.912&20.097&23.4947&20.661&22.248&27.594&23.624&23.420&24.669 \\
			INGP-Base  &22.193&20.348&24.599&23.626&22.364&29.269&26.439&28.548&30.337 \\
			INGP-Big    &22.171&20.652&25.069&23.466&22.373&29.690&26.691&29.479&30.685 \\
                Mip-NeRF 360  &24.37&\cellcolor{yellow!40}21.73&26.98&26.40&\cellcolor{red!40}22.87&\cellcolor{orange!40}31.63&\cellcolor{red!40}29.55&\cellcolor{red!40}32.23&\cellcolor{red!40}33.46 \\
                \hline
                3DGS  &\cellcolor{yellow!40}25.246&21.520&\cellcolor{yellow!40}27.410&\cellcolor{yellow!40}26.550&22.490&30.632&28.700&30.317&31.980 \\
                3DGS$^*$   &\cellcolor{orange!40}25.634&\cellcolor{orange!40}21.892&\cellcolor{orange!40}27.742&\cellcolor{orange!40}26.897&\cellcolor{orange!40}22.802&\cellcolor{yellow!40}31.506&\cellcolor{yellow!40}29.123&\cellcolor{yellow!40}31.561&\cellcolor{yellow!40}32.184 \\
			Pixel-GS (Ours)          &\cellcolor{red!40}25.739&\cellcolor{red!40}21.940&\cellcolor{red!40}27.834&\cellcolor{red!40}27.111&\cellcolor{yellow!40}22.597&\cellcolor{red!40}31.794&\cellcolor{orange!40}29.299&\cellcolor{orange!40}31.956&\cellcolor{orange!40}32.697 \\

			\hline
			\multicolumn{10}{c}{SSIM$\uparrow$}                                                                                                                                                                        \\
			\hline
			         & Bicycle                               & Flowers                              & Garden                               & Stump                              & Treehill                               & Room                          & Counter                                & Kitchen &Bonsai                               \\
                \hline
			Plenoxels &0.496&0.431&0.6063&0.523&0.509&0.8417&0.759&0.648&0.814 \\
		      INGP-Base &0.491&0.450&0.649&0.574&0.518&0.855&0.798&0.818&0.890 \\
		      INGP-Big &0.512&0.486&0.701&0.594&0.542&0.871&0.817&0.858&0.906 \\
		      Mip-NeRF360 &0.685&0.583&0.813&0.744&0.632&0.913&0.894&0.920&\cellcolor{yellow!40}0.941 \\
        \hline
		      3DGS &\cellcolor{yellow!40}0.771&\cellcolor{yellow!40}0.605&\cellcolor{yellow!40}0.868&\cellcolor{yellow!40}0.775&\cellcolor{yellow!40}0.638&\cellcolor{yellow!40}0.914&\cellcolor{yellow!40}0.905&\cellcolor{yellow!40}0.922&0.938 \\
			3DGS$^*$ &\cellcolor{orange!40}0.778&\cellcolor{orange!40}0.622&\cellcolor{orange!40}0.873&\cellcolor{orange!40}0.785&\cellcolor{orange!40}0.652&\cellcolor{orange!40}0.926&\cellcolor{orange!40}0.915&\cellcolor{orange!40}0.933&\cellcolor{orange!40}0.947 \\
			Pixel-GS (Ours)&\cellcolor{red!40}0.793&\cellcolor{red!40}0.652&\cellcolor{red!40}0.878&\cellcolor{red!40}0.796&\cellcolor{red!40}0.653&\cellcolor{red!40}0.930&\cellcolor{red!40}0.921&\cellcolor{red!40}0.936&\cellcolor{red!40}0.951 \\

			\hline
			\multicolumn{10}{c}{LPIPS$\downarrow$}                                                                                                                                                                      \\
			\hline
			         & Bicycle                               & Flowers                              & Garden                               & Stump                              & Treehill                               & Room                          & Counter                                & Kitchen &Bonsai                               \\
                \hline
			Plenoxels &0.506&0.521&0.3864&0.503&0.540&0.4186&0.441&0.447&0.398 \\
	       INGP-Base &0.487&0.481&0.312&0.450&0.489&0.301&0.342&0.254&0.227 \\
	       INGP-Big &0.446&0.441&0.257&0.421&0.450&0.261&0.306&0.195&0.205 \\
	       Mip-NeRF360 &0.301&0.344&\cellcolor{yellow!40}0.170&0.261&0.339&\cellcolor{yellow!40}0.211&\cellcolor{yellow!40}0.204&\cellcolor{yellow!40}0.127&\cellcolor{yellow!40}0.176 \\
        \hline
	       3DGS &\cellcolor{yellow!40}0.205&\cellcolor{yellow!40}0.336&\cellcolor{orange!40}0.103&\cellcolor{yellow!40}0.210&\cellcolor{orange!40}0.317&0.220&\cellcolor{yellow!40}0.204&0.129&0.205 \\
			3DGS$^*$  &\cellcolor{orange!40}0.204&\cellcolor{orange!40}0.328&\cellcolor{red!40}0.094&\cellcolor{orange!40}0.207&\cellcolor{yellow!40}0.319&\cellcolor{orange!40}0.193&\cellcolor{orange!40}0.179&\cellcolor{orange!40}0.113&\cellcolor{orange!40}0.174 \\
			Pixel-GS (Ours)  &\cellcolor{red!40}0.173&\cellcolor{red!40}0.251&\cellcolor{red!40}0.094&\cellcolor{red!40}0.181&\cellcolor{red!40}0.269&\cellcolor{red!40}0.183&\cellcolor{red!40}0.162&\cellcolor{red!40}0.106&\cellcolor{red!40}0.162 \\

			\hline
		\end{tabular}
	}
\end{table}

%% file: sections/table_and_figure/tankscene_compare.tex
\begin{table}[h]
	\centering
	\small
	\caption{\textbf{Per-scene quantitative results from the Tanks~\&~Temples (Part 1).}}
	\label{table:sub2_part1}
	\scalebox{0.7}{
		\begin{tabular}{l|ccccccccccc}
			\multicolumn{12}{c}{PSNR$\uparrow$}                                                                                                                                                                        \\
			\hline
			        &Auditorium &Ballroom &Barn &Caterpillar &Church &Courthouse &Courtroom &Family &Francis &Horse &Ignatius \\
			\hline
			3DGS$^*$  &\cellcolor{orange!40}24.453&\cellcolor{red!40}25.263&\cellcolor{orange!40}28.475&\cellcolor{orange!40}23.749&\cellcolor{orange!40}23.371&\cellcolor{orange!40}22.426&\cellcolor{orange!40}23.534&\cellcolor{orange!40}25.313&\cellcolor{orange!40}27.772&\cellcolor{red!40}24.591&\cellcolor{red!40}22.528 \\
			Pixel-GS (Ours)  &\cellcolor{red!40}24.772&\cellcolor{orange!40}25.066&\cellcolor{red!40}28.997&\cellcolor{red!40}24.079&\cellcolor{red!40}23.679&\cellcolor{red!40}22.679&\cellcolor{red!40}23.603&\cellcolor{red!40}25.491&\cellcolor{red!40}28.580&\cellcolor{orange!40}24.552&\cellcolor{orange!40}22.347 \\

			\hline
			\multicolumn{12}{c}{SSIM$\uparrow$}                                                                                                                                                                        \\
			\hline
			        &Auditorium &Ballroom &Barn &Caterpillar &Church &Courthouse &Courtroom &Family &Francis &Horse &Ignatius \\
                \hline
			3DGS$^*$  &\cellcolor{orange!40}0.876&\cellcolor{red!40}0.860&\cellcolor{orange!40}0.869&\cellcolor{orange!40}0.811&\cellcolor{red!40}0.835&\cellcolor{orange!40}0.789&\cellcolor{orange!40}0.809&\cellcolor{orange!40}0.889&\cellcolor{orange!40}0.909&\cellcolor{red!40}0.897&\cellcolor{orange!40}0.814 \\
			Pixel-GS (Ours)  &\cellcolor{red!40}0.882&\cellcolor{orange!40}0.858&\cellcolor{red!40}0.888&\cellcolor{red!40}0.832&\cellcolor{orange!40}0.833&\cellcolor{red!40}0.795&\cellcolor{red!40}0.810&\cellcolor{red!40}0.892&\cellcolor{red!40}0.916&\cellcolor{red!40}0.897&\cellcolor{red!40}0.819 \\

			\hline
			\multicolumn{12}{c}{LPIPS$\downarrow$}                                                                                                                                                                      \\
			\hline
			    &Auditorium &Ballroom &Barn &Caterpillar &Church &Courthouse &Courtroom &Family &Francis &Horse &Ignatius \\
                \hline
			3DGS$^*$  &\cellcolor{orange!40}0.222&\cellcolor{orange!40}0.121&\cellcolor{orange!40}0.182&\cellcolor{orange!40}0.211&\cellcolor{red!40}0.198&\cellcolor{orange!40}0.255&\cellcolor{orange!40}0.191&\cellcolor{orange!40}0.123&\cellcolor{orange!40}0.240&\cellcolor{orange!40}0.132&\cellcolor{orange!40}0.165 \\
			Pixel-GS (Ours)  &\cellcolor{red!40}0.213&\cellcolor{red!40}0.120&\cellcolor{red!40}0.144&\cellcolor{red!40}0.173&\cellcolor{orange!40}0.202&\cellcolor{red!40}0.246&\cellcolor{red!40}0.189&\cellcolor{red!40}0.108&\cellcolor{red!40}0.228&\cellcolor{red!40}0.117&\cellcolor{red!40}0.149 \\

			\hline
		\end{tabular}
	}
\end{table}

\begin{table}[h]
	\centering
	\small
	\caption{\textbf{Per-scene quantitative results from the Tanks~\&~Temples (Part 2).}}
	\label{table:sub2_part2}
	\scalebox{0.7}{
		\begin{tabular}{l|cccccccccc}
			\multicolumn{11}{c}{PSNR$\uparrow$}                                                                                                                                                                        \\
			\hline
			        &Lighthouse &M60 &Meetingroom &Museum &Palace &Panther &Playground &Temple &Train &Truck \\
			\hline
			3DGS$^*$  &\cellcolor{orange!40}22.107&\cellcolor{orange!40}27.829&\cellcolor{orange!40}25.750&\cellcolor{red!40}21.337&\cellcolor{orange!40}19.675&\cellcolor{red!40}28.485&\cellcolor{orange!40}25.783&\cellcolor{orange!40}17.930&\cellcolor{orange!40}22.117&\cellcolor{orange!40}25.425 \\
			Pixel-GS (Ours)  &\cellcolor{red!40}22.144&\cellcolor{red!40}27.968&\cellcolor{red!40}25.835&\cellcolor{orange!40}21.252&\cellcolor{red!40}19.957&\cellcolor{orange!40}28.466&\cellcolor{red!40}26.120&\cellcolor{red!40}18.698&\cellcolor{red!40}22.125&\cellcolor{red!40}25.491 \\

			\hline
			\multicolumn{11}{c}{SSIM$\uparrow$}                                                                                                                                                                        \\
			\hline
			        &Lighthouse &M60 &Meetingroom &Museum &Palace &Panther &Playground &Temple &Train &Truck \\
                \hline
			3DGS$^*$  &\cellcolor{orange!40}0.842&\cellcolor{orange!40}0.902&\cellcolor{orange!40}0.879&\cellcolor{red!40}0.794&\cellcolor{orange!40}0.736&\cellcolor{orange!40}0.911&\cellcolor{orange!40}0.864&\cellcolor{orange!40}0.754&\cellcolor{orange!40}0.813&\cellcolor{orange!40}0.878 \\
Pixel-GS (Ours)  &\cellcolor{red!40}0.843&\cellcolor{red!40}0.907&\cellcolor{red!40}0.880&\cellcolor{orange!40}0.789&\cellcolor{red!40}0.744&\cellcolor{red!40}0.914&\cellcolor{red!40}0.881&\cellcolor{red!40}0.767&\cellcolor{red!40}0.823&\cellcolor{red!40}0.883 \\

			\hline
			\multicolumn{11}{c}{LPIPS$\downarrow$}                                                                                                                                                                      \\
			\hline
			    &Lighthouse &M60 &Meetingroom &Museum &Palace &Panther &Playground &Temple &Train &Truck \\
                \hline
			3DGS$^*$  &\cellcolor{orange!40}0.207&\cellcolor{orange!40}0.145&\cellcolor{orange!40}0.180&\cellcolor{red!40}0.191&\cellcolor{orange!40}0.332&\cellcolor{orange!40}0.140&\cellcolor{orange!40}0.168&\cellcolor{orange!40}0.307&\cellcolor{orange!40}0.209&\cellcolor{orange!40}0.148 \\
			Pixel-GS (Ours)  &\cellcolor{red!40}0.197&\cellcolor{red!40}0.120&\cellcolor{red!40}0.175&\cellcolor{orange!40}0.194&\cellcolor{red!40}0.315&\cellcolor{red!40}0.118&\cellcolor{red!40}0.137&\cellcolor{red!40}0.286&\cellcolor{red!40}0.180&\cellcolor{red!40}0.122 \\

			\hline
		\end{tabular}
	}
\end{table}

%% file: sections/table_and_figure/ablation_mip.tex
\begin{table}[h]
	\centering
	\small
	\caption{\textbf{Per-scene quantitative results from our ablation study on the Mip-NeRF 360.}}
	\label{table:sub3}
	\scalebox{0.73}{
		\begin{tabular}{l|ccccccccc}
			\multicolumn{10}{c}{PSNR$\uparrow$}                                                                                                                                                                        \\
			\hline
			   & Bicycle & Flowers & Garden & Stump  &Treehill & Room  & Counter  & Kitchen &Bonsai  \\
			\hline

               3DGS$^*$  &25.634&\cellcolor{yellow!40}21.892&\cellcolor{yellow!40}27.742&\cellcolor{yellow!40}26.897&\cellcolor{red!40}22.802&\cellcolor{yellow!40}31.506&29.123&\cellcolor{yellow!40}31.561&32.184 \\
                Pixel-aware Gradient &\cellcolor{orange!40}25.709&21.780&\cellcolor{red!40}27.868&\cellcolor{red!40}27.113&22.466&31.195&\cellcolor{orange!40}29.274&\cellcolor{orange!40}31.774&\cellcolor{orange!40}32.488 \\
                Scaled Gradient Field &\cellcolor{yellow!40}25.662&\cellcolor{orange!40}21.895&27.717&26.860&\cellcolor{orange!40}22.789&\cellcolor{orange!40}31.659&\cellcolor{yellow!40}29.138&31.516&\cellcolor{yellow!40}32.282 \\
			Complete Model  &\cellcolor{red!40}25.739&\cellcolor{red!40}21.940&\cellcolor{orange!40}27.834&\cellcolor{orange!40}27.111&\cellcolor{yellow!40}22.597&\cellcolor{red!40}31.794&\cellcolor{red!40}29.299&\cellcolor{red!40}31.956&\cellcolor{red!40}32.679 \\

			\hline
			\multicolumn{10}{c}{SSIM$\uparrow$}                                                                                                                                                                        \\
			\hline
			   & Bicycle & Flowers & Garden & Stump  &Treehill & Room  & Counter  & Kitchen &Bonsai\\
                \hline

			3DGS$^*$  &0.778&\cellcolor{yellow!40}0.622&\cellcolor{orange!40}0.873&\cellcolor{yellow!40}0.785&\cellcolor{orange!40}0.652&\cellcolor{yellow!40}0.926&\cellcolor{yellow!40}0.915&\cellcolor{orange!40}0.933&\cellcolor{orange!40}0.947 \\
                Pixel-aware Gradient &\cellcolor{orange!40}0.792&\cellcolor{orange!40}0.651&\cellcolor{red!40}0.878&\cellcolor{orange!40}0.795&0.650&0.922&\cellcolor{orange!40}0.920&\cellcolor{red!40}0.936&\cellcolor{red!40}0.951 \\
                Scaled Gradient Field &\cellcolor{yellow!40}0.779&0.621&\cellcolor{orange!40}0.873&0.782&\cellcolor{yellow!40}0.651&\cellcolor{orange!40}0.927&\cellcolor{yellow!40}0.915&\cellcolor{yellow!40}0.932&\cellcolor{orange!40}0.947 \\
			Complete Model   &\cellcolor{red!40}0.793&\cellcolor{red!40}0.652&\cellcolor{red!40}0.878&\cellcolor{red!40}0.796&\cellcolor{red!40}0.653&\cellcolor{red!40}0.930&\cellcolor{red!40}0.921&\cellcolor{red!40}0.936&\cellcolor{red!40}0.951 \\

			\hline
			\multicolumn{10}{c}{LPIPS$\downarrow$}                                                                                                                                                                      \\
			\hline
			   & Bicycle & Flowers & Garden & Stump  &Treehill & Room  & Counter  & Kitchen &Bonsai  \\
                \hline

			3DGS$^*$  &\cellcolor{orange!40}0.204&\cellcolor{yellow!40}0.328&\cellcolor{yellow!40}0.103&\cellcolor{yellow!40}0.207&\cellcolor{yellow!40}0.319&0.193&\cellcolor{orange!40}0.179&\cellcolor{yellow!40}0.113&\cellcolor{yellow!40}0.174 \\
                Pixel-aware Gradient &\cellcolor{red!40}0.173&\cellcolor{red!40}0.248&\cellcolor{red!40}0.093&\cellcolor{orange!40}0.182&\cellcolor{red!40}0.269&\cellcolor{orange!40}0.189&\cellcolor{red!40}0.162&\cellcolor{orange!40}0.107&\cellcolor{red!40}0.161 \\
                Scaled Gradient Field &\cellcolor{yellow!40}0.205&0.329&\cellcolor{yellow!40}0.103&0.209&\cellcolor{orange!40}0.317&\cellcolor{yellow!40}0.192&\cellcolor{orange!40}0.179&0.114&\cellcolor{yellow!40}0.174 \\
			Complete Model  &\cellcolor{red!40}0.173&\cellcolor{orange!40}0.251&\cellcolor{orange!40}0.094&\cellcolor{red!40}0.181&\cellcolor{red!40}0.269&\cellcolor{red!40}0.183&\cellcolor{red!40}0.162&\cellcolor{red!40}0.106&\cellcolor{orange!40}0.162 \\

			\hline
		\end{tabular}
	}
\end{table}

%% file: sections/table_and_figure/ablation_tank.tex
\begin{table}[h]
	\centering
	\small
	\caption{\textbf{Per-scene quantitative results from our ablation study on the Tanks~\&~Temples (Part 1).}}
	\label{table:sub4_part1}
	\scalebox{0.65}{
		\begin{tabular}{l|ccccccccccc}
			\multicolumn{12}{c}{PSNR$\uparrow$}                                                                                                                                                                        \\
			\hline
			         &Auditorium &Ballroom &Barn &Caterpillar &Church &Courthouse &Courtroom &Family &Francis &Horse &Ignatius \\
			\hline
                3DGS$^*$  &\cellcolor{yellow!40}24.453&\cellcolor{red!40}25.263&\cellcolor{orange!40}28.475&\cellcolor{orange!40}23.749&\cellcolor{yellow!40}23.371&\cellcolor{orange!40}22.426&\cellcolor{yellow!40}23.534&\cellcolor{yellow!40}25.313&\cellcolor{yellow!40}27.772&\cellcolor{orange!40}24.591&\cellcolor{red!40}22.528 \\
                Pixel-aware Gradient &21.659&24.551&27.959&23.615&19.693&11.429&22.231&25.275&27.085&24.275&22.315 \\
                Scaled Gradient Field &\cellcolor{orange!40}24.677&\cellcolor{orange!40}25.238&\cellcolor{yellow!40}28.301&\cellcolor{yellow!40}23.655&\cellcolor{orange!40}23.438&\cellcolor{yellow!40}22.208&\cellcolor{red!40}23.616&\cellcolor{orange!40}25.356&\cellcolor{orange!40}28.525&\cellcolor{red!40}24.831&\cellcolor{orange!40}22.444 \\
                Complete Model  &\cellcolor{red!40}24.772&\cellcolor{yellow!40}25.066&\cellcolor{red!40}28.997&\cellcolor{red!40}24.079&\cellcolor{red!40}23.679&\cellcolor{red!40}22.679&\cellcolor{orange!40}23.603&\cellcolor{red!40}25.491&\cellcolor{red!40}28.580&\cellcolor{yellow!40}24.552&\cellcolor{yellow!40}22.347 \\

			\hline
			\multicolumn{12}{c}{SSIM$\uparrow$}                                                                                                                                                                        \\
			\hline
			         &Auditorium &Ballroom &Barn &Caterpillar &Church &Courthouse &Courtroom &Family &Francis &Horse &Ignatius \\
                \hline
                3DGS$^*$  &\cellcolor{yellow!40}0.876&\cellcolor{red!40}0.860&\cellcolor{yellow!40}0.869&\cellcolor{yellow!40}0.811&\cellcolor{red!40}0.835&\cellcolor{orange!40}0.789&\cellcolor{orange!40}0.809&\cellcolor{yellow!40}0.889&\cellcolor{yellow!40}0.909&\cellcolor{red!40}0.897&\cellcolor{orange!40}0.814 \\
                Pixel-aware Gradient &0.822&0.847&\cellcolor{orange!40}0.882&\cellcolor{orange!40}0.827&0.730&0.508&0.778&\cellcolor{orange!40}0.890&\cellcolor{orange!40}0.914&\cellcolor{orange!40}0.895&\cellcolor{red!40}0.819 \\
                Scaled Gradient Field &\cellcolor{orange!40}0.878&\cellcolor{orange!40}0.859&0.862&0.804&\cellcolor{yellow!40}0.832&\cellcolor{yellow!40}0.782&\cellcolor{yellow!40}0.808&\cellcolor{yellow!40}0.889&0.908&\cellcolor{red!40}0.897&\cellcolor{orange!40}0.814 \\
                Complete Model  &\cellcolor{red!40}0.882&\cellcolor{yellow!40}0.858&\cellcolor{red!40}0.888&\cellcolor{red!40}0.832&\cellcolor{orange!40}0.833&\cellcolor{red!40}0.795&\cellcolor{red!40}0.810&\cellcolor{red!40}0.892&\cellcolor{red!40}0.916&\cellcolor{red!40}0.897&\cellcolor{red!40}0.819 \\

			\hline
			\multicolumn{12}{c}{LPIPS$\downarrow$}                                                                                                                                                                      \\
			\hline
			     &Auditorium &Ballroom &Barn &Caterpillar &Church &Courthouse &Courtroom &Family &Francis &Horse &Ignatius \\
                \hline
                3DGS$^*$  &\cellcolor{orange!40}0.222&\cellcolor{orange!40}0.121&\cellcolor{yellow!40}0.182&\cellcolor{yellow!40}0.211&\cellcolor{red!40}0.198&\cellcolor{orange!40}0.255&\cellcolor{orange!40}0.191&0.123&\cellcolor{yellow!40}0.240&\cellcolor{yellow!40}0.132&\cellcolor{yellow!40}0.165 \\
                Pixel-aware Gradient &0.267&0.128&\cellcolor{red!40}0.135&\cellcolor{red!40}0.170&0.300&0.563&0.211&\cellcolor{orange!40}0.111&\cellcolor{red!40}0.226&\cellcolor{orange!40}0.120&\cellcolor{orange!40}0.151 \\
                Scaled Gradient Field &\cellcolor{yellow!40}0.225&\cellcolor{yellow!40}0.122&0.191&0.221&\cellcolor{yellow!40}0.208&\cellcolor{yellow!40}0.270&\cellcolor{yellow!40}0.194&\cellcolor{yellow!40}0.122&0.242&\cellcolor{yellow!40}0.132&0.166 \\
                Complete Model  &\cellcolor{red!40}0.213&\cellcolor{red!40}0.120&\cellcolor{orange!40}0.144&\cellcolor{orange!40}0.173&\cellcolor{orange!40}0.202&\cellcolor{red!40}0.246&\cellcolor{red!40}0.189&\cellcolor{red!40}0.108&\cellcolor{orange!40}0.228&\cellcolor{red!40}0.117&\cellcolor{red!40}0.149 \\

			\hline
		\end{tabular}
	}
\end{table}

\begin{table}[h]
	\centering
	\small
	\caption{\textbf{Per-scene quantitative results from our ablation study on the Tanks~\&~Temples (Part 2).}}
	\label{table:sub4_part2}
	\scalebox{0.7}{
		\begin{tabular}{l|cccccccccc}
			\multicolumn{11}{c}{PSNR$\uparrow$}                                                                                                                                                                        \\
			\hline
			        &Lighthouse &M60 &Meetingroom &Museum &Palace &Panther &Playground &Temple &Train &Truck \\
			\hline
                3DGS$^*$  &\cellcolor{yellow!40}22.107&\cellcolor{orange!40}27.829&\cellcolor{yellow!40}25.750&\cellcolor{orange!40}21.337&\cellcolor{yellow!40}19.675&\cellcolor{orange!40}28.485&\cellcolor{yellow!40}25.783&\cellcolor{yellow!40}17.930&\cellcolor{yellow!40}22.117&\cellcolor{orange!40}25.425 \\
                Pixel-aware Gradient &15.264&26.628&25.340&20.795&13.116&27.922&25.500&10.839&17.204&25.248 \\
                Scaled Gradient Field &\cellcolor{red!40}22.500&\cellcolor{yellow!40}27.776&\cellcolor{red!40}26.041&\cellcolor{red!40}21.374&\cellcolor{red!40}20.023&\cellcolor{red!40}28.557&\cellcolor{red!40}26.129&\cellcolor{red!40}18.928&\cellcolor{red!40}22.179&\cellcolor{yellow!40}25.377 \\
                Complete Model  &\cellcolor{orange!40}22.144&\cellcolor{red!40}27.968&\cellcolor{orange!40}25.835&\cellcolor{yellow!40}21.252&\cellcolor{orange!40}19.957&\cellcolor{yellow!40}28.466&\cellcolor{orange!40}26.120&\cellcolor{orange!40}18.698&\cellcolor{orange!40}22.125&\cellcolor{red!40}25.491 \\

			\hline
			\multicolumn{11}{c}{SSIM$\uparrow$}                                                                                                                                                                        \\
			\hline
			        &Lighthouse &M60 &Meetingroom &Museum &Palace &Panther &Playground &Temple &Train &Truck \\
                \hline
                3DGS$^*$  &\cellcolor{orange!40}0.842&\cellcolor{orange!40}0.902&\cellcolor{yellow!40}0.879&\cellcolor{orange!40}0.794&\cellcolor{yellow!40}0.736&\cellcolor{orange!40}0.911&\cellcolor{yellow!40}0.864&\cellcolor{yellow!40}0.754&\cellcolor{orange!40}0.813&\cellcolor{yellow!40}0.878 \\
                Pixel-aware Gradient &0.691&0.895&0.874&0.782&0.593&0.909&\cellcolor{orange!40}0.871&0.511&0.690&\cellcolor{orange!40}0.881 \\
                Scaled Gradient Field &\cellcolor{yellow!40}0.839&\cellcolor{yellow!40}0.899&\cellcolor{red!40}0.882&\cellcolor{red!40}0.797&\cellcolor{orange!40}0.739&\cellcolor{yellow!40}0.910&0.862&\cellcolor{red!40}0.768&\cellcolor{yellow!40}0.804&0.877 \\
                Complete Model  &\cellcolor{red!40}0.843&\cellcolor{red!40}0.907&\cellcolor{orange!40}0.880&\cellcolor{yellow!40}0.789&\cellcolor{red!40}0.744&\cellcolor{red!40}0.914&\cellcolor{red!40}0.881&\cellcolor{orange!40}0.767&\cellcolor{red!40}0.823&\cellcolor{red!40}0.883 \\

			\hline
			\multicolumn{11}{c}{LPIPS$\downarrow$}                                                                                                                                                                      \\
			\hline
			    &Lighthouse &M60 &Meetingroom &Museum &Palace &Panther &Playground &Temple &Train &Truck \\
                \hline
                3DGS$^*$  &\cellcolor{orange!40}0.207&\cellcolor{yellow!40}0.145&\cellcolor{yellow!40}0.180&\cellcolor{orange!40}0.191&\cellcolor{yellow!40}0.332&\cellcolor{yellow!40}0.140&\cellcolor{yellow!40}0.168&\cellcolor{yellow!40}0.307&\cellcolor{orange!40}0.209&\cellcolor{yellow!40}0.148 \\
                Pixel-aware Gradient &0.373&\cellcolor{orange!40}0.128&\cellcolor{orange!40}0.177&0.196&0.509&\cellcolor{orange!40}0.124&\cellcolor{orange!40}0.147&0.555&0.312&\cellcolor{red!40}0.120 \\
                Scaled Gradient Field &\cellcolor{yellow!40}0.213&0.151&0.182&\cellcolor{red!40}0.190&\cellcolor{orange!40}0.331&0.141&0.173&\cellcolor{orange!40}0.296&\cellcolor{yellow!40}0.227&0.150 \\
                Complete Model  &\cellcolor{red!40}0.197&\cellcolor{red!40}0.120&\cellcolor{red!40}0.175&\cellcolor{yellow!40}0.194&\cellcolor{red!40}0.315&\cellcolor{red!40}0.118&\cellcolor{red!40}0.137&\cellcolor{red!40}0.286&\cellcolor{red!40}0.180&\cellcolor{orange!40}0.122 \\

			\hline
		\end{tabular}
	}
\end{table}

%% file: sections/table_and_figure/sub_qua.tex
\begin{figure}[t]
    \centering
    \tiny
    \resizebox{1.0\linewidth}{!}{
        \begin{tabular}{*{4}{>{\centering\arraybackslash}m{0.25\linewidth}}}
        \includegraphics[width=\linewidth]{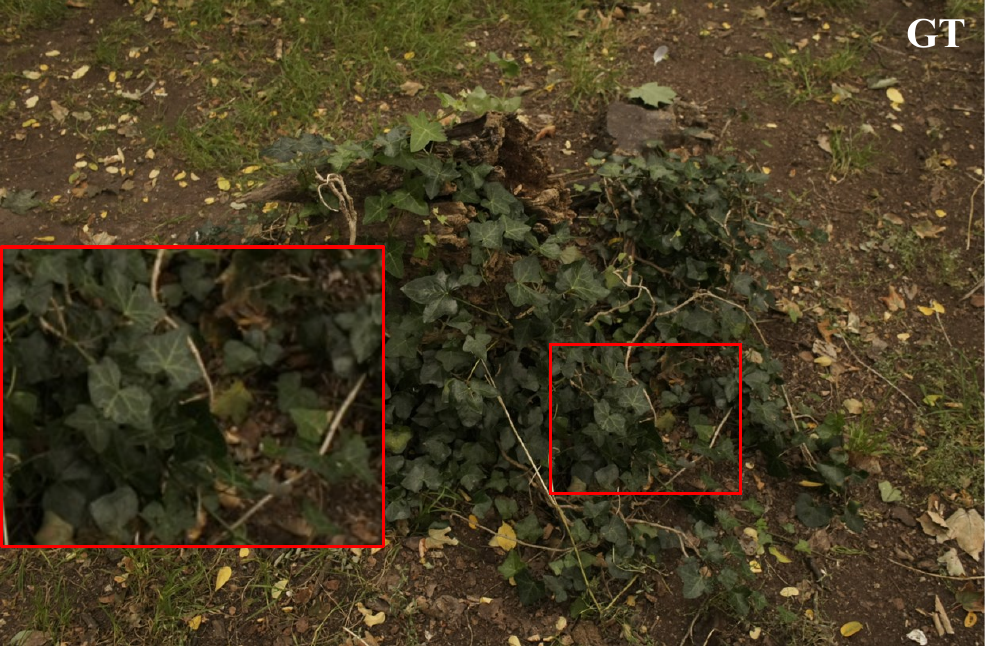} &
        \includegraphics[width=\linewidth]{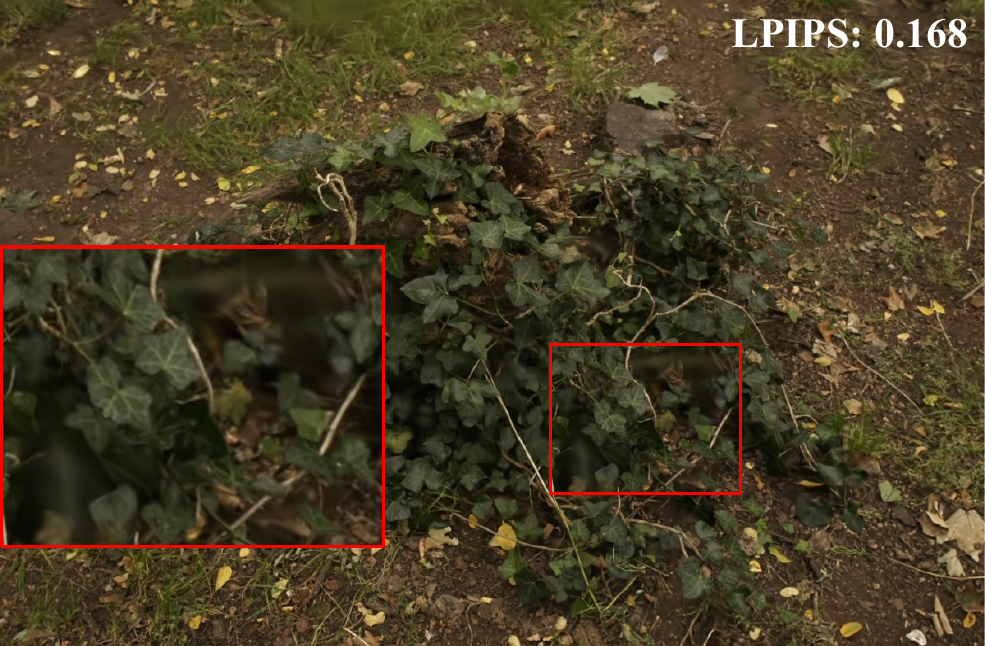} &
        \includegraphics[width=\linewidth]{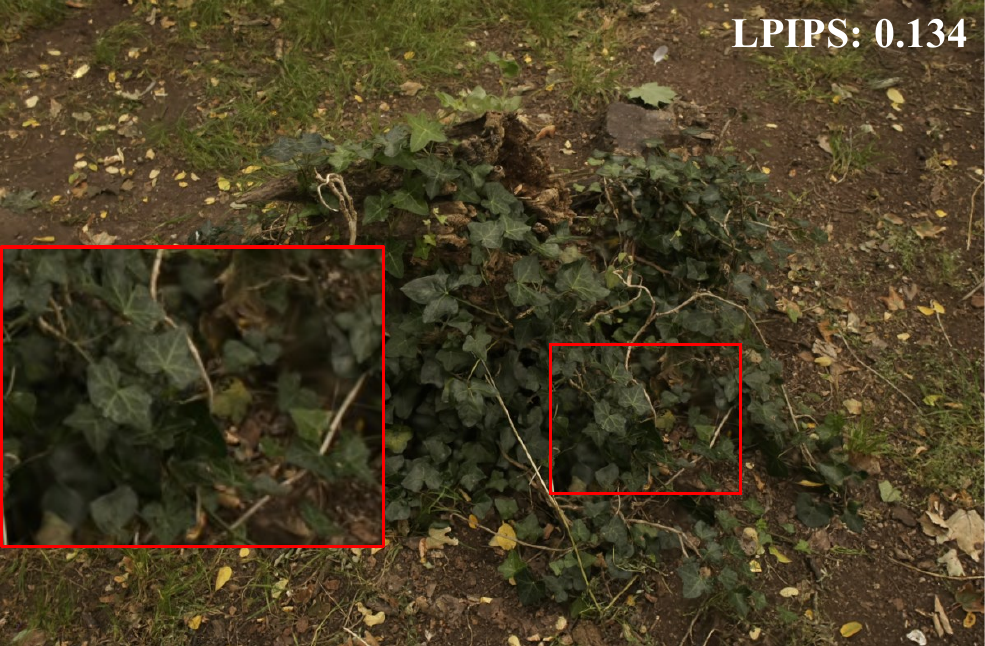} &
        \includegraphics[width=\linewidth]{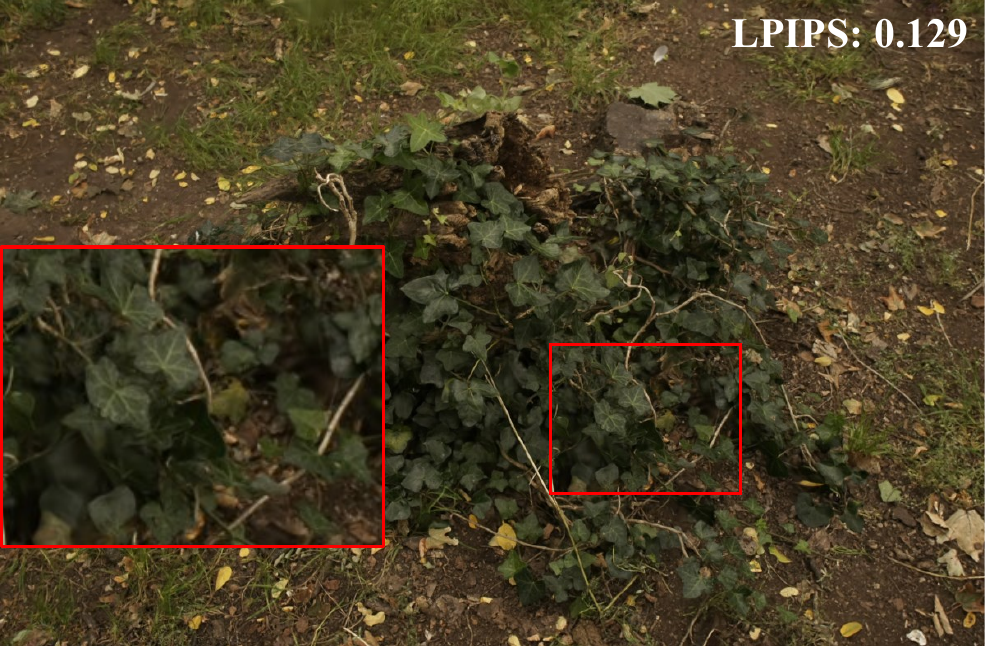} \\
        \includegraphics[width=\linewidth]{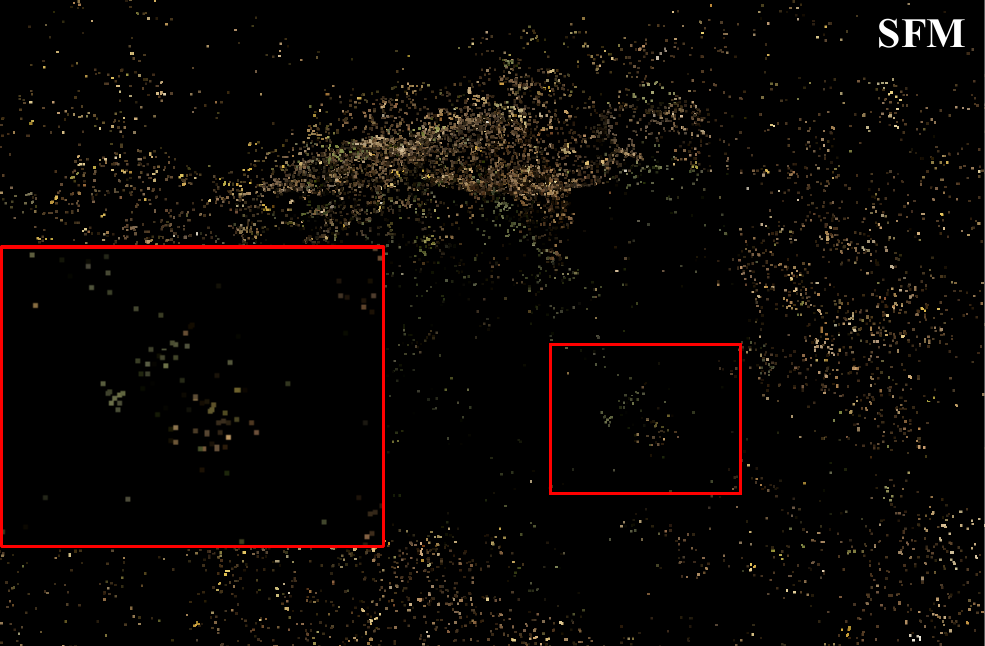} &
        \includegraphics[width=\linewidth]{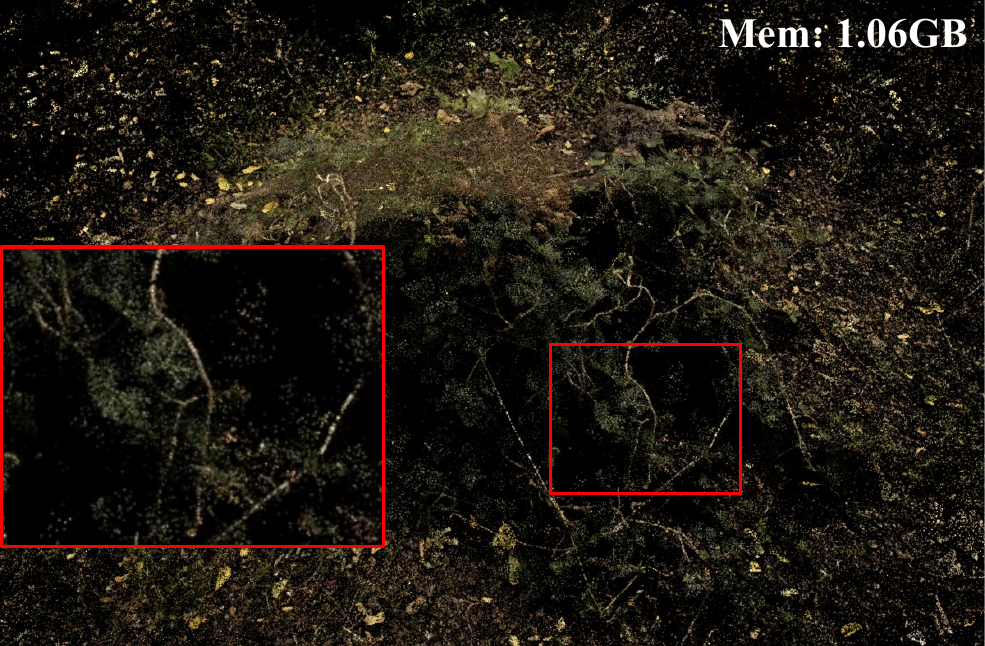} &
        \includegraphics[width=\linewidth]{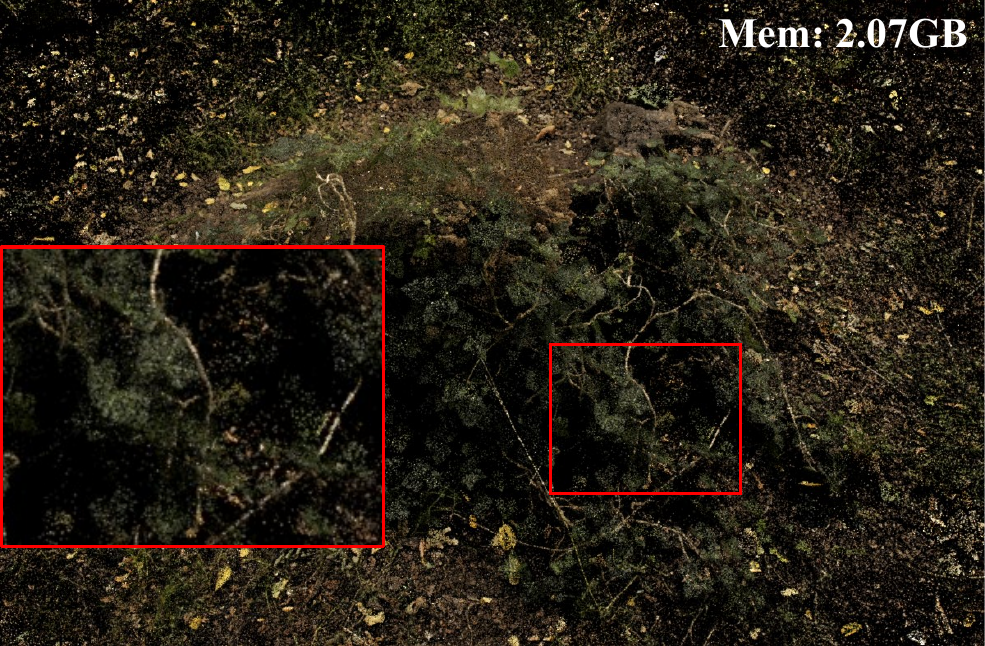} &
        \includegraphics[width=\linewidth]{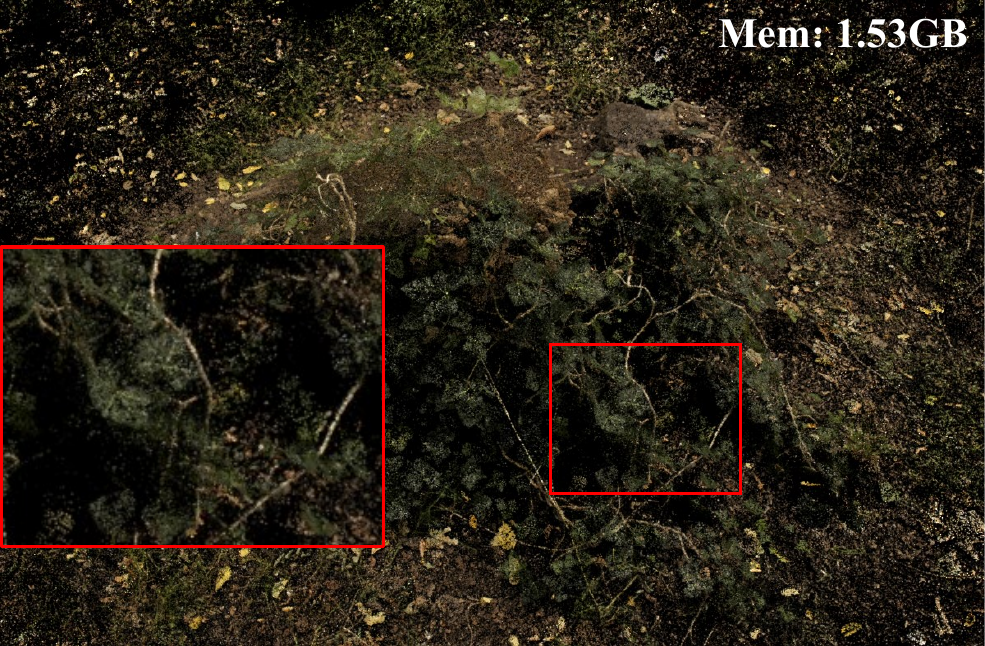} \\
        \includegraphics[width=\linewidth]{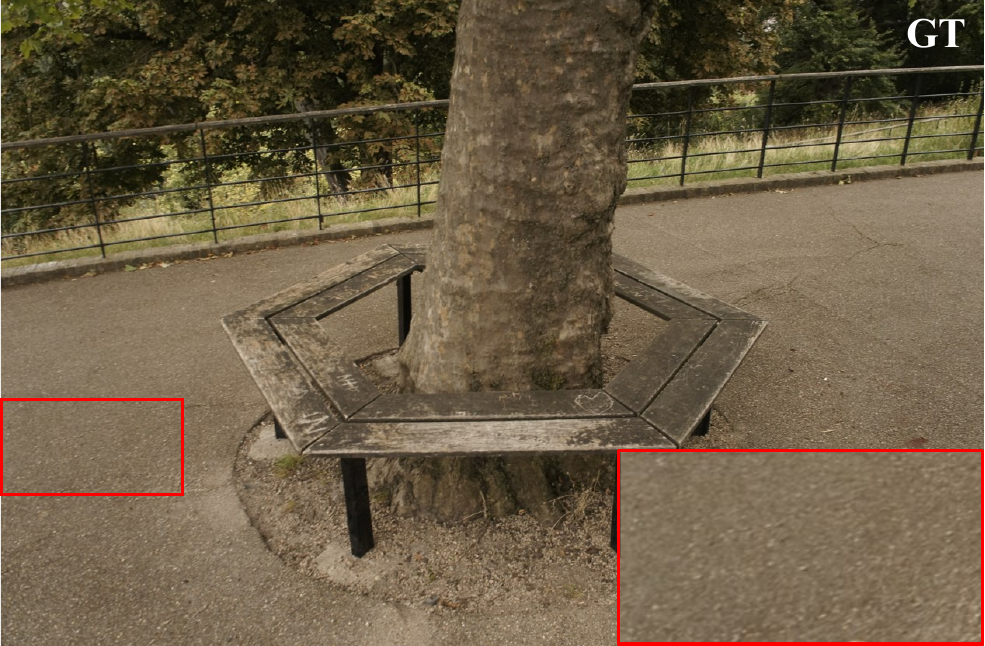} &
        \includegraphics[width=\linewidth]{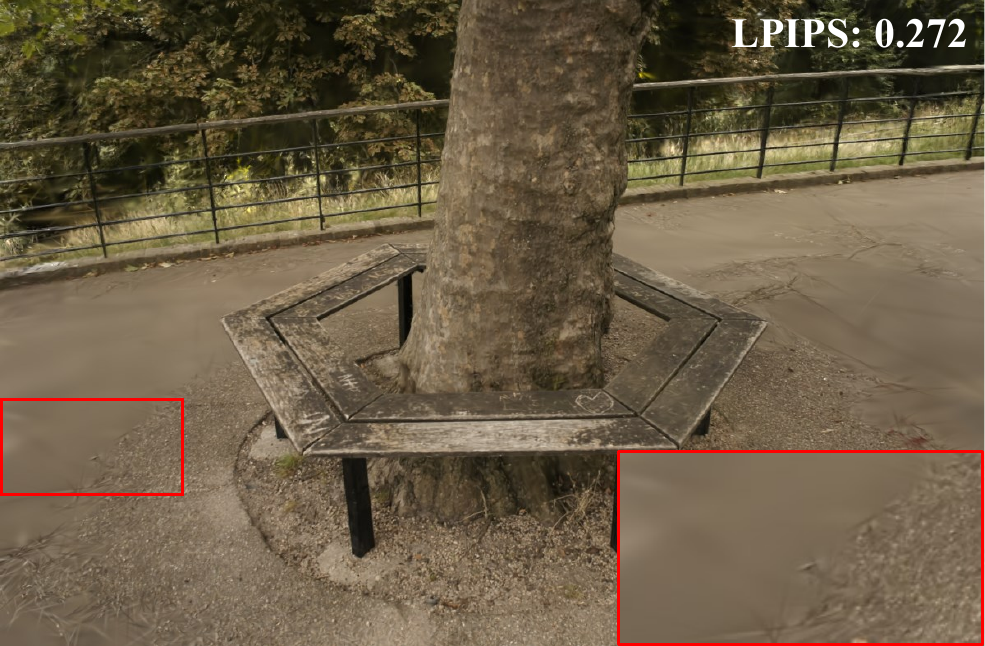} &
        \includegraphics[width=\linewidth]{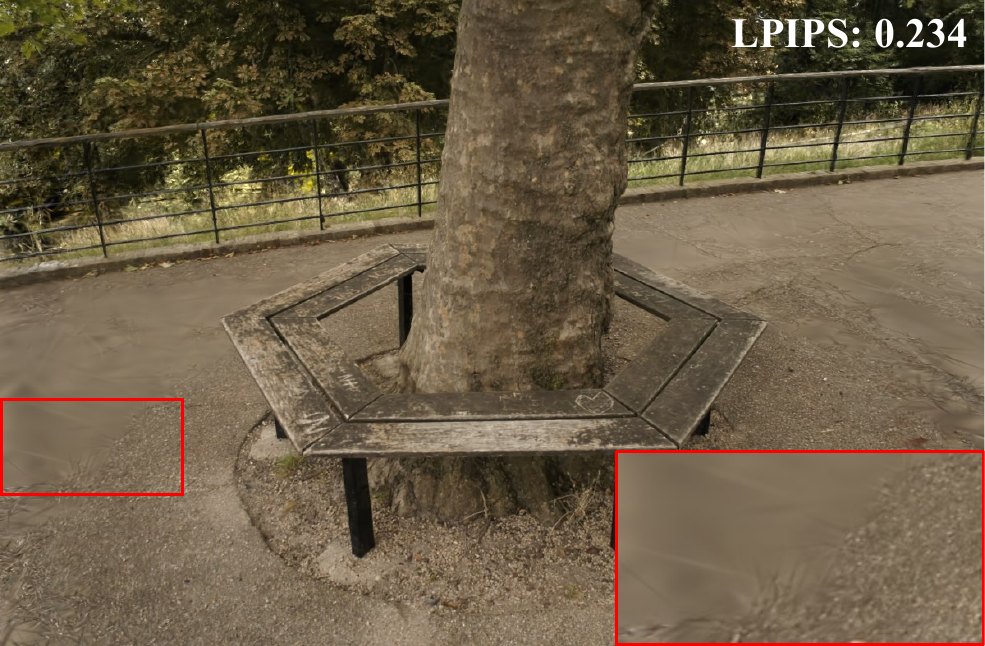} &
        \includegraphics[width=\linewidth]{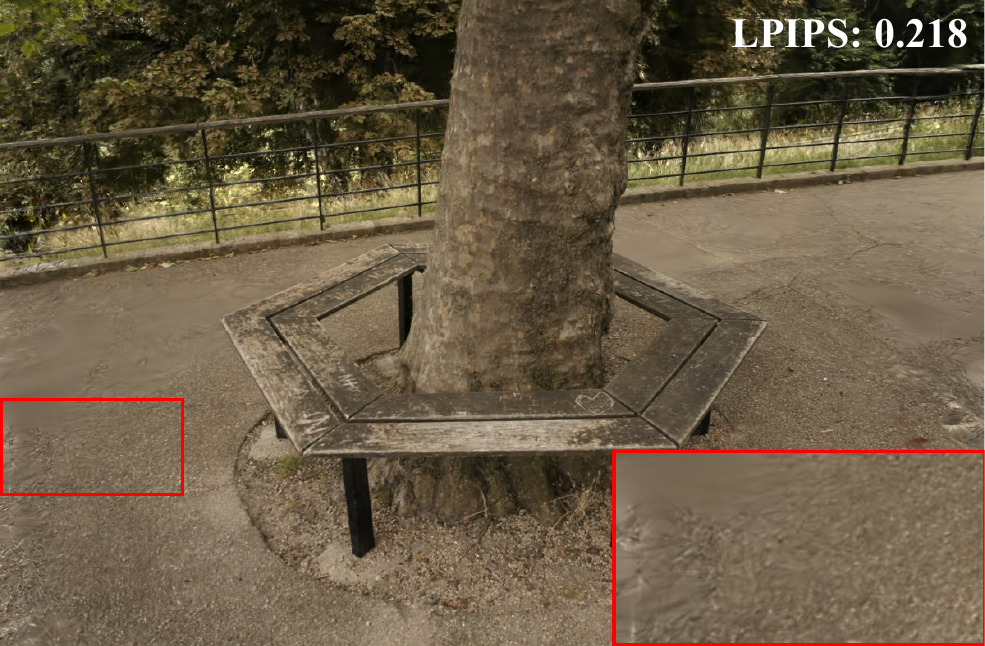} \\
        \includegraphics[width=\linewidth]{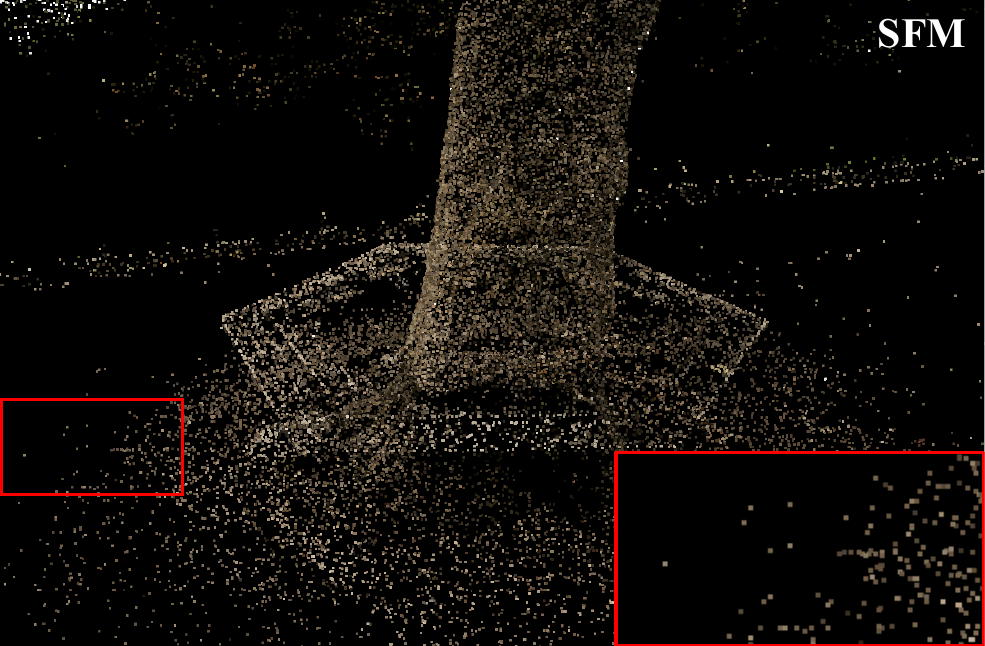} &
        \includegraphics[width=\linewidth]{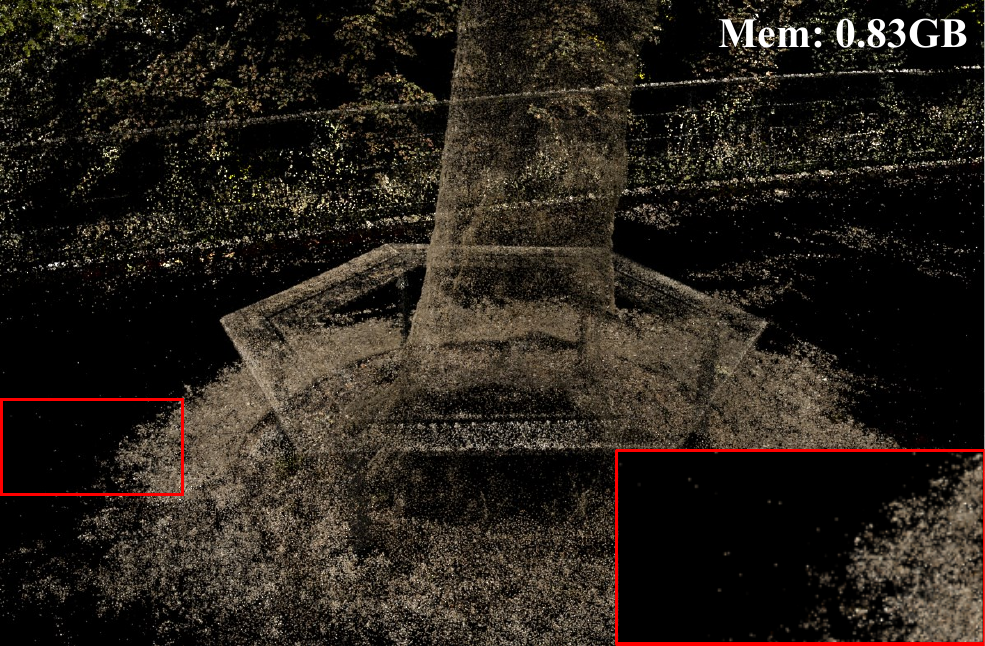} &
        \includegraphics[width=\linewidth]{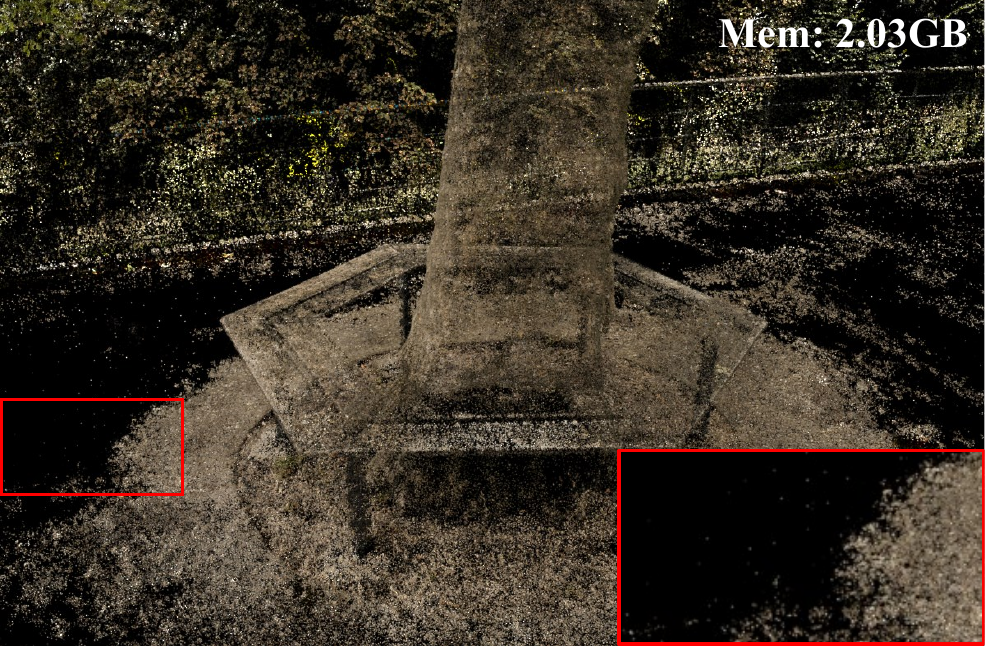} &
        \includegraphics[width=\linewidth]{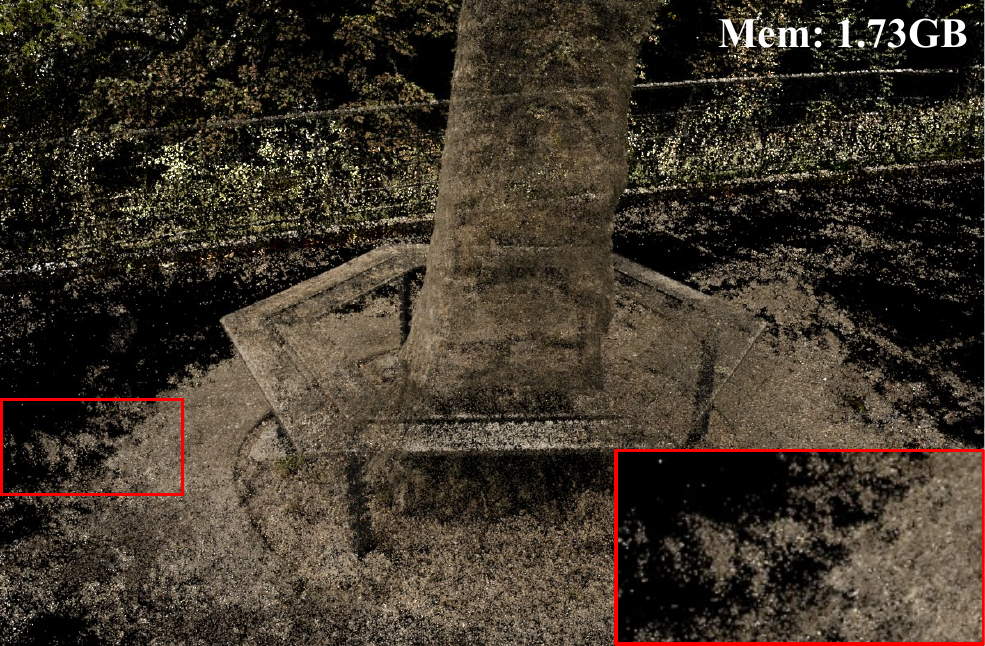} \\
        \includegraphics[width=\linewidth]{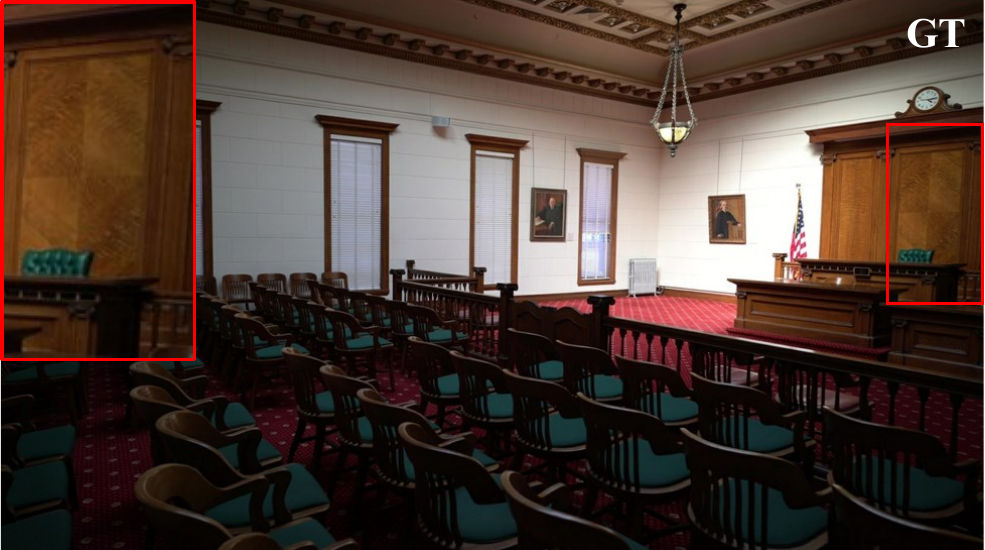} &
        \includegraphics[width=\linewidth]{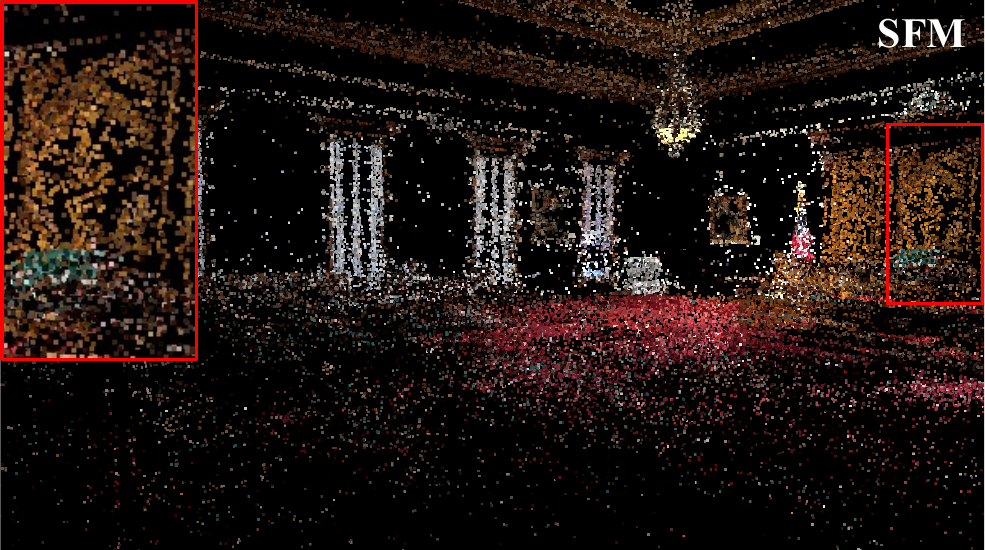} &
        \includegraphics[width=\linewidth]{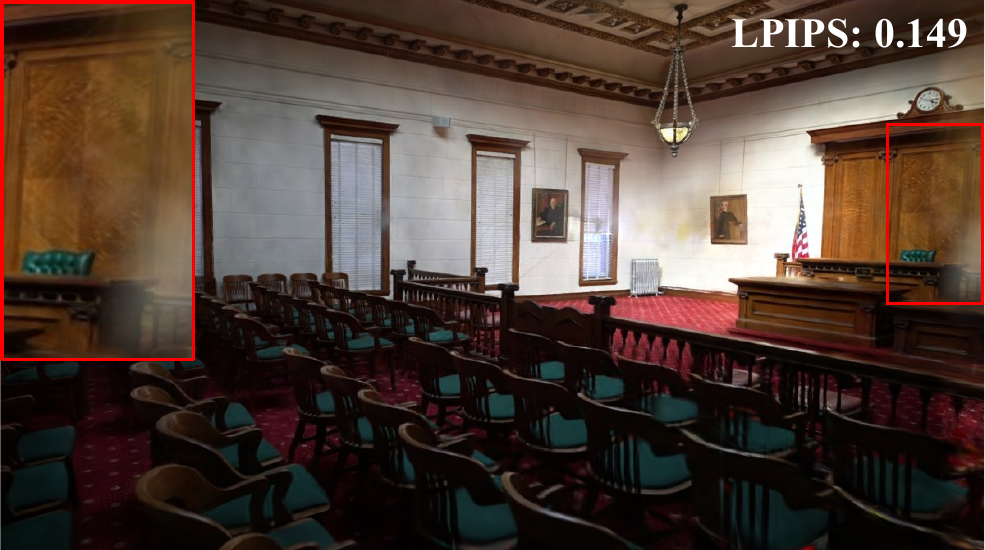} &
        \includegraphics[width=\linewidth]{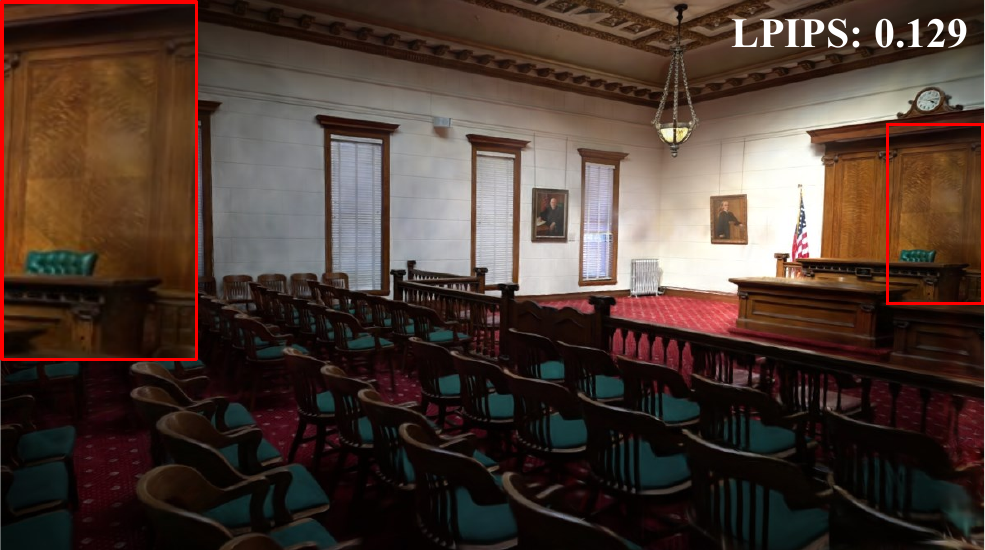} \\
        \includegraphics[width=\linewidth]{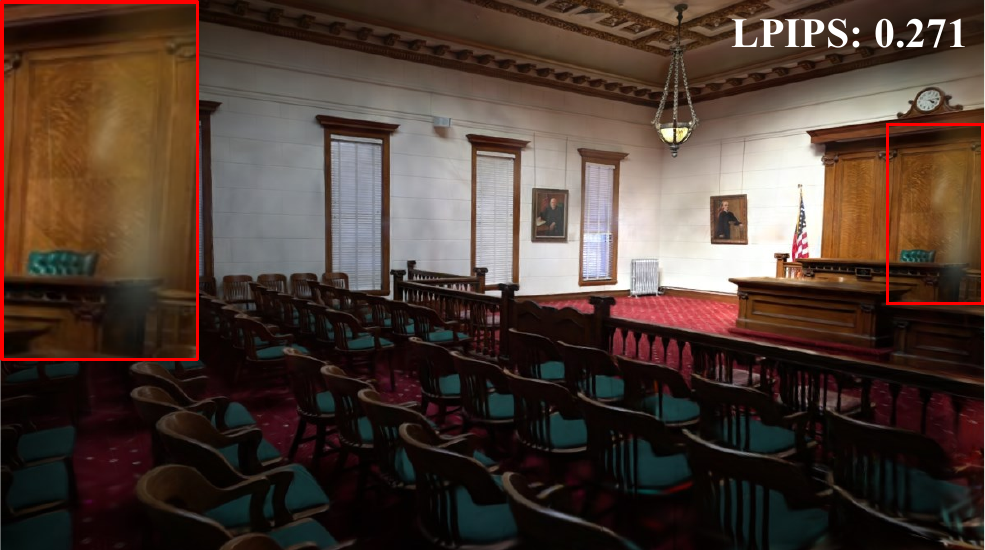} &
        \includegraphics[width=\linewidth]{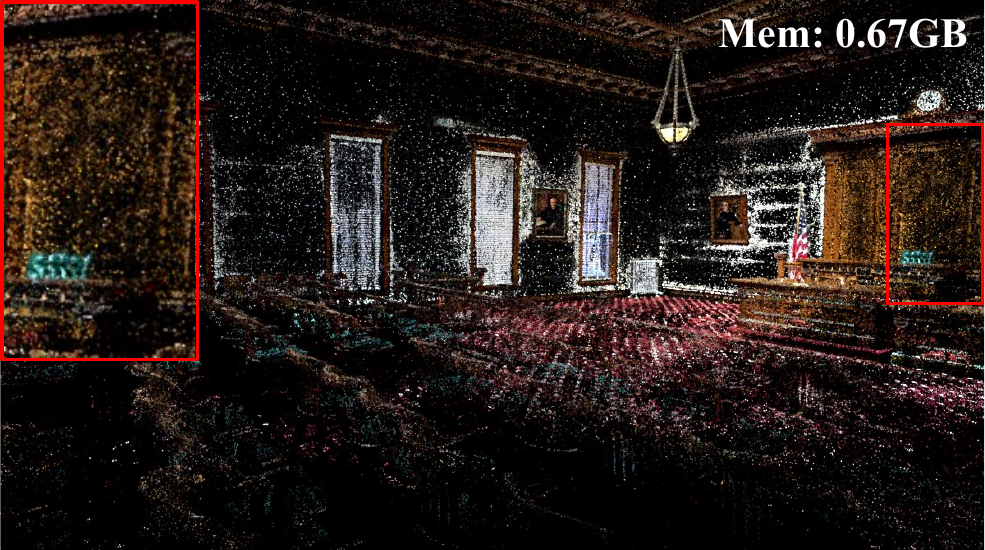} &
        \includegraphics[width=\linewidth]{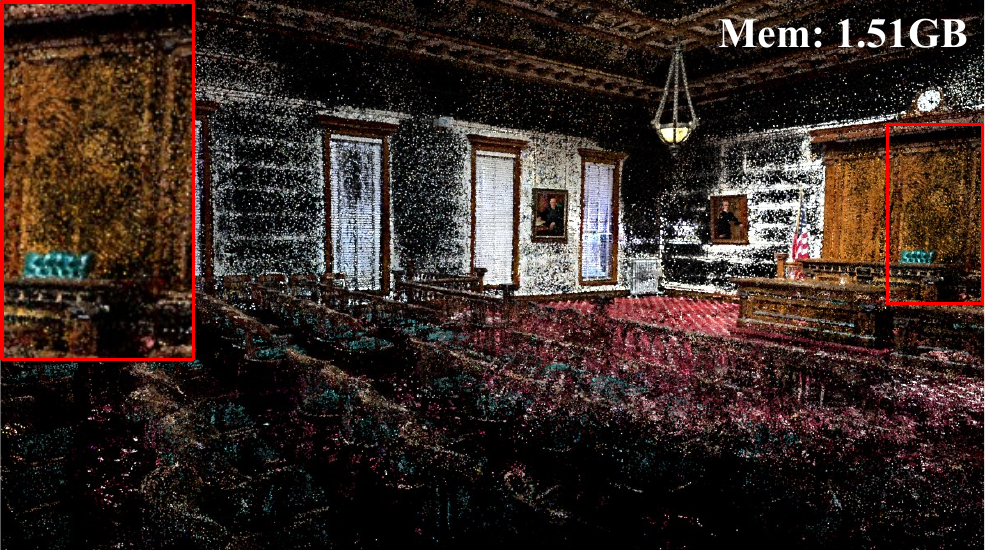} &
        \includegraphics[width=\linewidth]{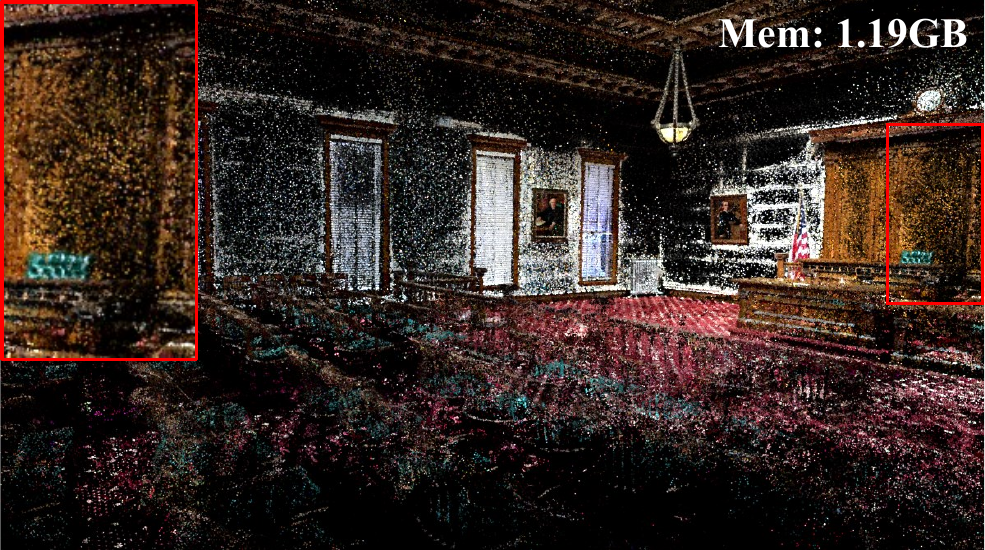} \\
        \includegraphics[width=\linewidth]{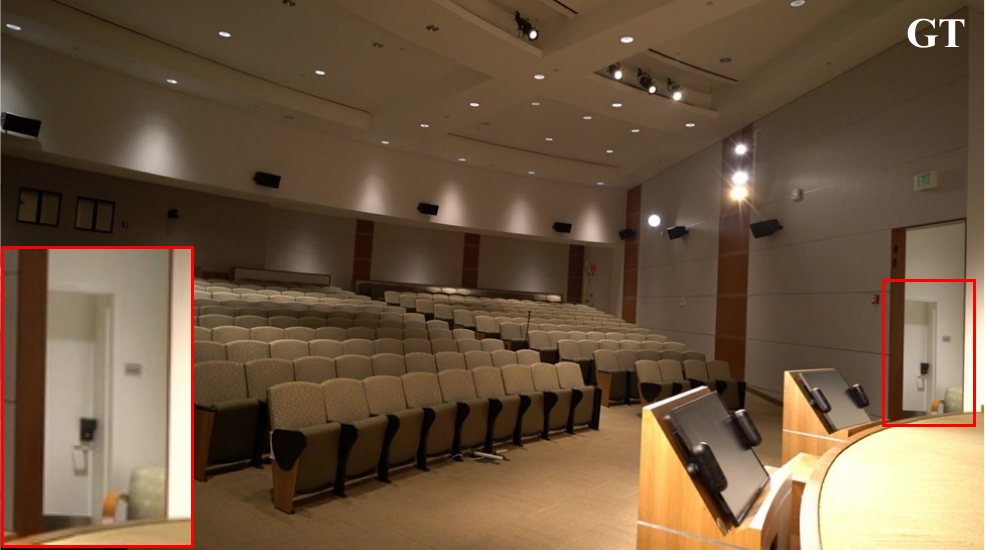} &
        \includegraphics[width=\linewidth]{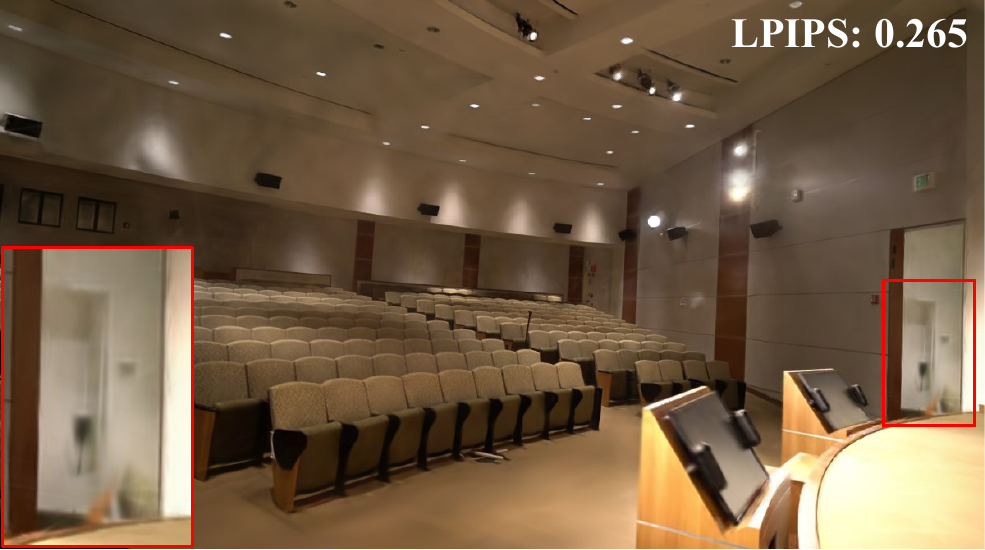} &
        \includegraphics[width=\linewidth]{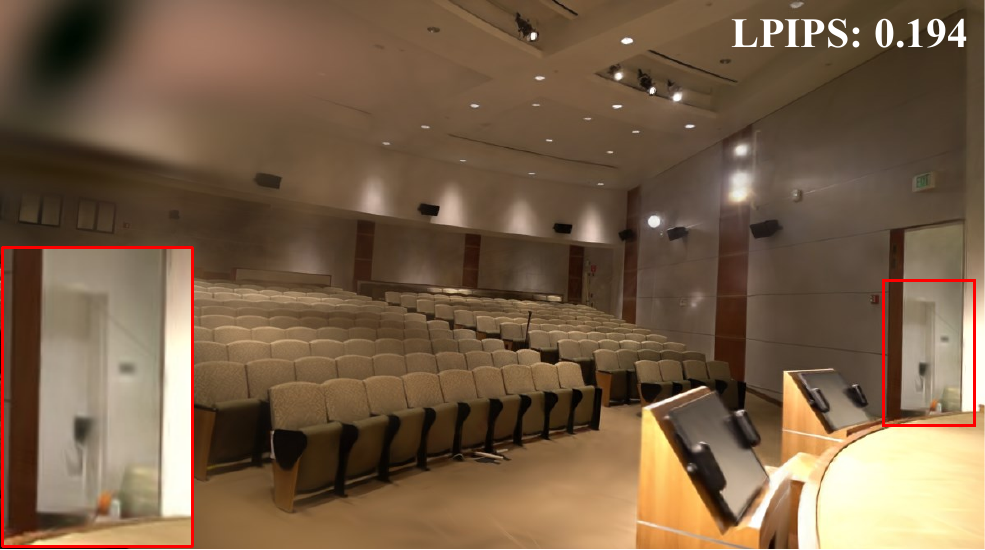} &
        \includegraphics[width=\linewidth]{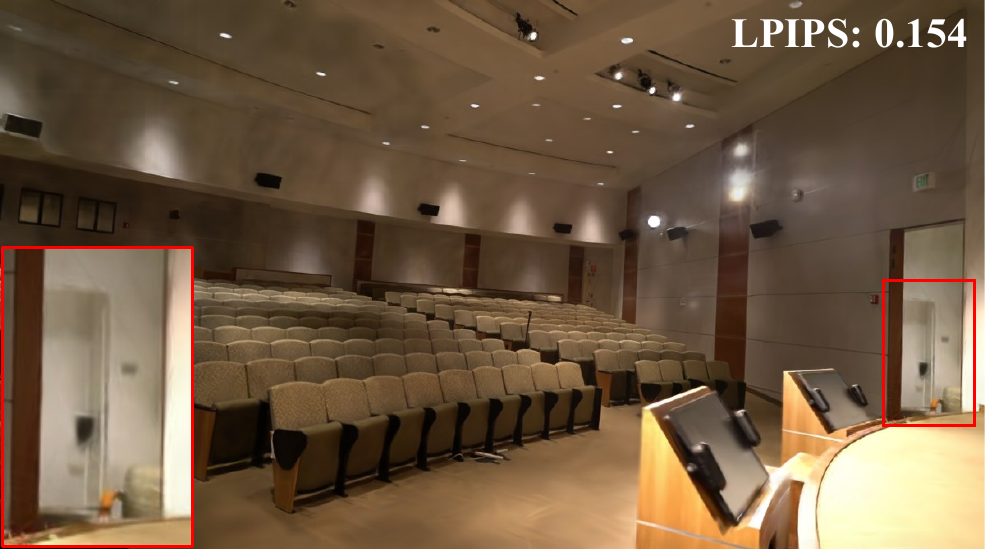} \\
        \includegraphics[width=\linewidth]{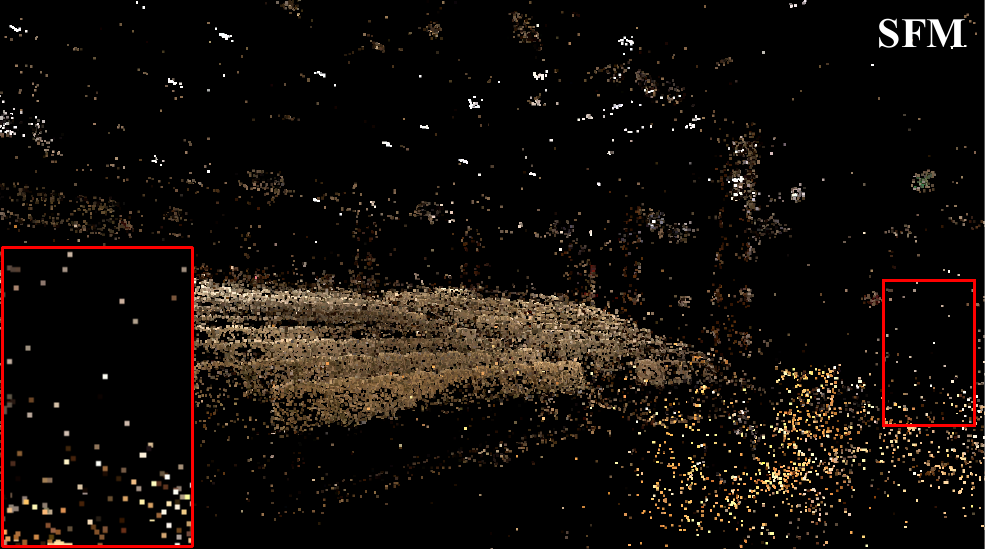} &
        \includegraphics[width=\linewidth]{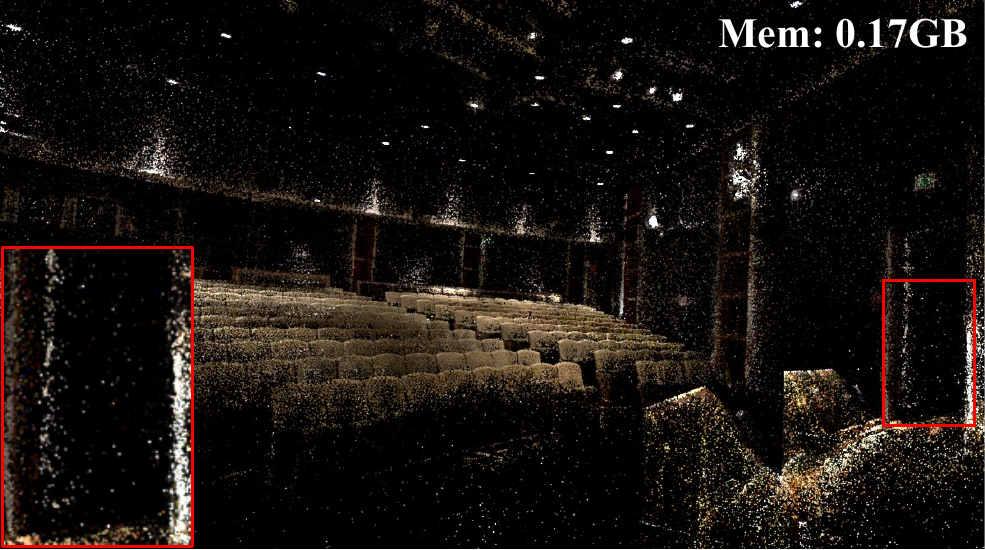} &
        \includegraphics[width=\linewidth]{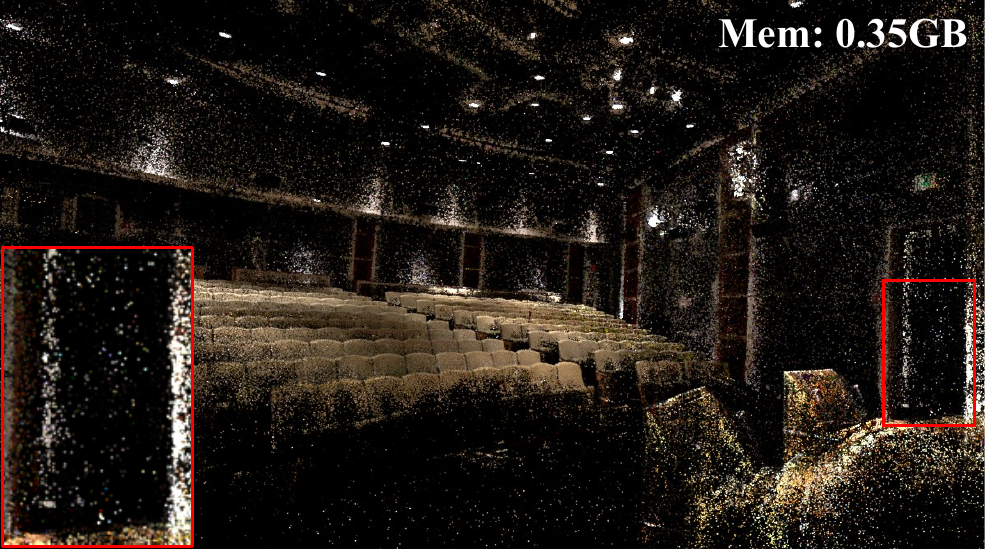} &
        \includegraphics[width=\linewidth]{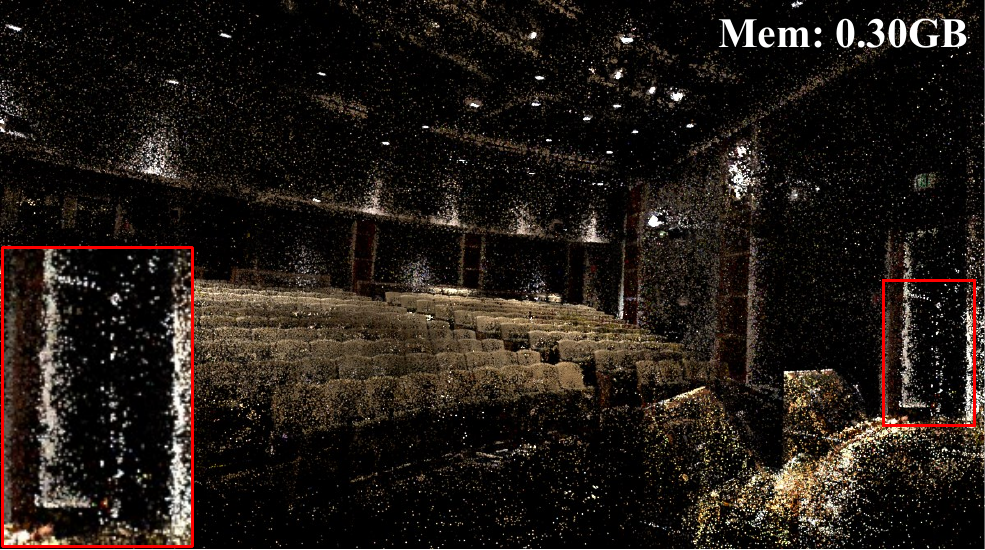} \\
        \includegraphics[width=\linewidth]{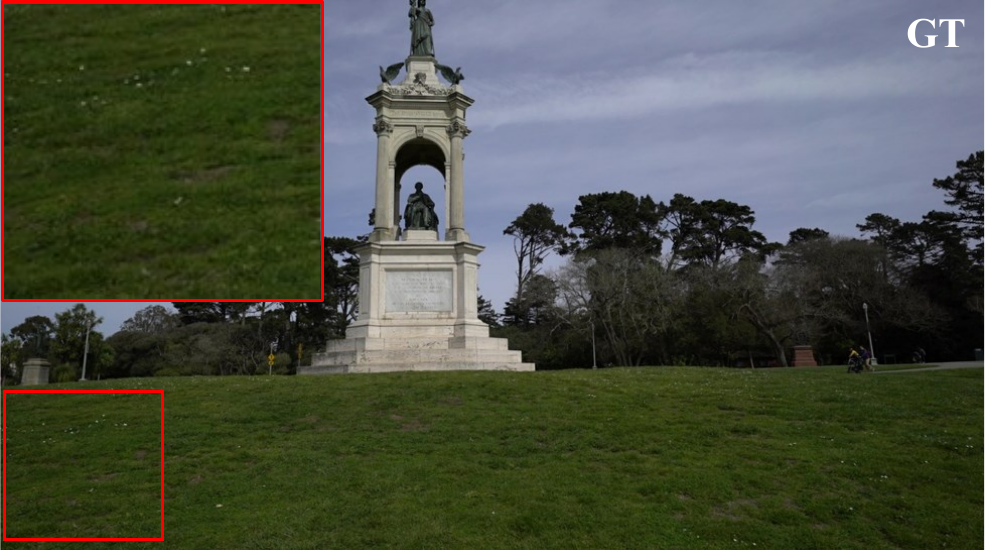} &
        \includegraphics[width=\linewidth]{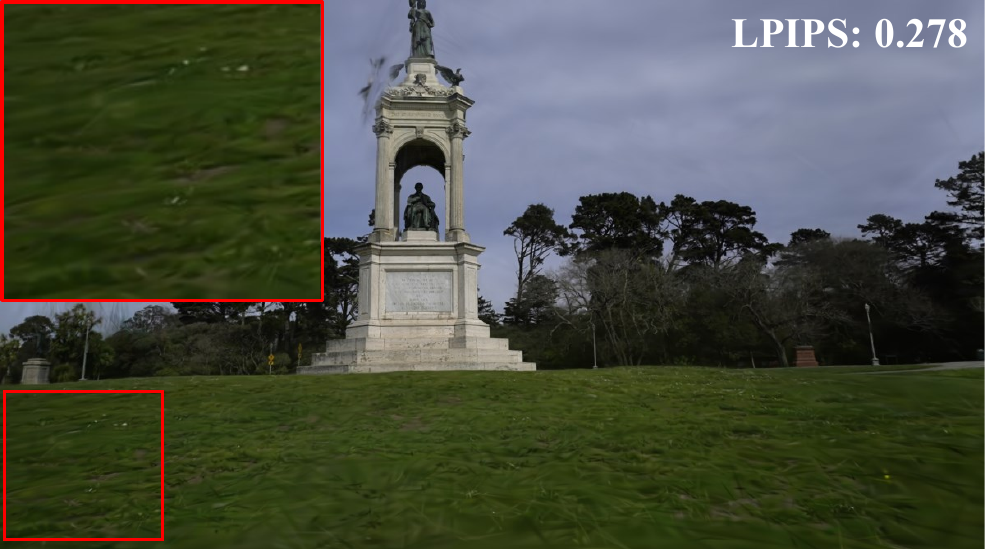} &
        \includegraphics[width=\linewidth]{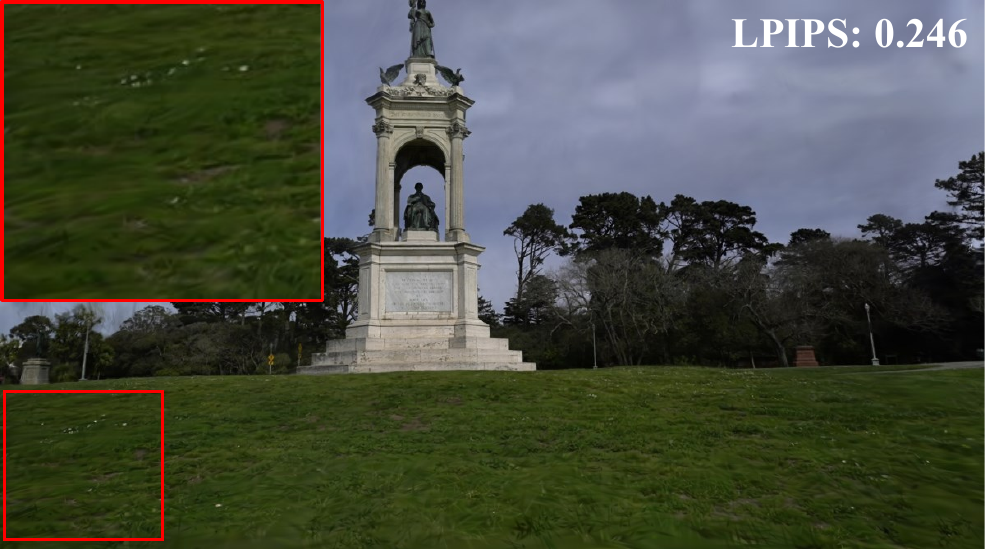} &
        \includegraphics[width=\linewidth]{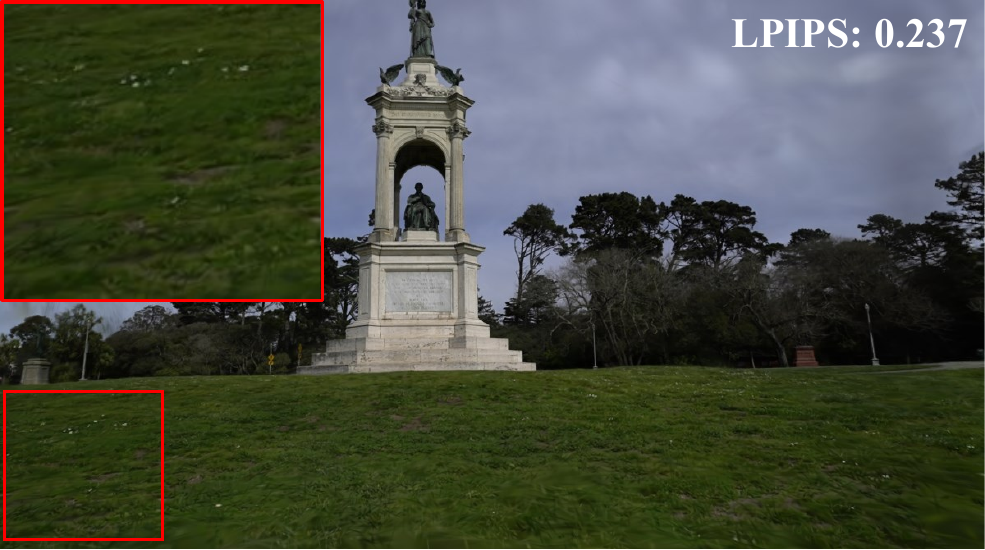} \\
        \includegraphics[width=\linewidth]{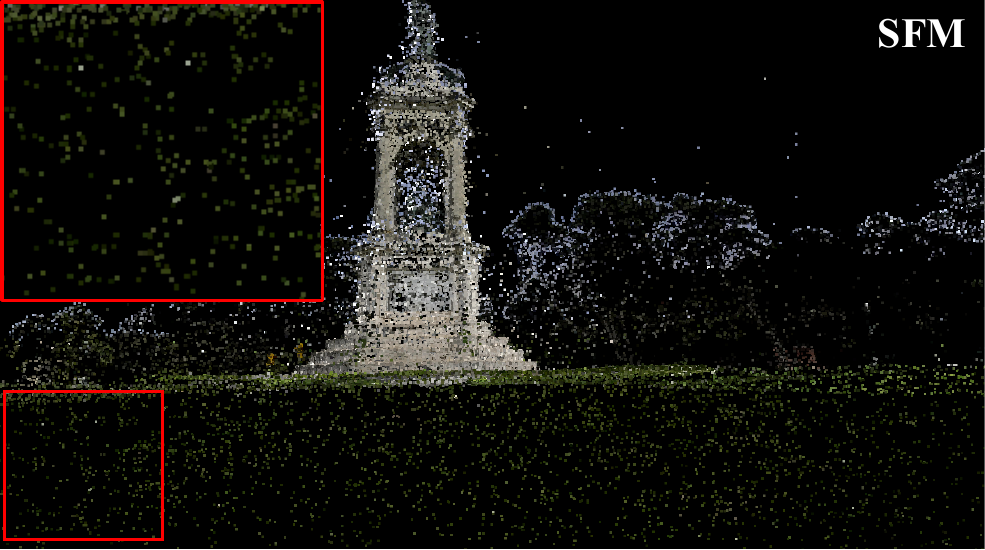} &
        \includegraphics[width=\linewidth]{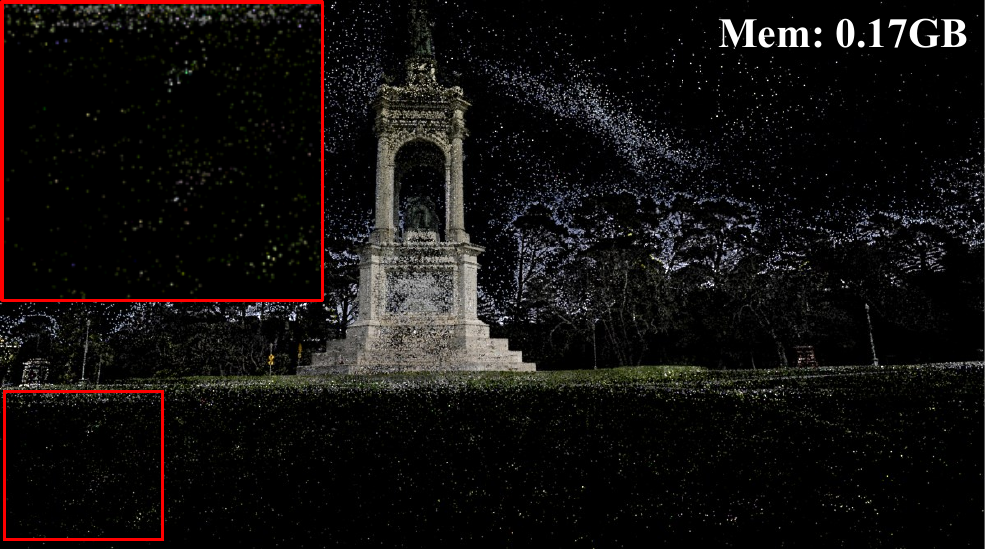} &
        \includegraphics[width=\linewidth]{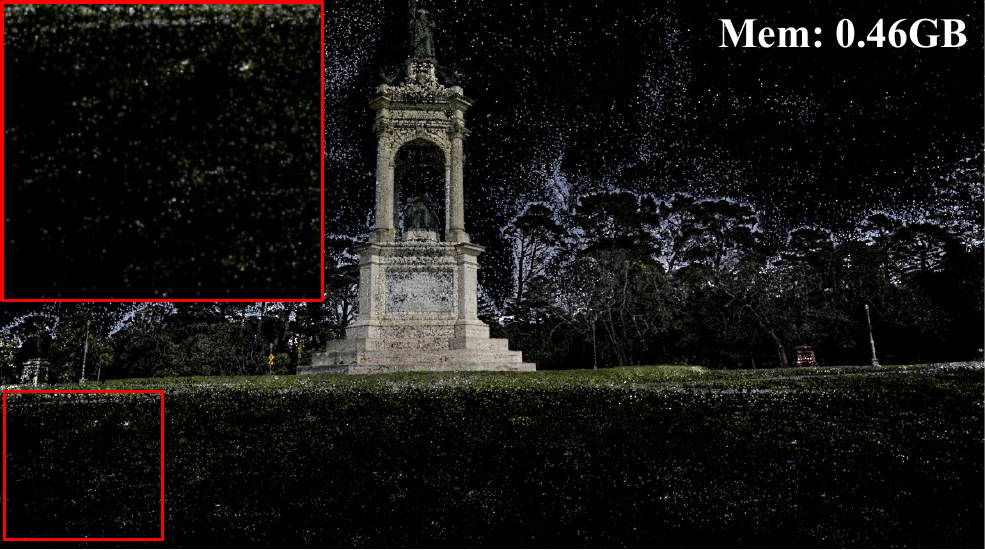} &
        \includegraphics[width=\linewidth]{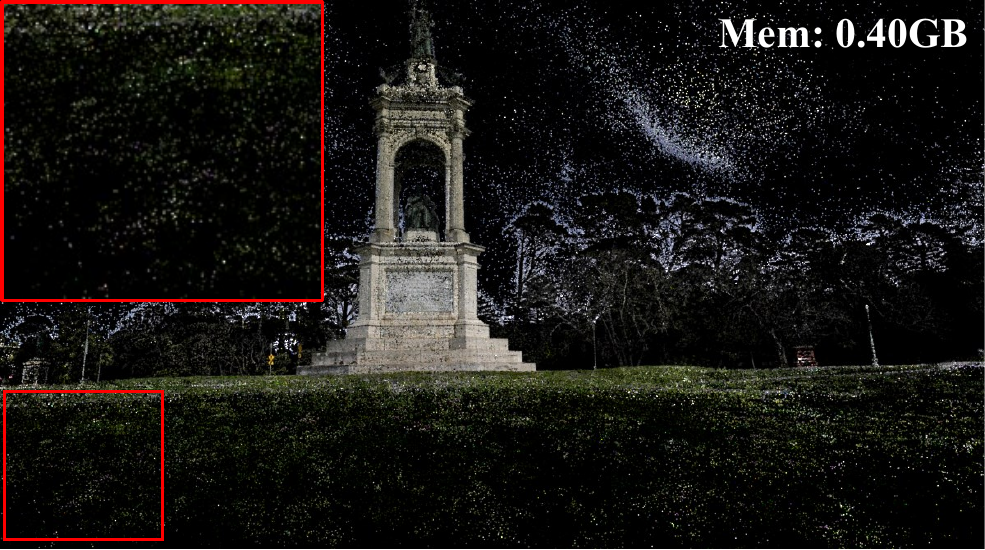} \\
        (a) Ground Truth & (b) 3DGS$^*$ (original threshold) & (c) 3DGS$^*$ (lower threshold) & (d) Pixel-GS (Ours) \\
    \end{tabular}}

    \vspace{-2mm} %

    \caption{
        Pixel-GS (d) enhances the modeling capability in areas with insufficient initial points (a). In contrast, 3DGS, even when using a lower point cloud growth threshold (c), still tends to grow additional points in areas where the initial points are denser.
    }
    \vspace{-3mm}
    \label{subfig}
\end{figure}